\pdfoutput=1
\documentclass[sigconf]{acmart}
\usepackage{colortbl}
\usepackage{balance}
\usepackage{booktabs} 
\usepackage{amsmath,color}
\usepackage{relsize}
\usepackage{makecell}
\usepackage{caption}
\usepackage{subcaption}
\usepackage{tablefootnote}
\usepackage{placeins}
\usepackage{afterpage}
\usepackage{float}
\usepackage{todonotes}
\usepackage{tabularx}
\newcolumntype{b}{X} 
\newcolumntype{s}{>{\hsize=.5\hsize}X} 
\usepackage{flushend}
\usepackage{mathtools}
\usepackage{multirow}
\usepackage{array}
\usepackage{makecell}
\usepackage[T1]{fontenc}
\usepackage{soul}
\usepackage{pgfplots}
\usepackage{pgfplotstable}
\usepackage[eulergreek]{sansmath}
\usepackage{balance}
\usepackage{arydshln}
\usepackage{xparse}
\usepackage{multicol}
\usepackage{tikz}
\usetikzlibrary{calc}

\pgfplotsset{
  tick label style = {font=\sansmath\sffamily},
  every axis label = {font=\sansmath\sffamily},
  legend style = {font=\sansmath\sffamily},
  label style = {font=\sansmath\sffamily},
  compat = 1.14
}

\newcommand{\ourheading}[1]{\vspace{1.5mm}
\noindent
{\bf #1.}}

\newcommand{\squishlist}{
  \begin{list}{$\bullet$}
    { \setlength{\itemsep}{0pt}      \setlength{\parsep}{3pt}
      \setlength{\topsep}{3pt}       \setlength{\partopsep}{0pt}
      \setlength{\leftmargin}{1.5em} \setlength{\labelwidth}{1em}
      \setlength{\labelsep}{0.5em} } }
\newcommand{\squishlisttwo}{
  \begin{list}{$\bullet$}
    { \setlength{\itemsep}{0pt}    \setlength{\parsep}{0pt}
      \setlength{\topsep}{0pt}     \setlength{\partopsep}{0pt}
      \setlength{\leftmargin}{0.9em} \setlength{\labelwidth}{0.5em}
      \setlength{\labelsep}{0.5em} } }

 \newcommand{\squishend}{
     \end{list} 
 }

\newcommand{\fillcol}{blue!20}
\newcommand{\bordercol}{blue}

\newcommand{\setfillcolor}[1]{\renewcommand{\fillcol}{#1}}
\newcommand{\setbordercolor}[1]{\renewcommand{\bordercol}{#1}}

\makeatletter
\tikzset{%
     remember picture with id/.style={%
       remember picture,
       overlay,
       save picture id=#1,
     },
     save picture id/.code={%
       \edef\pgf@temp{#1}%
       \immediate\write\pgfutil@auxout{%
         \noexpand\savepointas{\pgf@temp}{\pgfpictureid}}%
     },
     if picture id/.code args={#1#2#3}{%
       \@ifundefined{save@pt@#1}{%
         \pgfkeysalso{#3}%
       }{
         \pgfkeysalso{#2}%
       }
     }
   }

   \def\savepointas#1#2{%
  \expandafter\gdef\csname save@pt@#1\endcsname{#2}%
}

\def\tmk@labeldef#1,#2\@nil{%
  \def\tmk@label{#1}%
  \def\tmk@def{#2}%
}

\tikzdeclarecoordinatesystem{pic}{%
  \pgfutil@in@,{#1}%
  \ifpgfutil@in@%
    \tmk@labeldef#1\@nil
  \else
    \tmk@labeldef#1,(0pt,0pt)\@nil
  \fi
  \@ifundefined{save@pt@\tmk@label}{%
    \tikz@scan@one@point\pgfutil@firstofone\tmk@def
  }{%
  \pgfsys@getposition{\csname save@pt@\tmk@label\endcsname}\save@orig@pic%
  \pgfsys@getposition{\pgfpictureid}\save@this@pic%
  \pgf@process{\pgfpointorigin\save@this@pic}%
  \pgf@xa=\pgf@x
  \pgf@ya=\pgf@y
  \pgf@process{\pgfpointorigin\save@orig@pic}%
  \advance\pgf@x by -\pgf@xa
  \advance\pgf@y by -\pgf@ya
  }%
}

\NewDocumentCommand{\tikzmarkin}{m D(){0.825,-0.10} D(){-0.175,0.27}}{%
      \tikz[remember picture,overlay]
      \draw[line width=2.2pt,rectangle,fill=\fillcol,draw=\bordercol]
      (pic cs:#1) ++(#2) rectangle (#3)
      ;}

\newcommand\tikzmarkend[2][]{%
\tikz[remember picture with id=#2] #1;}

\begin{document}
\title{Story-oriented Image Selection and Placement}

\author{Sreyasi Nag Chowdhury\hspace{0.75cm} Simon Razniewski\hspace{0.75cm} Gerhard Weikum
\mbox{\large sreyasi, srazniew, weikum@mpi-inf.mpg.de}}
\affiliation{\institution{Max Planck Institute for Informatics, Saarbr{\"u}cken, Germany}}

\begin{abstract}

Multimodal contents have become commonplace on the Internet today, manifested as news articles, social media posts, and personal or business blog posts. Among the various kinds of media (images, videos, graphics, icons, audio) used in such multimodal stories, images are the most popular. The selection of images from a collection -- either author's personal photo album, or web repositories -- and their meticulous placement within a text, builds a succinct multimodal commentary for digital consumption. In this paper we present a system that automates the process of selecting relevant images for a story and placing them at contextual paragraphs within the story for a multimodal narration. We leverage automatic object recognition, user-provided tags, and commonsense knowledge, and use an unsupervised combinatorial optimization to solve the selection and placement problems seamlessly as a single unit.
\end{abstract}

\maketitle

\section{Introduction}

\vspace{0.5em}
It is well-known (and supported by
studies~\cite{lester2013visual, messaris2001role}) that the most powerful messages are delivered with a combination of words and pictures. On the Internet, such multimodal content is abundant in the form of news articles, social media posts, and personal blog posts where authors enrich their stories with carefully chosen and placed images. 
As an example, consider a vacation trip report, to be posted on a blog site or online community. The backbone of the travel report is a textual narration, but the user typically places illustrative images in appropriate spots, carefully selected from her photo collection from this trip. These images can either show specific highlights such as waterfalls, mountain hikes or animal encounters, or may serve to depict feelings and the general mood of the trip, e.g., by showing nice sunsets or bar scenes.
Another example is brochures for research institutes or other organizations. Here, the text describes the mission, achievements and ongoing projects, and it is accompanied with judiciously selected and placed photos of buildings, people, 
products and  other images depicting the subjects and phenomena of interest, e.g., galaxies or telescopes for research in astrophysics.

The generation of such multimodal stories requires substantial human judgement and reasoning, and is thus time-consuming and labor-intensive. In particular, the effort on the human side includes 
selecting the right images from a pool of story-specific photos
(e.g., the traveler's own photos)
and possibly also from a broader pool for visual illustration
(e.g., images licensed from a PR company's catalog or 
a big provider such as Pinterest).
Even if the set of photos were exactly given, there is still considerable effort to place them within or next to appropriate paragraphs, paying attention to the semantic coherence between surrounding text and image. 
In this paper, we set out to automate this human task, formalizing it as a {\em story-images alignment} problem.

\begin{figure}[t]
\begin{center}
\vspace{0.2cm}
\includegraphics[width=1.03\columnwidth]{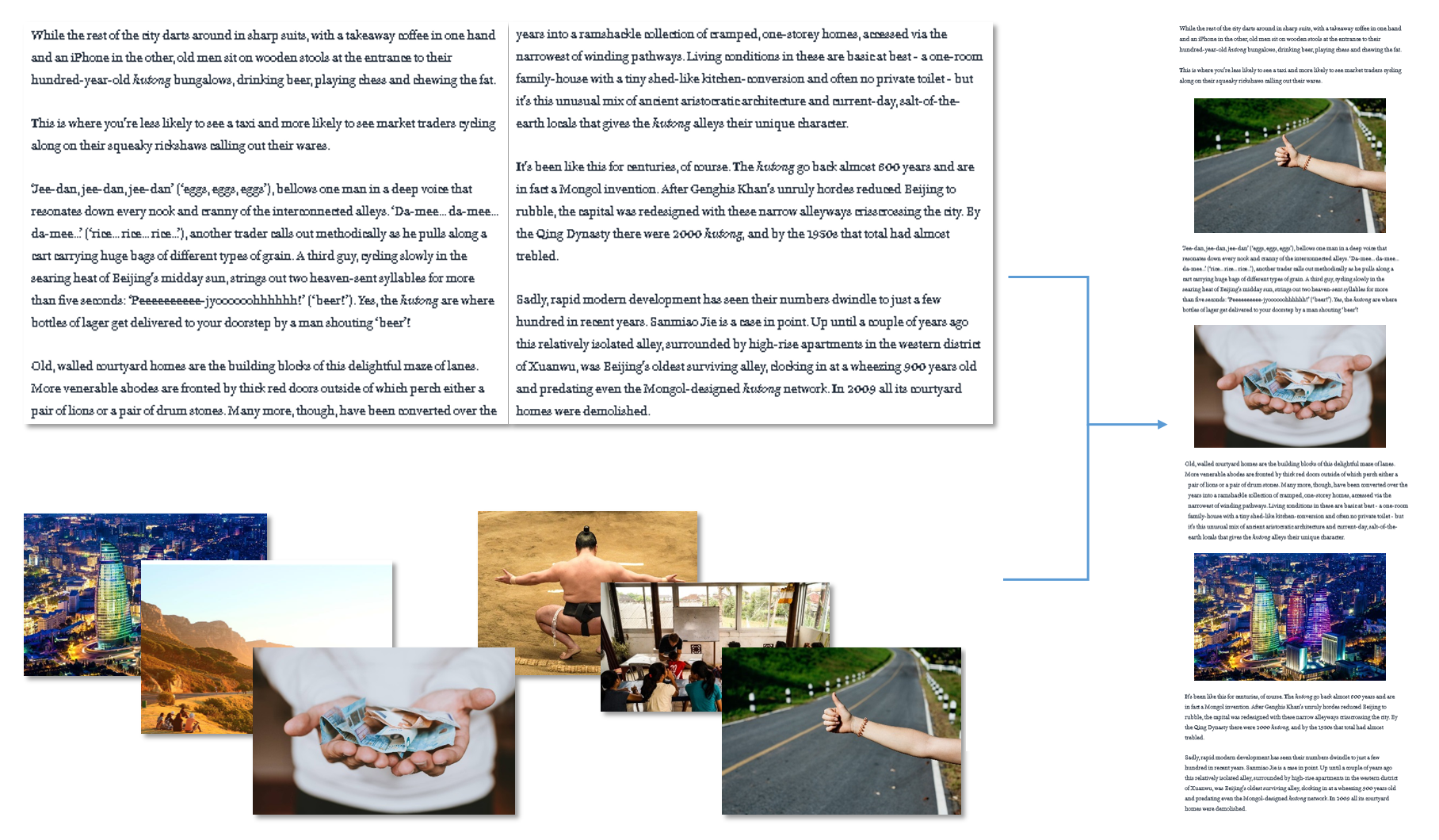}
\caption{The story-and-images alignment problem.}
\label{fig:overview}
\end{center}
\vspace{0.2cm}
\end{figure}

\vspace{0.8em}
\ourheading{Problem Statement} Given a story-like text document and a set of images, the problem is to automatically decide where individual images are placed in the text. 
Figure~\ref{fig:overview} depicts this task. 
The problem comes in different variants: either all images in the
given set need to be placed, or a subset of given cardinality must be selected and aligned with text paragraphs.
Formally, given $n$ paragraphs and $m \le n$ images,
assign these images to a subset of the paragraphs,
such that each paragraph has at most one image.
The variation with image selection assumes that
$m > n$ and requires a budget $b \le n$ for the
number of images to be aligned with the paragraphs.


\vspace{0.8em}
\ourheading{Prior Work and its Inadequacy}
There is ample literature on computer support for multimodal content creation, most notably, on generating image tags and captions.
Closest to our problem is prior work on story illustration \cite{DBLP:journals/tomccap/JoshiWL06, DBLP:conf/kes/SchwarzRCGGL10}, where the task is to select illustrative images from a large pool. However, the task is quite different from ours, making prior approaches inadequate for the setting of this paper. 
First, unlike in general story illustration, we need to consider the text-image alignments jointly for all pieces of a story, rather than making context-free choices one piece at a time. Second, we typically start with a pool of story-specific photos and expect high semantic coherence between each narrative paragraph and the respective image, whereas general story illustration operates with a broad pool of unspecific images that serve many topics.
Third, prior work assumes that each image in the pool has an informative caption or set of tags, by which the selection algorithm computes its choices. Our model does not depend on pre-defined set of tags, but detects image concepts on the fly.

\vspace{0.8em}
Research on Image Tagging may be based on community input, leading to so-called ``social tagging'' \cite{DBLP:journals/sigkdd/GuptaLYH10}, or based on computer-vision methods, called ``visual tagging''. In the latter case, bounding boxes are automatically annotated with image labels, and relationships between objects may also be generated \cite{DBLP:conf/cvpr/RedmonF17, DBLP:conf/eccv/LuKBL16}. Recent works have investigated how to leverage commonsense knowledge as a background asset to further enhance such automatically computed tags \cite{DBLP:conf/wsdm/ChowdhuryTFW18}. Also, deep-learning methods have led to expressive forms of multimodal embeddings, where textual descriptions and images are projected into a joint latent space \cite{DBLP:conf/nips/FromeCSBDRM13, DBLP:conf/bmvc/FaghriFKF18} in order to compute multimodal similarities.

\vspace{0.8em}
In this paper, in addition to manual image tags where available, we harness visual tags from deep neural network based object-detection frameworks and incorporate background commonsense knowledge, as automatic steps to enrich the semantic interpretation of images. This, by itself, does not address the alignment problem, though. The alignment problem is solved by combinatorial optimization.
Our method is experimentally compared to baselines that makes use of multimodal embeddings.


\vspace{0.8em}
\ourheading{Our Approach -- SANDI}
We present a framework that casts the story-images alignment task into a combinatorial optimization problem. The objective function, to be maximized, captures the semantic coherence between each paragraph and the image that is placed there. To this end, we consider a suite of features, most notably, the visual tags associated with an image
(user-defined tags as well as tags from automatic computer-vision tools), 
text embeddings, and also background knowledge in the form of commonsense assertions. The optimization is constrained by the number of images that the story should be enriched with. As a solution algorithm, we devise an integer linear program (ILP) and employ the Gurobi ILP solver for computing the exact optimum. Experiments show that SANDI produces semantically coherent alignments.


\vspace{0.8em}
\ourheading{Contributions} 
To the best of our knowledge, this is the first work to
address the story-images alignment problem. 
Our salient contributions are:\\

\begin{enumerate}
\item We introduce and define the problem of story-images alignment.
\item We analyze two real-world datasets of stories with rich visual illustrations, and derive insights on alignment decisions and quality measures.
\item We devise relevant features, formalize the alignment task as a combinatorial optimization problem, and develop an exact-solution algorithm using integer linear programming.
\item We present experiments that compare our method against baselines that use multimodal embeddings.
\end{enumerate}


\begin{figure*}[t]
	\centering
		\begin{subfigure}[t]{0.5\textwidth}
			\centering
  			 \begin{tabular}{|p{3cm}|p{4.6cm}|}
        		\hline
        		Image & Ground Truth Paragraph\\
                \hline
                \raisebox{-\totalheight}{\includegraphics[height=20mm]{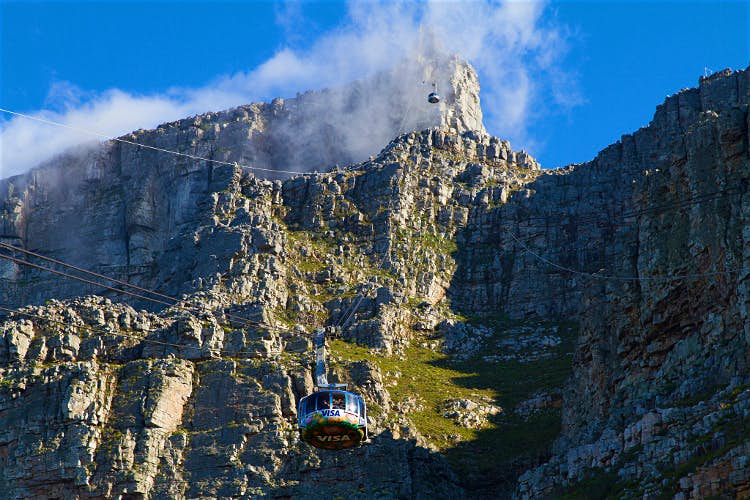}}\vspace{0.5em}
										 & \vspace{0.25em} \ldots Table Mountain Cableway. The revolving car provides 360 degree views as you ascend this mesmerising 60-million-year-old mountain. From the upper cableway station\ldots\\
        		\hline
                \raisebox{-\totalheight}{\includegraphics[height=20.4mm]{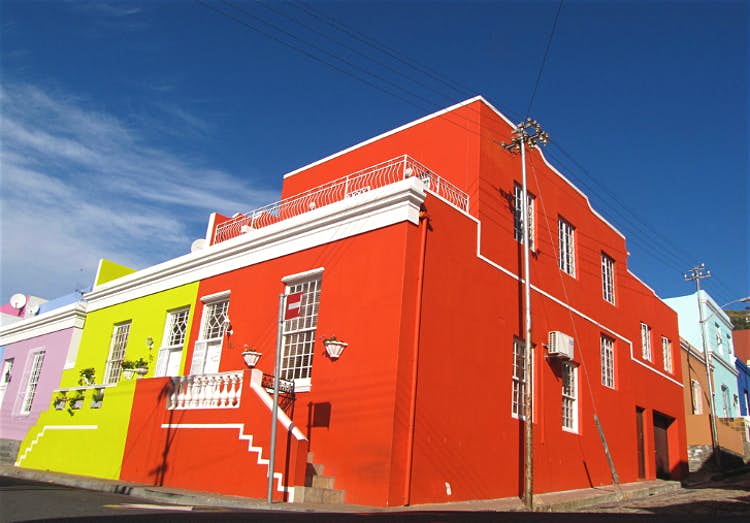}}\vspace{0.5em}
										 & \vspace{0.25em}\ldots On the east flank of the hill is the old Muslim quarter of the Bo-Kaap; have your camera ready to capture images of the photogenic pastel-painted colonial period homes\ldots\\
\hline
   			\end{tabular}
   			\caption{Sample image and corresponding paragraph from Lonely Planet}
        \label{image-para-lonelyPlanet}
		\end{subfigure}%
        ~
		\begin{subfigure}[t]{0.5\textwidth}
			\centering
  			 \begin{tabular}{|p{3cm}|p{4.6cm}|}
        		\hline
        		Image & Ground Truth Paragraph\\
                \hline
                \raisebox{-\totalheight}{\includegraphics[height=20mm]{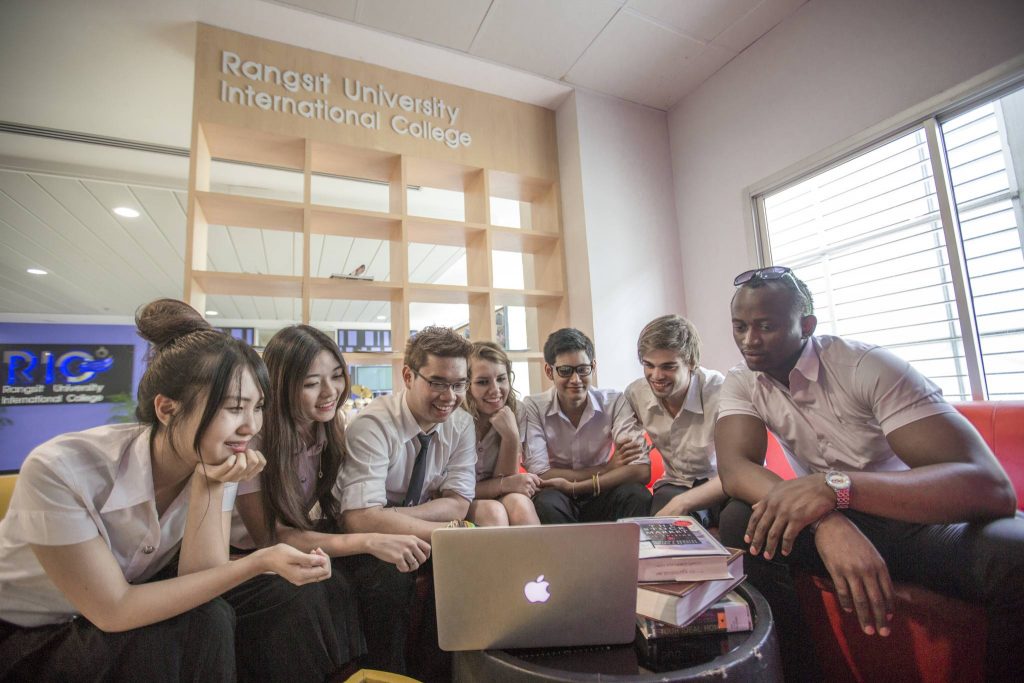}}\vspace{0.6em}
										 & \vspace{0.25em} If you are just looking for some peace and quiet or hanging out with other students...library on campus, a student hangout space in the International College building\ldots.\\
        		\hline
                \raisebox{-\totalheight}{\includegraphics[height=20mm]{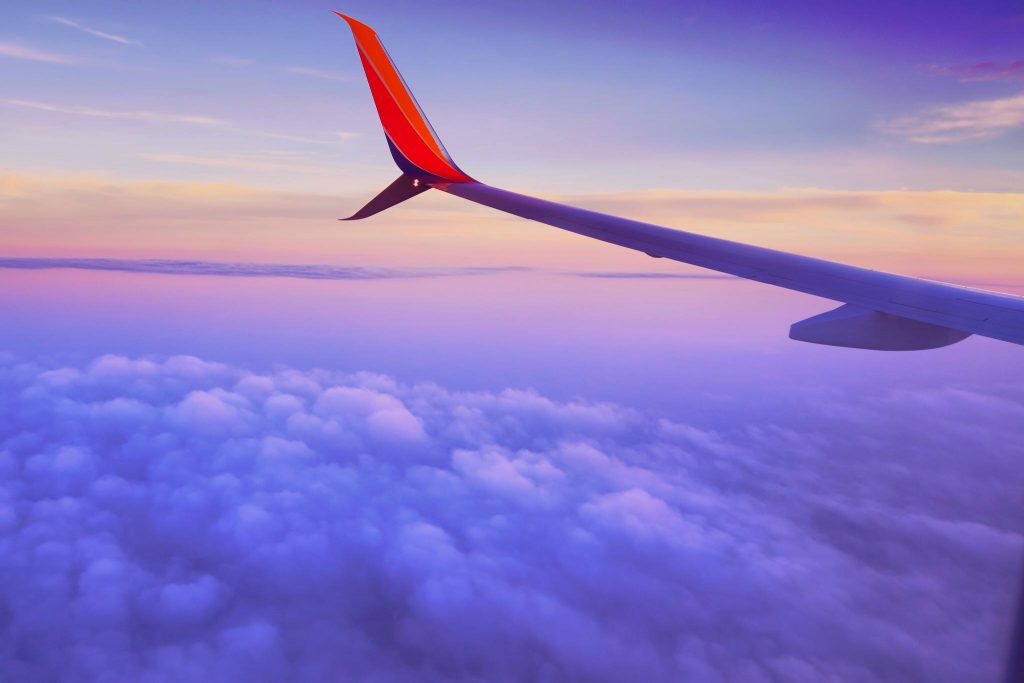}}\vspace{0.7em}
										 & \vspace{0.25em}\ldots I was scared to travel alone. But I quickly realized that there's no need to be afraid. Leaving home and getting out of your comfort zone is an important part of growing up.\ldots\\
\hline
   			\end{tabular}
   			\caption{Sample image and corresponding paragraph from Asia Exchange}
        \label{image-para-studyAbroad}
		\end{subfigure}
\caption{Image-text semantic coherence in datasets.}
\end{figure*}


\section{Related Work}
Existing work that has studied associations between images and text can be categorized into the following areas.


\vspace{0.2cm}
\ourheading{Image Attribute Recognition. }
High level concepts in images lead to better results in Vision-to-Language problems~\cite{DBLP:conf/cvpr/WuSLDH16}.
Identifying image attributes is the starting point toward text-image alignment. To this end, several deep-learning based modern architectures detect concepts in images -- object recognition~\cite{DBLP:conf/nips/HoffmanGTHDGDS14, DBLP:conf/cvpr/RedmonF17, DBLP:conf/nips/RenHGS15}, scene recognition~\cite{DBLP:conf/nips/ZhouLXTO14}, activity recognition~\cite{DBLP:conf/iccv/GkioxariGM15, DBLP:conf/iccv/YaoJKLGF11, DBLP:conf/iccv/ZhaoMY17}. Since all these frameworks work with low level image features like color, texture, gradient etc., noise creep in often leading to incoherent or incorrect detections -- for example, a blue wall could be detected as ``ocean''. While some of the incoherence can be refined using background knowledge~\cite{DBLP:conf/wsdm/ChowdhuryTFW18}, there still exists considerable inaccuracy. We leverage some frameworks from this category in our model to detect visual concepts in images.


\vspace{0.2cm}
\ourheading{Story Illustration} Existing research finds suitable images from a big image collection to illustrate personal stories~\cite{DBLP:journals/tomccap/JoshiWL06} or news posts~\cite{DBLP:conf/kes/SchwarzRCGGL10,DBLP:conf/semco/DelgadoMC10}. Traditionally, images are searched based on textual tags associated with image collections. Occasionally they use visual similarity measures to prune out images very similar to each other. More recent frameworks use deep neural networks to find suitable representative images for a story~\cite{DBLP:conf/cvpr/RaviWMSMK18}. Story Illustration only addresses the problem of image selection, whereas we solve two problems simultaneously:
image selection and image placement -- making a 
joint decision on all pieces of the story.
\cite{DBLP:conf/cvpr/RaviWMSMK18} operates on small stories (5 sentences) with simple content,
and retrieves 1 image per sentence. Our stories are 
much longer texts, the sentences are more complex,
and the stories refer to both general concepts and
named entities.
This makes our problem distinct.
We cannot systematically compare our full-blown model with 
prior works on story illustration alone.


\vspace{0.2cm}
\ourheading{Multimodal Embeddings} 
A popular method of semantically comparing images and text has been to map textual and visual features into a common space of multimodal embeddings~\cite{DBLP:conf/nips/FromeCSBDRM13, DBLP:journals/corr/VendrovKFU15, DBLP:conf/bmvc/FaghriFKF18}. Semantically similar concepts across modalities can then be made to occur nearby in the embedding space. Visual-Semantic-Embeddings (VSE) has been used for generating captions for the whole image~\cite{DBLP:conf/bmvc/FaghriFKF18}, or to associate textual cues to small image regions~\cite{DBLP:conf/cvpr/KarpathyL15} thus aligning text and visuals.
Visual features of image content -- for example color, geometry, aspect-ratio -- have also been used to align image regions to nouns (e.g. ``chair''), attributes (e.g. ``big''), and pronouns (e.g. ``it'') in corresponding explicit textual descriptions~\cite{DBLP:conf/cvpr/KongLBUF14}. However, alignment of small image regions to text snippets play little role in jointly interpreting the correlation between the whole image and a larger body of text. We focus on the latter in this work.


\vspace{0.2cm}
\ourheading{Image-text Comparison} 
Visual similarity of images has been leveraged to associate single words~\cite{DBLP:journals/mta/ZhangQPF17} or commonly occurring phrases~\cite{DBLP:journals/pr/ZhouF15} to a cluster of images. While this is an effective solution for better indexing and retrieval of images, it can hardly be used for contextual text-image alignment. For example, an image with a beach scene may be aligned with either ``relaxed weekend'' or ``this is where I work best'' depending on the context of the full text.

Yet another framework combines visual features from images with verbose image descriptions to find semantically closest paragraphs in the corresponding novels~\cite{DBLP:conf/iccv/ZhuKZSUTF15}, looking at images and paragraphs in isolation. In a similar vein, \cite{DBLP:conf/ism/ChuK17} align images with one semantically closest sentence in the corresponding article for viewing on mobile devices. In contrast, we aim to generate a complete longer multimodal content to be read as a single unit. This calls for distinction between paragraphs and images, and continuity of the story-line.


\ourheading{Image Caption Generation}
Generation of natural language image descriptions is a popular problem at the intersection of computer vision, natural language processing, and artificial intelligence~\cite{DBLP:journals/jair/BernardiCEEEIKM16}.
Alignment of image regions to textual concepts is a prerequisite for generating captions. While most existing frameworks generate factual captions~\cite{DBLP:conf/icml/XuBKCCSZB15, DBLP:conf/accv/TanC16, DBLP:conf/cvpr/LuXPS17}, some of the more recent architectures venture into producing stylized captions~\cite{DBLP:conf/cvpr/GanGHGD17} and stories~\cite{DBLP:conf/iccv/ZhuKZSUTF15,DBLP:conf/cvpr/KrauseJKF17}. Methodologically they are similar to visual-semantic-embedding in that visual and textual features are mapped to the same multimodal embedding space to find semantic correlations. Encoder-decoder LSTM networks are the most common architecture used. An image caption can be considered as a precise focused description of an image without much superfluous or contextual information. For e.g., the caption of an image of the Eiffel Tower would not ideally contain the author's detailed opinion of Paris. However, in a multimodal document, the paragraphs surrounding the image could contain detailed thematic descriptions. We try to capture the thematic indirection between an image and surrounding text, thus making the problem quite different from crisp caption generation.


\ourheading{Commonsense Knowledge for Story Understanding} One of the earliest applications of Commonsense Knowledge to interpret the connection between images and text is a photo agent which automatically annotated images from user's multi-modal (text and image) emails or web pages, while also inferring additional commonsense concepts~\cite{DBLP:conf/ah/LiebermanL02}. Subsequent works used commonsense reasoning to infer causality in stories~\cite{DBLP:conf/commonsense/WilliamsLW17}, especially applicable to question answering. The most commonly used database of commonsense concepts is ConceptNet~\cite{DBLP:conf/aaai/SpeerCH17}. We enhance automatically detected concepts in an image with relevant commonsense assertions. This often helps to capture more context about the image.

\section{Dataset and Problem Analysis}
\label{sec:analysis}

\subsection{Datasets}
\label{dataset}

To the best of our knowledge, there is no experimental dataset for text-image alignment, and existing datasets on image tagging
or image caption generation are not suitable in our setting.
We therefore compile and analyze 
two datasets of blogs from Lonely Planet\footnote{\url{www.lonelyplanet.com/blog}} and Asia Exchange\footnote{\url{www.asiaexchange.org}}.
\squishlist
\item Lonely Planet: 
2178 multimodal articles containing on average 20 paragraphs and 4.5 images per article. Most images are accompanied by captions. Figure~\ref{image-para-lonelyPlanet} shows two image-paragraph pairs from this dataset. Most of the images and come from the author's personal archives and adhere strictly to the content of the article.

\item Asia Exchange: 200 articles about education opportunities in Asia, with an average of 13.5 paragraphs and 4 images per article. The images may be strongly adhering to the content of the article (top image in Figure~\ref{image-para-studyAbroad}), or they may be generic stock images complying with the abstract theme as seen in the bottom image in Figure~\ref{image-para-studyAbroad}). Most images have captions.
\squishend

\ourheading{Text-Image Semantic Coherence} 
To understand the specific nature of this data, we 
had two annotators analyze the given placement of 50 randomly chosen images in articles from the Lonely Planet dataset. The annotators assessed whether the images were specific to the surrounding paragraphs as opposed to merely being relevant for entire articles. The annotators also determined 
to how many paragraphs an image was specifically fitting, and indicated the main reason for the given alignments. For this purpose, we defined 6 possibly overlapping meta-classes: (i) specific man-made entities such as monuments, buildings or paintings, (ii) natural objects such as lakes and mountains, (iii) general nature scenes such as fields or forest, (iv) human activities such as biking or drinking, (v) generic objects such as animals or cars, and (vi) geographic locations such as San Francisco or Rome.


\begin{table}
\begin{tabular}{p{0.2cm}p{4cm}p{2.8cm}}
                                          & Criterion                                                                                                               & \% of images \\ \hline
\multirow{3}{*}{\rotatebox{90}{Relevance}} & \begin{tabular}[c]{@{}l@{}}Placement specific to \\surrounding paragraphs 
\end{tabular} 
& 91\%                  \\
                                          & Relevant text after image   & 86\%\\
                                          & Avg. \#relevant paragraphs  & 1.65\\ [0.5em]
\hline
\multirow{6}{*}{\rotatebox{90}{Main reason}}   
                                          & Natural named objects	& 9\%\\
                                          & Human activities	& 12\%\\
                                          & Generic objects	& 15\%\\
                                          & General nature scenes	& 20\% \\
                                          & Man-made named objects	& 21\%\\
                                          & Geographic locations	& 29\%\\ [0.5em] \hline
\vspace{-1em}                                          
\end{tabular}
\caption{Analysis of image placement for 50 images from Lonely Planet travel blogs.}
\label{tbl:alignment-reasons}
\vspace{-0.6cm}
\end{table}


The outcome of the annotation is shown in Table~\ref{tbl:alignment-reasons}. As one can see, 91\% of the images were indeed more specifically relevant to surrounding text than to the article in general, and 86\% of these were placed before the relevant text. We therefore assume the paragraph following the image as ground truth. 
As to the main reasons for this relevance, we observe quite a mix of reasons, with geographic locations being most important at 29\%, followed by man-made objects at 21\% and general nature scenes at 20\% and so on. 


\begin{figure*}[t]
\captionsetup[subfigure]{labelformat=empty}
\begin{subfigure}[t]{0.235\textwidth}
\includegraphics[width=\textwidth]{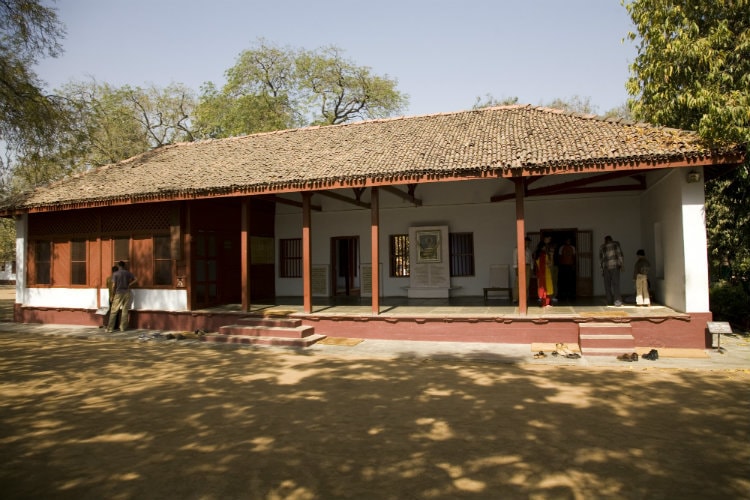}
\caption{\footnotesize{CV: country store, person, bench, lodge outdoor\\MAN: unassuming ashram, Mahatma Ghandi\\BD: Sabarmati Ashram}}
\end{subfigure}\hspace{0.015\textwidth}
\begin{subfigure}[t]{0.235\textwidth}
\includegraphics[width=\textwidth]{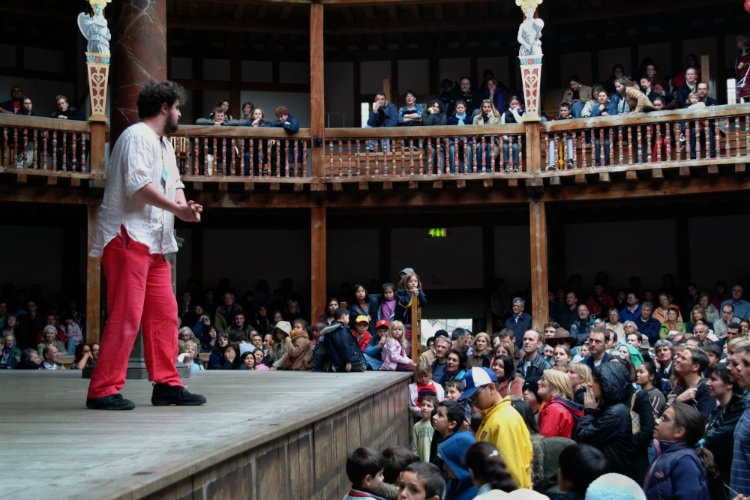}
\caption{\footnotesize{CV: person, sunglasses, stage\\MAN: Globe Theatre, performance, Shakespeare, Spectators\\BD: Shakespeare's Globe\\CSK: show talent, attend concert, entertain audience}}
\end{subfigure}\hspace{0.015\textwidth}
\begin{subfigure}[t]{0.235\textwidth}
\includegraphics[width=\textwidth]{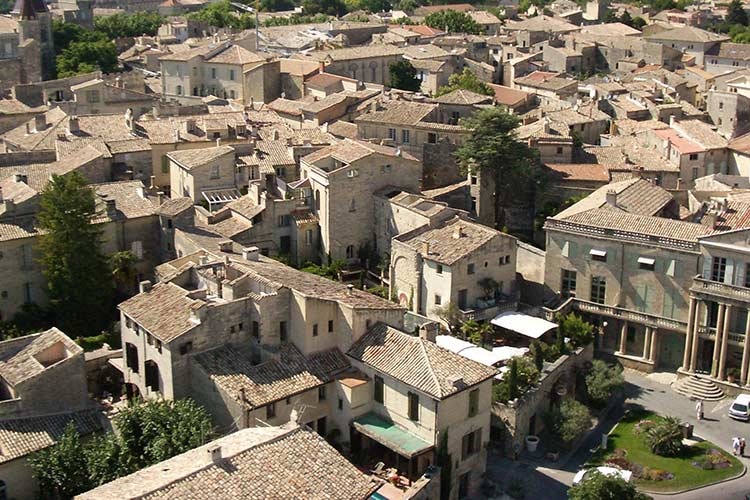}
\caption{\footnotesize{CV: adobe brick, terra cotta, vehicle, table, village\\MAN: tiled rooftops\\BD: uzes languedoc, languedoc roussillon\\CSK: colony, small town}}
\end{subfigure}\hspace{0.015\textwidth}
\begin{subfigure}[t]{0.235\textwidth}
\includegraphics[width=\textwidth]{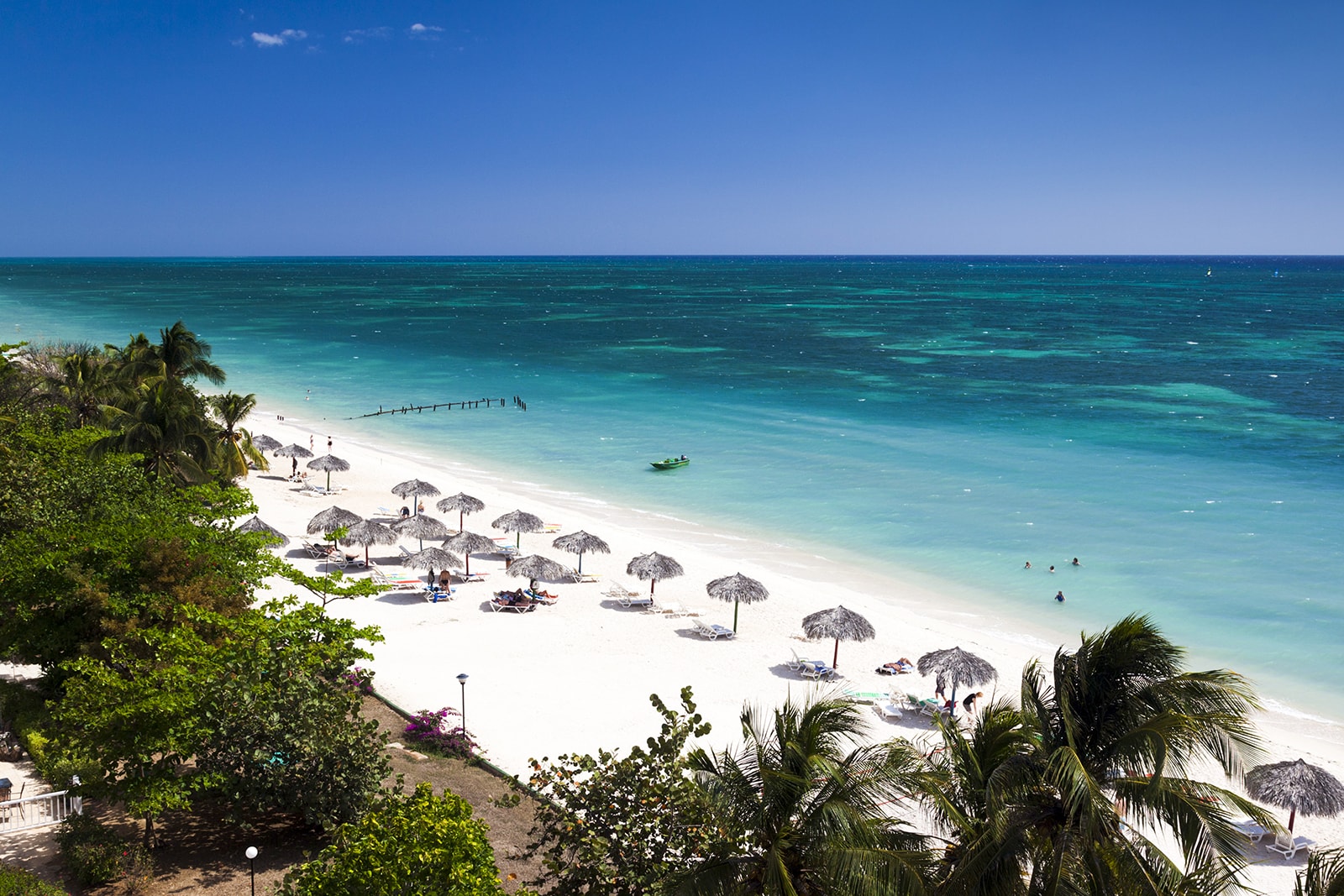}
\caption{\footnotesize{CV: umbrella, beach\\MAN: white sands, Playa Ancon\\BD: ancon cuba, playa ancon\\CSK: sandy shore, vacation}}
\end{subfigure}
\vspace{-0.3cm}
\caption{Characterization of image descriptors: CV adds visual objects/scenes, MAN and BD add location details, CSK adds high-level concepts.}
\label{tagSources}
\end{figure*}


\subsection{Image Descriptors}
\label{features}
Based on the analysis in Table~\ref{tbl:alignment-reasons}, we consider the following kinds of tags for describing images:


\ourheading{Visual Tags (CV)}
State-of-the-art computer-vision methods for object and scene detection
yield visual tags from low-level image features. 
We use three frameworks for this purpose. 
First, 
deep convolutional neural networks based architectures like
LSDA~\cite{DBLP:conf/nips/HoffmanGTHDGDS14} and YOLO~\cite{DBLP:conf/cvpr/RedmonF17}, are used to detect objects like \emph{person}, \emph{frisbee} or \emph{bench}. These models have been trained on ImageNet object classes and denote ``Generic objects'' from Table~\ref{tbl:alignment-reasons}. For stories, general scene descriptors like \emph{restaurant} or \emph{beach} play a major role, too. Therefore, our second
asset is scene detection, specifically from the MIT Scenes Database ~\cite{DBLP:conf/nips/ZhouLXTO14}. Their publicly released pre-trained model ``Places365-CNN'', trained on 365 scene categories with ~5000 images per category, predicts scenes in images with corresponding confidence scores. We pick the most confident scene for each image. These constitute ``General nature scenes'' from Table~\ref{tbl:alignment-reasons}. Thirdly, since stories often abstract away from explicit visual concepts, a framework that incorporates generalizations and abstractions into visual detections~\cite{DBLP:conf/wsdm/ChowdhuryTFW18} is also leveraged. 
For e.g., the concept ``hiking'' is supplemented with the concepts ``walking'' (Hypernym of ``hiking'' from WordNet) and ``fun'' (from ConceptNet~\cite{DBLP:conf/aaai/SpeerCH17} assertion ``hiking, HasProperty, fun'').

\ourheading{User Tags (MAN)} 
Owners of images often have additional knowledge about content and context -- for e.g., 
activities or geographical information (``hiking near Lake Placid''), which, from Table~\ref{tbl:alignment-reasons} play a major role in text-image alignment.
In a down-stream application, users would have the provision to specify tags for their images. For experimental purposes, we use the nouns and adjectives from image captions from our datasets as proxy for user tags.

\ourheading{Big-data Tags (BD)} 
Big data and crowd knowledge allow to infer additional context that may not be visually apparent. 
We utilize the Google reverse image search API\footnote{\url{www.google.com/searchbyimage}} to incorporate such tags. This API allows to search by image, and suggests tags based on those accompanying visually similar images in the vast web image repository\footnote{\label{web}The image descriptors would be made publicly available along with the dataset so that changes in web search results do not affect the reproducibility of our results.}. These tags often depict popular places and entities, such as ``Savarmati Ashram'', or ``Mexico City insect market'', and thus constitute ``Natural names objects'', ``Man-made named objects'', as well as ''Geographic locations'' from Table~\ref{tbl:alignment-reasons}.


\ourheading{Commonsense Knowledge (CSK)} 
Commonsense Knowledge can bridge the gap between visual and textual concepts~\cite{DBLP:conf/akbc/ChowdhuryTW16}. 
We use the following ConceptNet relations to enrich the image tag space: \textit{used for, has property, causes, at location, located near, conceptually related to}. 
As ConceptNet is somewhat noisy, subjective, and diverse, we additionally filter its concepts by \emph{informativeness} for a given image following~\cite{DBLP:conf/naacl/WuG13}. If the top-10 web search results of a CSK concept are semantically similar to the image context (detected image tags), the CSK concept is considered to be \emph{informative} for the image. For example, consider the image context ``\textit{hike, Saturday, waterproof boots}''. CSK derived from ``hike'' are \textit{outdoor activity}, and \textit{fun}. The top-10 Bing search results\textsuperscript{\ref{web}} for the concept \textit{outdoor activity} are semantically similar to the image context. However, those for the term \textit{fun} are semantically varied. Hence, \textit{outdoor activity} is more informative than \textit{fun} for this particular image. Cosine similarity between the mean vectors of the image context and the search results is used as a measure of semantic similarity.

\vspace{0.5em}
\noindent
Figure~\ref{tagSources} shows examples for the different kinds of image tags. 

\section{Model for Story-Images Alignment}
\label{models}
\noindent
Without substantial amounts of labeled
training data, there is no point in
considering machine-learning methods.
Instead, we tackle the task as
a 
Combinatorial Optimization problem 
in an unsupervised way. 

Our \textit{story-images alignment} model constitutes an Integer Linear Program (ILP) 
which jointly optimizes the placement of selected images within an article. 
The main ingredient for this alignment is the pairwise similarity between images and units of text. We consider a paragraph as a text unit.

\newcommand{\srel}{\textit{srel}}

\ourheading{Text-Image Pairwise Similarity} 
Given an image, each of the four kinds of descriptors of Section~\ref{features}
gives rise to a bag of features.
We use these features to compute 
{\em text-image semantic relatedness scores} $\srel(i,t)$ for an image $i$ and a paragraph $t$.%
\begin{equation}
\label{srel}
    srel(i, t) = cosine(\vec{i}, \vec{t})
\end{equation}
where $\vec{i}$ and $\vec{t}$ are the mean word embeddings for the image tags and the paragraph respectively. For images, we use all detected tags. For paragraphs, we consider only the top 50\% of concepts w.r.t. their TF-IDF ranking over the entire dataset. Both paragraph concepts and image tags capture unigrams as well as bigrams.
We use word embeddings from word2vec trained on Google News Corpus.

$srel(i,t)$ scores serve as weights for variables in the ILP.
Note that 
model for
text-image 
similarity 
is
orthogonal to the 
combinatorial problem solved by the ILP.
Cosine distance between concepts (as in Eq.~\ref{srel}) could be easily replaced by other similarity measures over the multimodal
embedding space.

\ourheading{Tasks} Our problem can be divided into 
two
distinct tasks: 
\squishlist 
\item Image Selection -- to select relevant images from an image pool.
\item Image Placement -- to place selected images in the story. 
\squishend
These two components are modelled into one ILP where Image Placement is achieved by maximizing an objective function, while the constraints dictate Image Selection. In the following subsections we discuss two flavors of our model consisting of one or both of the above tasks.


\vspace{-0.1cm}
\subsection{Complete Alignment}
\label{optimalAlignment}

{Complete Alignment} constitutes the problem of aligning {\em all images} in a given image pool with relevant text units of a story. Hence, only Image Placement is applicable.
For a story with $|T|$ text units and an associated image pool with $|I|$ images, the alignment of images $i\in I$ to text units $t\in T$ can be modeled as an ILP with the following definitions:

\vspace{0.5em}
\noindent
{\bf Decision Variables:} The following binary decision variables are introduced:\\
\smallskip
$X_{it} = 1$ if image $i$ should be aligned with text unit $t$, $0$ otherwise.

\vspace{0.5em}
\noindent
{\bf Objective:}
Select image $i$ to be aligned with text unit $t$ such that the semantic relatedness over all text-image pairs is maximized:

\begin{equation}
\label{objective}
    max \bigg[ \sum\limits_{i\in I}\sum\limits_{t\in T}srel(i,t)X_{it}\bigg]
\end{equation}

where $srel(i, t)$ is the text-image semantic relatedness from Eq.~\ref{srel}.

\vspace{0.5em}
\noindent
{\bf Constraints:}

\vspace{-0.6cm}
\begin{multicols}{2}
\begin{equation}
\label{rowConstraint}
\mathlarger\sum\limits_i X_{it} \leq 1 \forall t 
\end{equation}
\break
\begin{equation}
\label{columnConstraint}
\mathlarger\sum\limits_t X_{it} = 1 \forall i 
\end{equation}
\end{multicols}
We make two 
assumptions for text-image alignments:
no paragraph may be aligned with multiple images (\ref{rowConstraint}), and each image is used exactly once in the story (\ref{columnConstraint}). The former is an 
observation from multimodal presentations on the web such as in blog posts 
or brochures. The latter assumption is made based on the nature of our datasets, which are fairly
typical for web contents.
Both are designed as hard constraints that
a solution must satisfy.
In principle, we could relax them into
soft constraints by incorporating
violations as a loss-function penalty into the
objective function.
However, we do not pursue this further,
as typical web contents would indeed
mandate hard constraints.
Note also that the ILP has no
hyper-parameters; so it is completely
unsupervised.


\subsection{Selective Alignment}
\label{selectionAlingment}

{Selective Alignment} is the flavor of the model which {\em selects a 
subset}
of thematically relevant images from a big image pool, and places them within the story. Hence, it constitutes both tasks -- Image Selection and Image Placement.
Along with the constraint in (\ref{rowConstraint}), Image Selection entails the following additional constraints:

\vspace{-0.6cm}
\begin{multicols}{2}
\begin{equation}
\label{columnConstraint_selection}
\mathlarger\sum\limits_t X_{it} \leq 1 \forall i 
\end{equation}
\break
\begin{equation}
\mathlarger\sum\limits_i\mathlarger\sum\limits_t X_{it} = b
\end{equation}
\end{multicols}

where $b$ is the budget for the number of images for the story.
$b$ may be trivially defined as the number of paragraphs in the story, following our assumption that each paragraph may be associated with a maximum of one image.
(\ref{columnConstraint_selection}) is an adjustment to (\ref{columnConstraint}) which implies that not all images from the image pool need to be aligned with the story.
The objective function from (\ref{objective}) rewards the
selection of best fitting images from the image pool.

\section{Quality Measures}
\label{qualityMeasures}

In this section we define metrics for automatic evaluation of text-image alignment models. The two tasks involved -- Image Selection and Image Placement -- call for separate evaluation metrics as discussed below.

\vspace{-0.2em}
\subsection{Image Selection}
\label{image_selection_metrics}

Representative images for a story are selected from a big pool of images. There are multiple conceptually similar images in our image pool since they have been gathered from blogs of the domain ``travel''. 
Hence evaluating the results on strict precision (based on exact matches between selected and ground-truth images) does not necessarily assess true quality.
We therefore define a relaxed precision metric (based on semantic similarity) in addition to the strict metric. Given a set of selected images $I$ and the set of ground truth images $J$, where $|I| = |J|$, the precision metrics are:

\begin{equation}
\label{semPrec}
    Relaxed Precision = \frac{\sum\limits_{\substack{i\in I}} \max\limits_{\substack{j\in J}}(cosine(\vec{i}, \vec{j}))}{|I|}
\end{equation}%
\begin{equation}
\label{strictPrec}
    Strict Precision = \frac{|I\cap J|}{|I|}
\end{equation}




\subsection{Image Placement}
\label{image_placement_metrics}

For each image in a multimodal story, the ground truth (GT) paragraph is assumed to be the one following the image in our datasets. 
To evaluate the quality of SANDI's text-image alignments, we compare the GT paragraph and the paragraph assigned to the image by SANDI (henceforth referred to as ``aligned paragraph'').
We propose the following metrics for evaluating the quality of alignments:

\vspace{0.2cm}
\ourheading{BLEU and ROUGE} BLEU and ROUGE are classic n-gram-overlap-based metrics for evaluating machine translation and text summarization. Although known to be limited insofar as they do not recognize synonyms and semantically equivalent formulations, they are in widespread use. 
We consider them as basic measures of concept overlap between GT and aligned paragraphs.

\vspace{0.2cm}
\ourheading{Semantic Similarity}
To alleviate the shortcoming of requiring exact matches, we consider a metric based on embedding similarity. We compute the similarity between two text units $t_i$ and $t_j$ by the average similarity of their word embeddings, considering all
unigrams and bigrams as words.

\begin{equation}
SemSim(t_i,t_j) = cosine(\vec{t_i}, \vec{t_j})
\label{semRel}
\vspace{0.2cm}
\end{equation}

where $\vec{x}$ is the mean vector of words in $x$.
For this calculation, we drop uninformative words
by keeping only the top 50\% with regard to their TF-IDF weights over the whole dataset.

\vspace{0.2cm}
\ourheading{Average Rank of Aligned Paragraph} We associate each paragraph in the story with a ranked list of all the paragraphs on the basis of semantic similarity (Eq.~\ref{semRel}), where rank 1 is the paragraph itself. Our goal is to produce alignments ranked higher with the GT paragraph. The average rank of alignments produced by a method is computed as follows:

\begin{equation}
\mathit{ParaRank} = 1 - \bigg[ \bigg(\frac{\sum\limits_{t\in T'} rank(t)}{|I|} -1\bigg)\bigg/ \bigg(|T| - 1 \bigg)\bigg]
\vspace{0.2cm}
\end{equation}

where $|I|$ is the number of images and $|T|$ is the number of paragraphs in the article. ${T}'\subset {T}$ is the set of paragraphs aligned to images.
Scores are normalized between 0 and 1; 1 being the perfect alignment and 0 being the worst alignment.

\vspace{0.2cm}
\ourheading{Order Preservation} Most stories 
follow a storyline. Images placed at meaningful spots within the story would ideally adhere to this sequence. Hence the measure of pairwise ordering provides a sense of respecting the storyline. Lets define order preserving image pairs as:
$P = \{(i,i'): i,i'\in I, i\neq i', i' \text{ follows } i$ in both GT and SANDI alignments$\}$, where $I$ is the set of images in the story. The measure can be defined as number of order preserving image pairs normalized by the total number of GT ordered image pairs.

\begin{equation}
\mathit{OrderPreserve} = \frac{|P|}{ (|I| (|I| - 1)/2)}
\vspace{0.2cm}
\end{equation}

\begin{table*}[t]
\begin{minipage}{0.5\linewidth}
\centering
\caption{Complete Alignment on the Lonely Planet dataset.}
\begin{tabular}{l|lllll}
\hline
          & \rotatebox{60}{BLEU} & \rotatebox{60}{ROUGE} & \rotatebox{60}{SemSim} & \rotatebox{60}{ParaRank} & \rotatebox{60}{\makecell{Order\\ Preserve}}\\ \hline
Random & 3.1 & 6.9 & 75.1 & 50.0 & 50.0\\
VSE++ \cite{DBLP:conf/bmvc/FaghriFKF18}      & 11.0 & 9.5   & 84.6   & 59.1    & 55.2\\
VSE++ ILP     & 12.56 & 11.23 & 83.98 & 58.08 & 47.93 \\[0.2em]
\hdashline
SANDI-CV  & 18.2 & 17.6  & 86.3   & 63.7    & 54.5 \\
SANDI-MAN & \textbf{45.6} & \textbf{44.5}  & \textbf{89.8}   & 72.5    & \textbf{77.4} \\
SANDI-BD  & 26.6 & 25.1  & 84.7   & 61.3    & 61.2 \\
SANDI{\LARGE $\ast$}    & 44.3 & 42.9  & 89.7   & \textbf{73.2}    & 76.3 \\
\hline
\end{tabular}
\label{evaluation}
\end{minipage}%
\begin{minipage}{0.5\linewidth}
\centering
\caption{Complete Alignment on the Asia Exchange dataset.}
\begin{tabular}{l|lllll}
\hline
          & \rotatebox{60}{BLEU} & \rotatebox{60}{ROUGE} & \rotatebox{60}{SemSim} & \rotatebox{60}{ParaRank} & \rotatebox{60}{\makecell{Order\\Preserve}}\\%
          \hline
Random & 6.8 & 8.9 & 70.8 & 50.0 & 50.0\\
VSE++ \cite{DBLP:conf/bmvc/FaghriFKF18}      & 19.4 & 17.7   & 85.7   & 51.9    & 48.0  \\
VSE++ ILP    & 23.5 & 20.11 & 85.98 & 52.55 & 46.13\\[0.2em]
\hdashline
SANDI-CV  & 21.5 & 20.6  & 87.8   & 58.4    & 52.0  \\
SANDI-MAN & \textbf{35.2} & \textbf{32.2}  & 89.2   & 61.5    & 61.5  \\
SANDI-BD  & 24.1 & 22.3  & 86.7   & 56.0    & 53.6  \\
SANDI{\LARGE $\ast$}    & 33.4 & 31.5  & \textbf{89.7}   & \textbf{62.4}    & \textbf{62.5}  \\
\hline
\end{tabular}
\label{evaluationStudyAbroad}
\end{minipage}
\end{table*}

\section{Experiments and Results}

We evaluate the two flavors of SANDI -- Complete Alignment and Selective Alignment -- based on the quality measures described in Section~\ref{qualityMeasures}.

\subsection{Setup}
\label{exp-setup}

\ourheading{Tools}
Deep convolutional neural network based architectures similar to LSDA~\cite{DBLP:conf/nips/HoffmanGTHDGDS14}, YOLO~\cite{DBLP:conf/cvpr/RedmonF17}, VISIR~\cite{DBLP:conf/wsdm/ChowdhuryTFW18} 
and Places-CNN~\cite{DBLP:conf/nips/ZhouLXTO14} are used as sources of \emph{Visual tags}. Google reverse image search tag suggestions are used as \emph{Big-data tags}. We use the Gurobi Optimizer for solving the ILP. 
A Word2Vec~\cite{DBLP:conf/nips/MikolovSCCD13} model trained on the Google News Corpus encompasses a large cross-section of domains, and hence is a well-suited source of word embeddings for our purposes.

\vspace{0.5em}
\ourheading{SANDI Variants}
The variants of our text-image alignment model are based on the use of image descriptors from Section~\ref{features}.%
\squishlist
\item SANDI-CV, SANDI-MAN, and SANDI-BD use CV, MAN, and BD tags as image descriptors respectively.
\item SANDI{\LARGE $\ast$} combines tags from all sources.
\item +CSK: With this setup we study the role of commonsense knowledge as a bridge between visual features and textual features.%
\squishend

\ourheading{Alignment sensitivity}
The degree to which alignments are specific to certain paragraphs varies from article to article. For some articles, alignments have little specificity, for instance, when the whole article talks about a hiking trip, and images generically show forests and mountains. We measure alignment sensitivity of articles by 
comparing the semantic relatedness of an image to its ground-truth paragraph
against all other paragraphs in the same article.
We use the cosine similarity between the image's vector of MAN tags and 
the text vectors, for this purpose.
The alignment sensitivity of an article then is the average of these similarity scores
over all its images.
We restrict our experiments to the top-100 most alignment-sensitive articles in each dataset.

\subsection{Complete Alignment}
\label{completeAlignment}
In this section we evaluate our Complete Alignment model (defined in Section~\ref{optimalAlignment}), which places \textit{all} images from a given image pool within a story.

\ourheading{Baselines}
To the best of our knowledge, there is no existing work on story-image alignment in the literature. Hence we modify methods on joint visual-semantic-embeddings (VSE) \cite{DBLP:journals/corr/KirosSZ14, DBLP:conf/bmvc/FaghriFKF18} to serve as baselines. Our implementation of VSE is similar to ~\cite{DBLP:conf/bmvc/FaghriFKF18}, henceforth referred to as VSE++.
We compare SANDI with the following baselines:
\squishlist
\item Random: a simple baseline with random image-text alignments.
\item VSE++ Greedy or simply VSE++: for a given image, VSE++ is adapted to produce a ranked list of paragraphs from the corresponding story. The best ranked paragraph is considered as an alignment, with a greedy constraint that one paragraph can be aligned to at most one image.
\item VSE++ ILP: using cosine similarity scores between image and paragraph from the joint embedding space, we solve an ILP for the alignment with the same constraints as that of SANDI.
\squishend

Since there are no existing story-image alignment datasets, VSE++ has been trained on the MSCOCO captions dataset \cite{DBLP:conf/eccv/LinMBHPRDZ14}, which contains 330K images with 5 captions per image.




\vspace{0.5em}
\ourheading{Evaluation}
Tables~\ref{evaluation} and \ref{evaluationStudyAbroad} show the performance of SANDI variants across the different evaluation metrics (from Section~\ref{image_placement_metrics}) on the Lonely Planet and Asia Exchange datasets respectively. 
On both datasets, SANDI outperforms VSE++, especially in terms of paragraph rank (+14.1\%/+10.5\%) and order preservation (+11.1\%/+14.5\%). While VSE++ looks at each image in isolation, SANDI captures context better by considering all text units of the article and all images from the corresponding album at once in a constrained optimization problem. 
VSE++ ILP, although closer to SANDI in methodology, does not outperform SANDI.
The success of SANDI can also be attributed to the fact that it is less tied to a particular type of images and text, relying only on word2vec embeddings that are trained on a much larger corpus than MSCOCO.

On both datasets, SANDI-MAN is the single best configuration, while the combination, SANDI{\LARGE $\ast$} marginally outperforms it on the Asia Exchange dataset. The similarity of scores across both datasets highlights the robustness of the SANDI approach.

\vspace{0.5em}
\ourheading{Role of Commonsense Knowledge}
While in alignment-sensitive articles the connections between paragraphs and images are often immediate, this is less the case for articles with low alignment sensitivity.   Table~\ref{evaluation-csk} shows the impact of adding common sense knowledge on the 100 least alignment sensitive articles from the Lonely Planet dataset. As one can see, adding CSK tags leads to a minor improvement in terms of semantic similarity (+0.1/+0.4\%), although the improvement is too small to argue that CSK is an important ingredient in text-image alignments.

\begin{table}[t]
\centering
\caption{Role of Commonsense Knowledge.}
\begin{tabular}{p{2cm}|p{2cm}p{1.5cm}p{1.5cm}}
\hline
 &				&{Standard} &{+CSK}\\
\hline
\multirow{2}{*}{SANDI-CV}							& \emph{SemSim} & 86.2 & \textbf{86.3}\\
									& \emph{ParaRank} & 59.9 & 59.7 \\[0.2em]
\hline
\multirow{2}{*}{SANDI-MAN}							& \emph{SemSim} & 85.1 & \textbf{85.5}\\
									& \emph{ParaRank} & 53.8 & \textbf{55.0} \\[0.1em]
\hline
\end{tabular}
\label{evaluation-csk}
\end{table}

\begin{table*}[t]
\begin{minipage}{0.5\linewidth}
\centering
\caption{Image Selection on the Lonely Planet dataset.}
\vspace{-0.2cm}
\begin{tabular}{l|l|llll}
\hline
                  {\makecell{Tag Space}} & Precision & {Random} & {NN} & {VSE++} & {SANDI}\\
\hline
\multirow{2}{*}{CV} & \small{$Strict$}            & 0.4   & 2.0   & 1.14 & \textbf{4.18}\\
                    & \small{$Relaxed$}    & 42.16 & 52.68 & 29.83 & \textbf{53.54}\\
\hline
\multirow{2}{*}{MAN} & \small{$Strict$}            & 0.4   & 3.95   & - & \textbf{14.57}\\
                    & \small{$Relaxed$}    & 37.14 & 42.73 &  & \textbf{49.65}\\
\hline
\multirow{2}{*}{BD} & \small{$Strict$}            & 0.4   & 1.75   & - & \textbf{2.71}\\
                    & \small{$Relaxed$}    & 32.59 & 37.94 &  & \textbf{38.86}\\
\hline
\multirow{2}{*}{\Huge{$\ast$}} & \small{$Strict$}            & 0.4   & 4.8   & - & \textbf{11.28}\\
                    & \small{$relaxed$}    & 43.84 & 50.06 &  & \textbf{54.34}\\
\hline
\end{tabular}
\label{storyIllustration-lonelyPlanet}
\end{minipage}%
\begin{minipage}{0.5\linewidth}
\centering
\caption{Image Selection on the Asia Exchange dataset.}
\vspace{-0.2cm}
\begin{tabular}{l|l|llll}
\hline
                  {\makecell{Tag Space}} & Precision & {Random} & {NN} & {VSE++} & {SANDI}\\
\hline
\multirow{2}{*}{CV} & \small{$Strict$}            & 0.45   & 0.65   & 0.44 & \textbf{0.79}\\
                    & \small{$Relaxed$}    & 55.0 & \textbf{57.64} & 30.05 & 57.2\\
\hline
\multirow{2}{*}{MAN} & \small{$Strict$}            & 0.45   &  0.78  & - & \textbf{3.42}\\
                    & \small{$Relaxed$}    & 40.24 & 52.0 &  & \textbf{52.87}\\
\hline
\multirow{2}{*}{BD} & \small{$Strict$}            & 0.45   &  0.82  & - & \textbf{0.87}\\
                    & \small{$Relaxed$}    & 31.12 & \textbf{33.27}  &  & 33.25\\
\hline
\multirow{2}{*}{\Huge{$\ast$}} & \small{$Strict$}            & 0.45   &  1.04 & - & \textbf{1.7}\\
                    & \small{$relaxed$}    & 55.68 & 58.1 &  & \textbf{58.2}\\
\hline
\end{tabular}
\label{storyIllustration-asiaExchange}
\end{minipage}%
\vspace{-0.3cm}
\end{table*}

\begin{table}[b]
\centering
\caption{Selective Alignment on the Lonely Planet dataset.}
\begin{tabular}{l|llll}
\hline
          & \rotatebox{60}{BLEU} & \rotatebox{60}{ROUGE} & \rotatebox{60}{SemSim} & \rotatebox{60}{ParaRank} \\ \hline
Random & 0.31 & 0.26 & 69.18 &  48.16 \\
VSE++ \cite{DBLP:conf/bmvc/FaghriFKF18}      & 1.04 & 0.8  &  79.18  &   53.09  \\
VSE++ ILP    & 1.23 & 1.03 & 79.04 & 53.96 \\[0.2em]
\hdashline
SANDI-CV  & 1.70 & 1.60  &  83.76  &  61.69   \\
SANDI-MAN & \textbf{8.82} & \textbf{7.40}  &  82.95  &  66.83   \\
SANDI-BD  & 1.77 & 1.69 &  \textbf{84.66}  &  \textbf{76.18}    \\
SANDI{\LARGE $\ast$}    & 6.82 & 6.57 &  84.50 &   75.84    \\ 
\hline
\end{tabular}
\label{evaluation-selectiveAlignment-lonelyPlanet}
\end{table}

\subsection{Selective Alignment}
This variation of our model, as defined in Section~\ref{selectionAlingment}, solves two problems simultaneously -- selection of representative images for the story from a big pool of images, and placement of the selected images within the story. The former sub-problem relates to the topic of ``Story Illustration''~\cite{DBLP:journals/tomccap/JoshiWL06, DBLP:conf/kes/SchwarzRCGGL10, DBLP:conf/semco/DelgadoMC10, DBLP:conf/cvpr/RaviWMSMK18}, but work along these lines
has focused on very short texts with simple content
(in contrast to the long and content-rich stories in our datasets).\\

\vspace{-0.2em}
\noindent
\textbf{\large6.3.1 \hspace{0.6em} Image Selection}

\ourheading{Setup} In addition to the setup described in Section~\ref{exp-setup}, following are the requirements for this task:
\squishlist
\item Image pool -- We pool images from all stories in the slice of the dataset we use in our experiments. 
Stories from a particular domain -- for e.g. travel blogs from Lonely Planet -- are largely quite similar. 
This entails that images in the pool may also be very similar in content -- for e.g., stories on \textit{hiking} contain images with similar tags like \textit{mountain, person, backpack}. 
\item Image budget -- For each story, the number of images in the ground truth is considered as the image budget $b$ for Image Selection (Equation~\ref{selectionAlingment}).
\squishend



\ourheading{Baselines}
We compare SANDI with the following baselines:
\squishlist
\item Random: a baseline of randomly selected images from the pool.
\item NN: a selection of nearest neighbors from a common embedding space of images and paragraphs.
Images are represented as centroid vectors of their tags,
and paragraphs are represented as centroid vectors of their distinctive concepts. The basic vectors are obtained from Word2Vec trained on Google News Corpus.
\item VSE++: state-of-the-art on joint visual-textual embeddings; the method presented in \cite{DBLP:conf/bmvc/FaghriFKF18} is
adapted to retrieve the top-$b$ images for a story. 
\squishend

\vspace{-0.2em}
\ourheading{Evaluation}
We evaluate Image Selection by the measures in Section~\ref{image_selection_metrics}.
Table~\ref{storyIllustration-lonelyPlanet} and Table~\ref{storyIllustration-asiaExchange} show
the results for Story Illustration, that is, image selection, for SANDI and the baselines. For the Lonely Planet dataset (Table~\ref{storyIllustration-lonelyPlanet}), 
a pool of 500 images was used. We study the effects of a bigger images pool (1000 images) in our experiments with the Asia Exchange dataset (Table~\ref{storyIllustration-asiaExchange}). As expected, average strict precision (exact matches with ground truth) drops. Recall from Section~\ref{dataset} that the Asia Exchange dataset often has stock images for general illustration rather than only story-specific images. Hence the average relaxed precision on image selection is higher.
The nearest-neighbor baseline (NN) and SANDI, both use Word2Vec embeddings for text-image similarity. SANDI's better scores are attributed to the joint optimization over the entire story, as opposed to greedy selection in case of NN. VSE++ uses a joint text-image embeddings space for similarity scores.
The results in the tables clearly show SANDI's advantages over the baselines.

Our evaluation metric $Relaxed Precision$ (Eq.~\ref{semPrec}) factors in the
semantic similarity between images which in turn depends on the image descriptors (Section~\ref{features}). Hence we compute results on the different image tag spaces, where `{\LARGE $\ast$}' refers to the combination of CV, MAN, and BD. The baseline VSE++ however, operates only on 
visual features; hence we report its performance only for CV tags.

\setfillcolor{white}
\setbordercolor{green}
\begin{figure*}[t]
\setlength\tabcolsep{2pt}
\begin{tabular}{lllllllll}
\hline
    \rotatebox{90}{Story} & 
    \multicolumn{8}{l}{\makecell[l]{\small{Clatter into Lisbon's steep, tight-packed Alfama aboard a classic yellow tram...England. Ride a regular bus for a squeezed-in-with-the-natives view of}\\
    \small{the metropolis...Venice, Italy...opting for a public vaporetto (water taxi) instead of a private punt...Hungary. Trundle alongside the Danube, with views}\\
    \small{up to the spires and turrets of Castle Hill...Istanbul, Turkey...Travel between Europe and Asia...Ferries crossing the Bosphorus strait...Sail at sunset...} \\
    \small{Monte Carlo's electric-powered ferry boats...The `Coast Tram' skirts Belgium's North Sea shoreline...Pretty but pricey Geneva...travel on buses,} \\
    \small{trams and taxi-boats...Liverpool, England...Hop aboard Europe's oldest ferry service...just try to stop yourself bursting into song.}}} \vspace{0.1cm}\\
    \hline
    \rotatebox{90}{GT} & 
    \includegraphics[width=2.0cm]{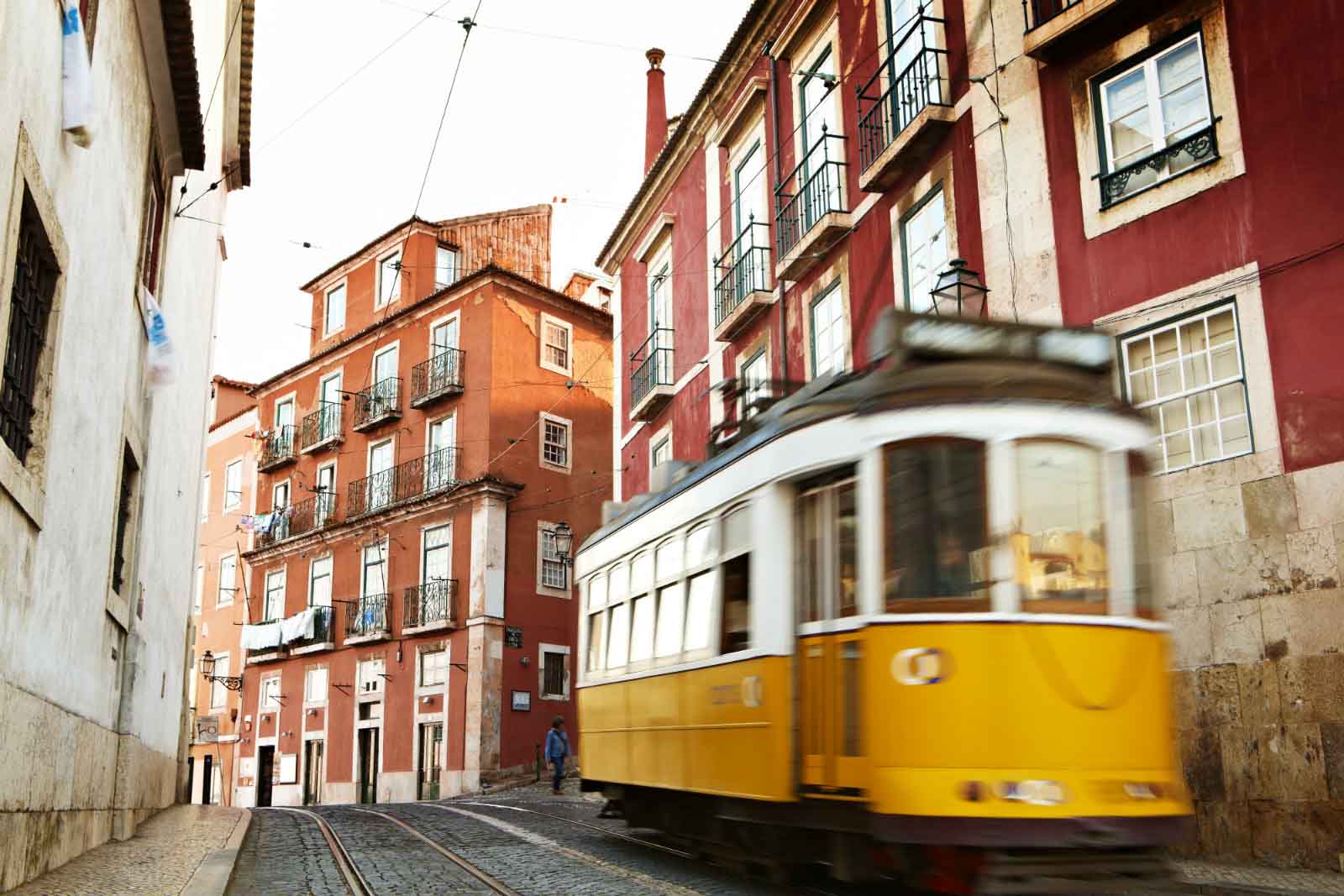} & 
    \includegraphics[width=2.0cm]{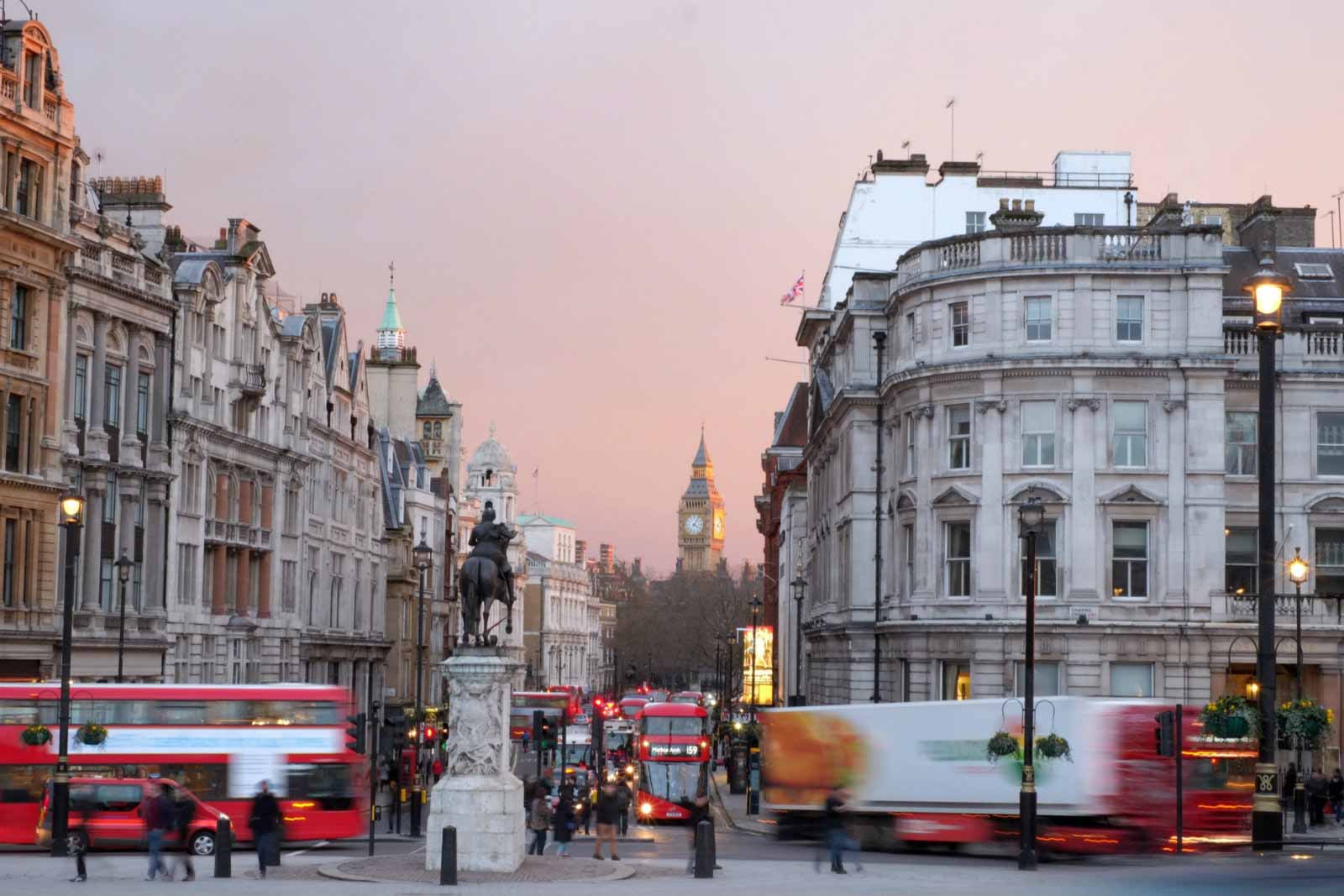} & 
    \includegraphics[width=2.0cm]{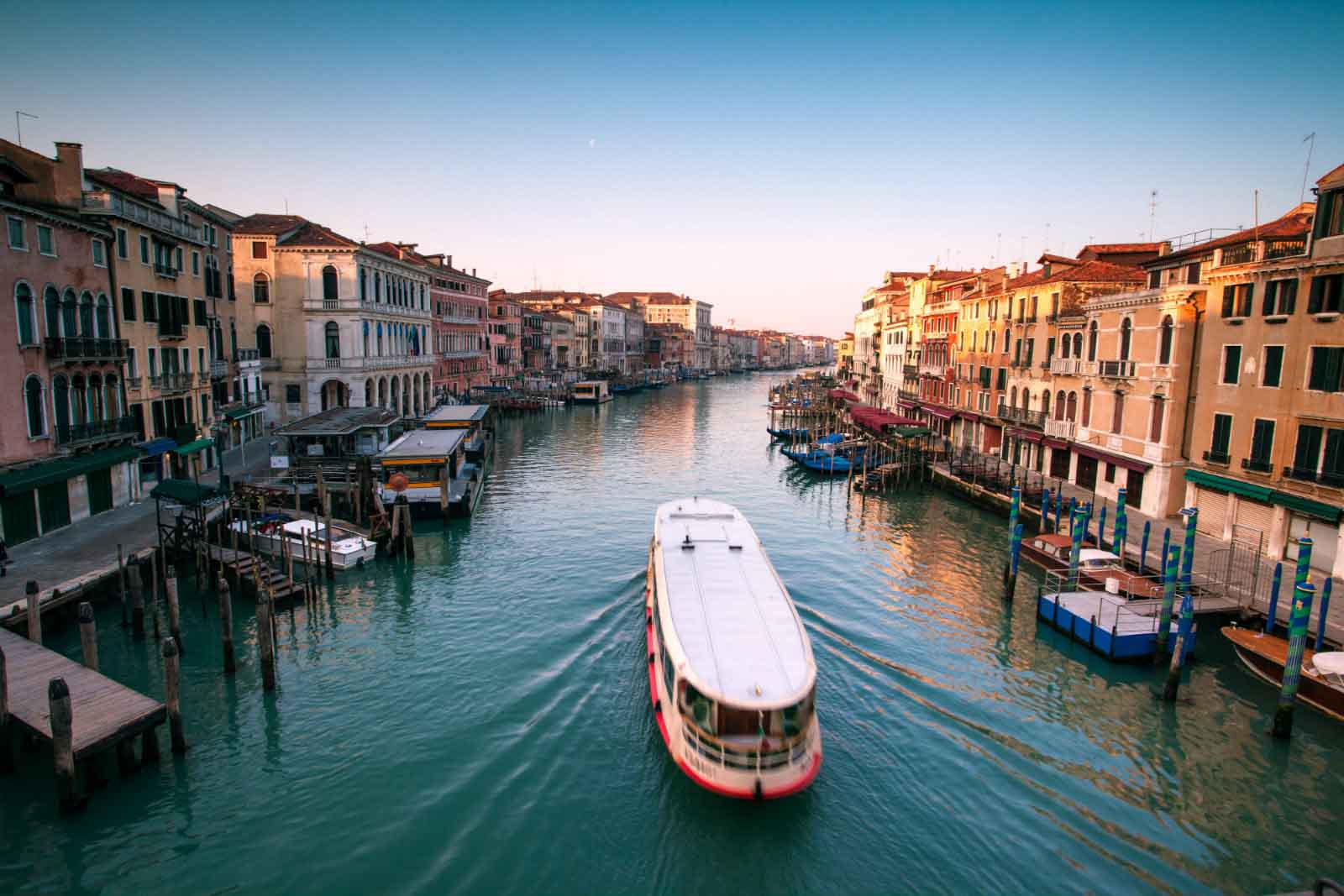} & 
    \includegraphics[width=2.0cm]{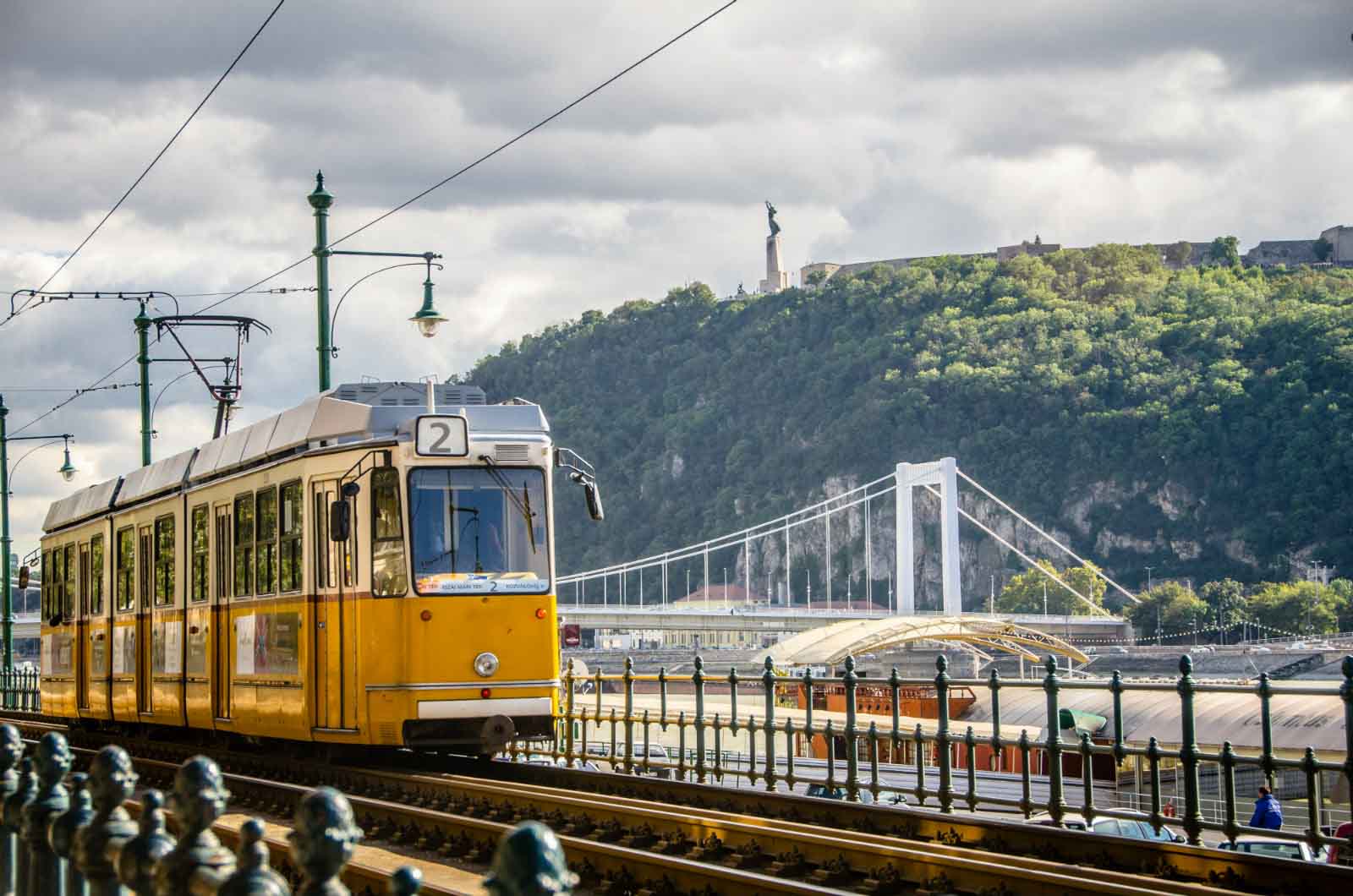} & 
    \includegraphics[width=2.0cm]{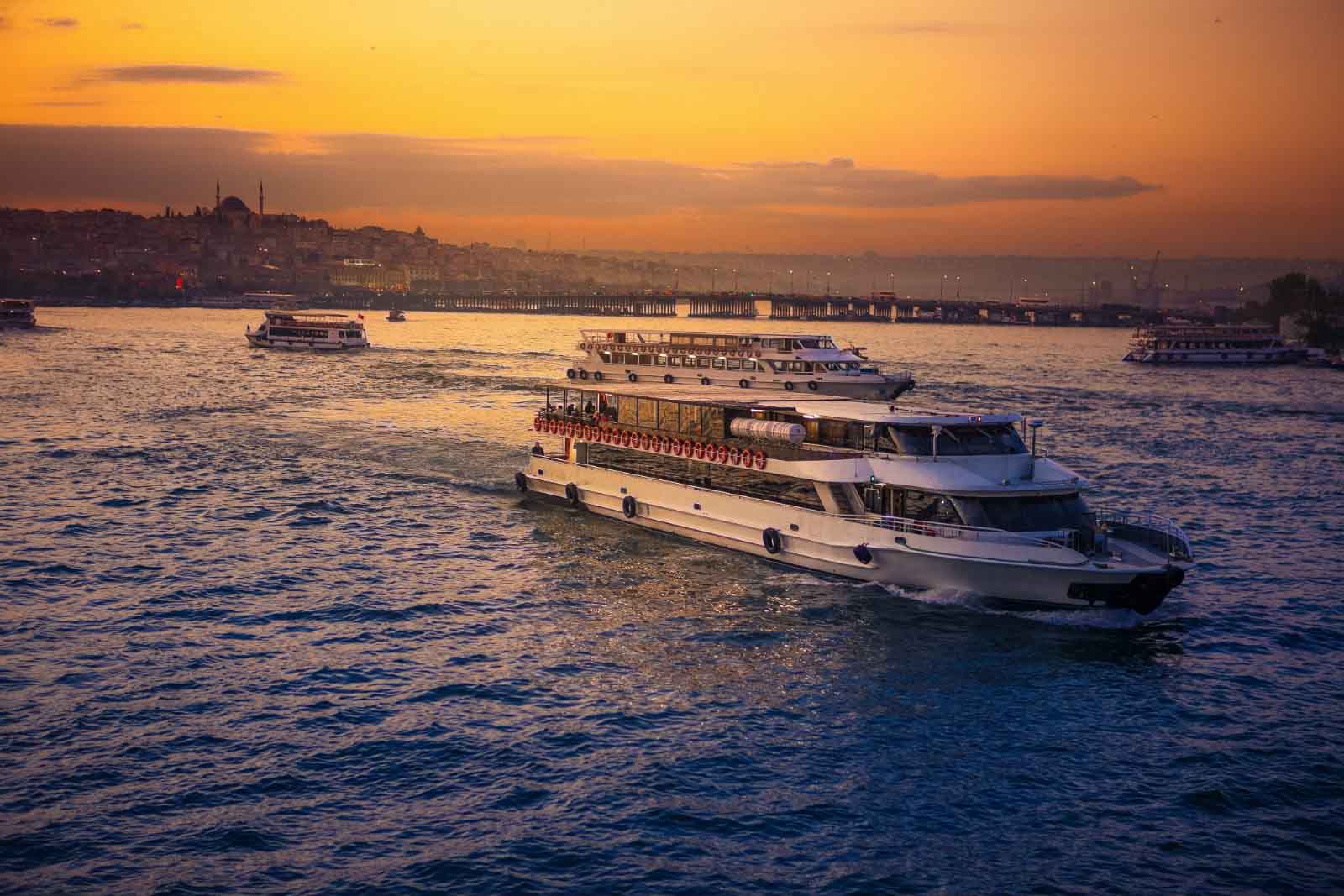} & 
    \includegraphics[width=2.0cm]{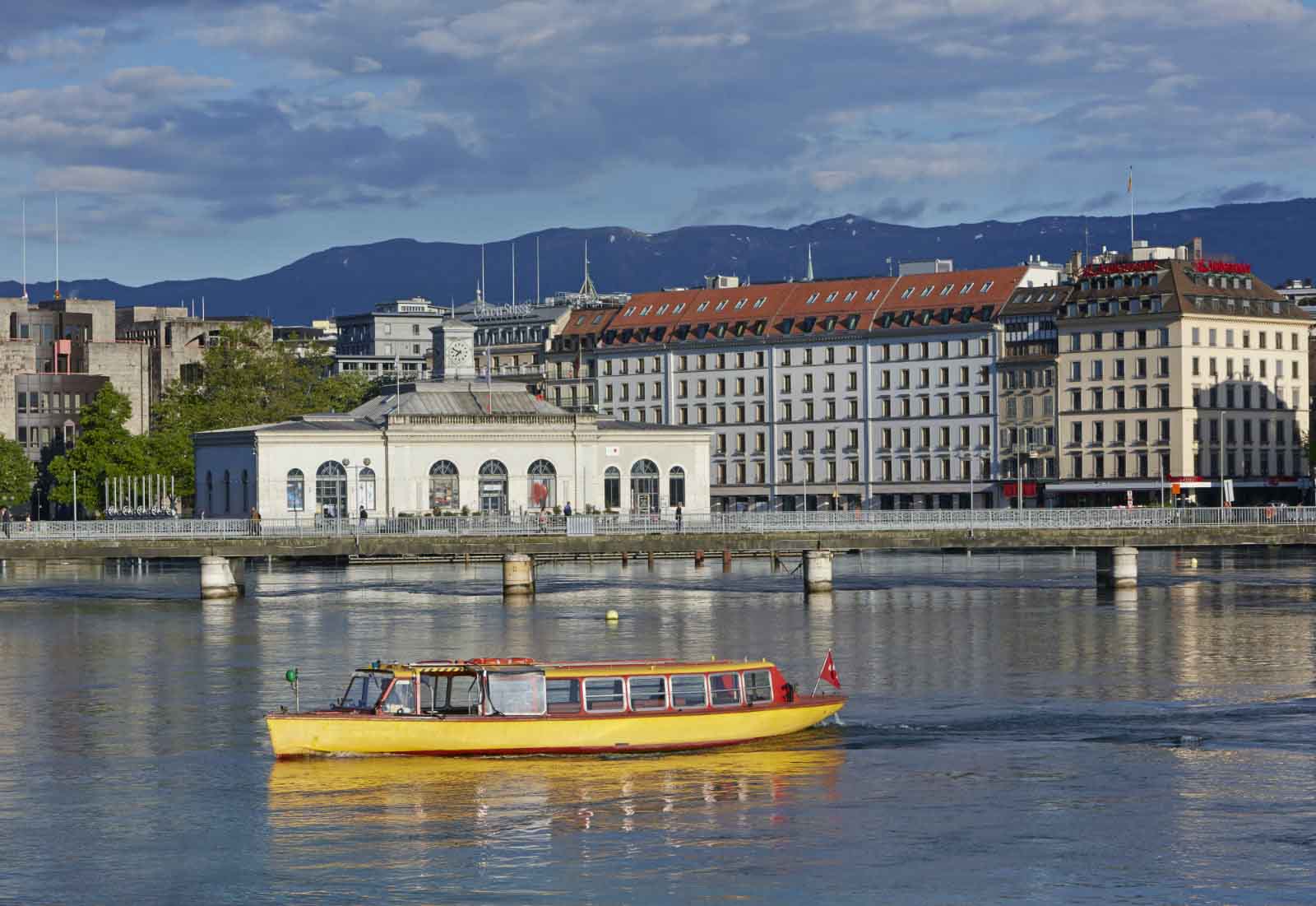} & 
    \includegraphics[width=2.0cm]{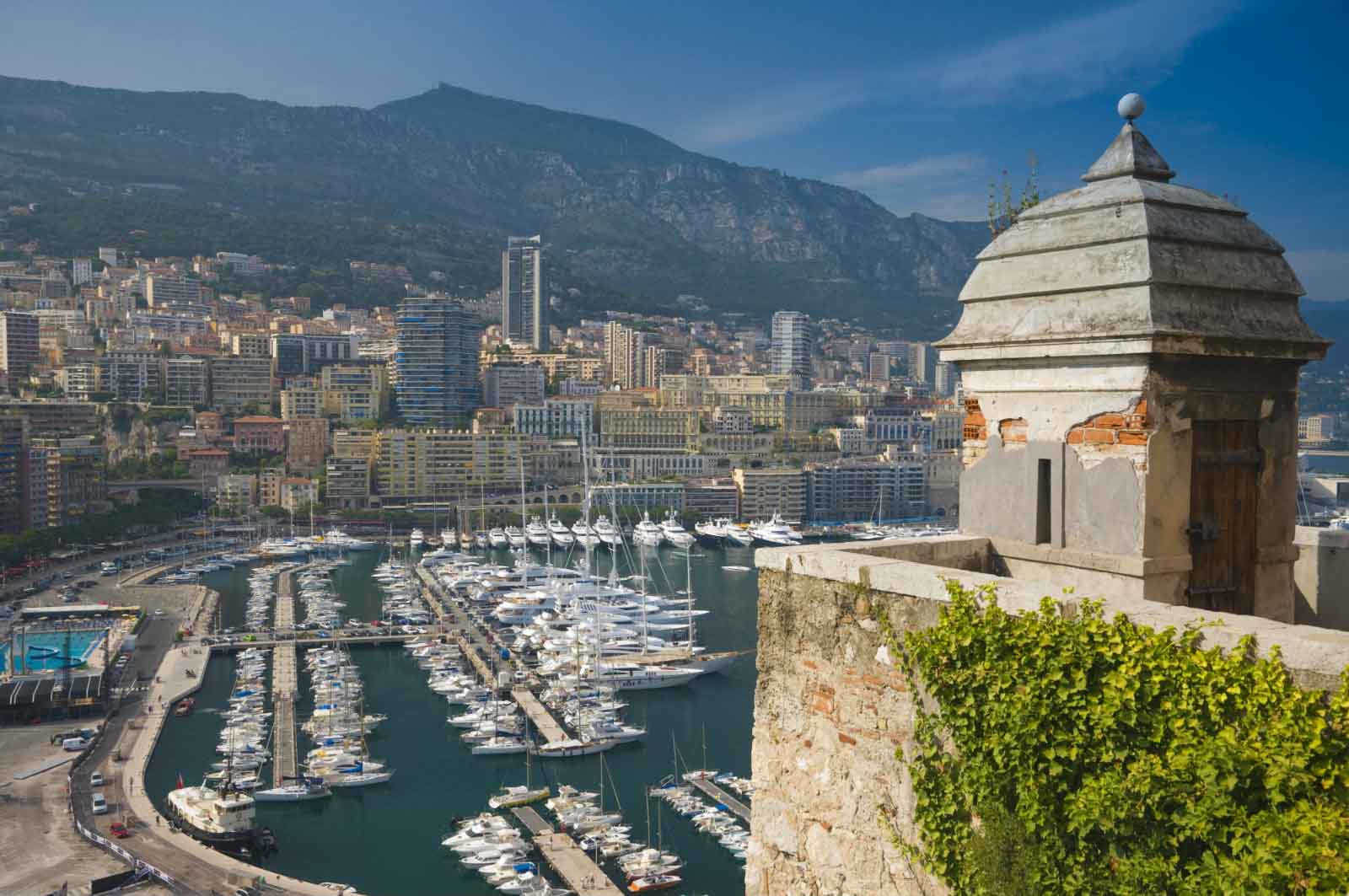} & 
    \includegraphics[width=2.0cm]{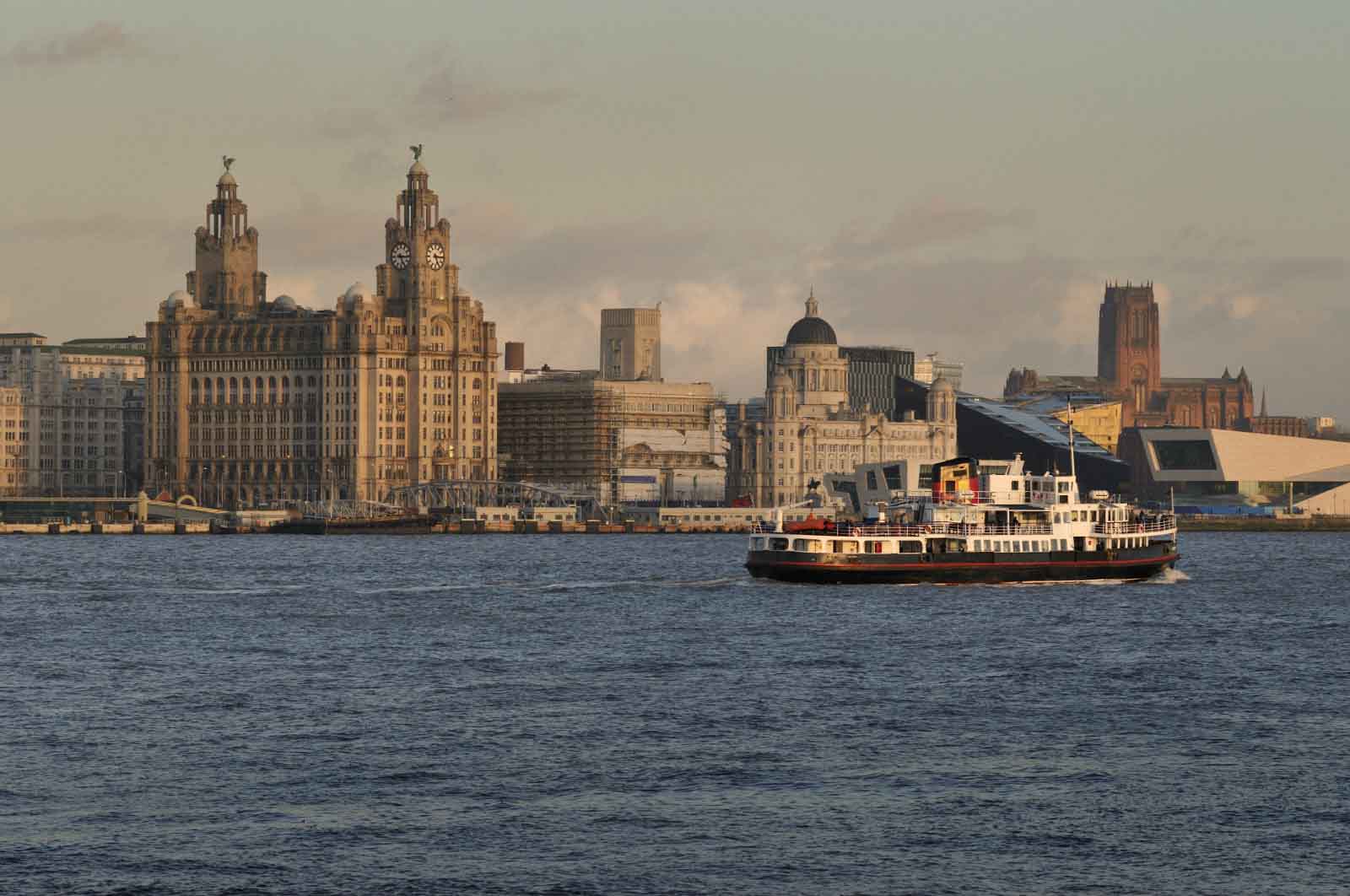}\vspace{0.02cm}\\
    \rotatebox{90}{NN} & 
    \includegraphics[width=2.0cm]{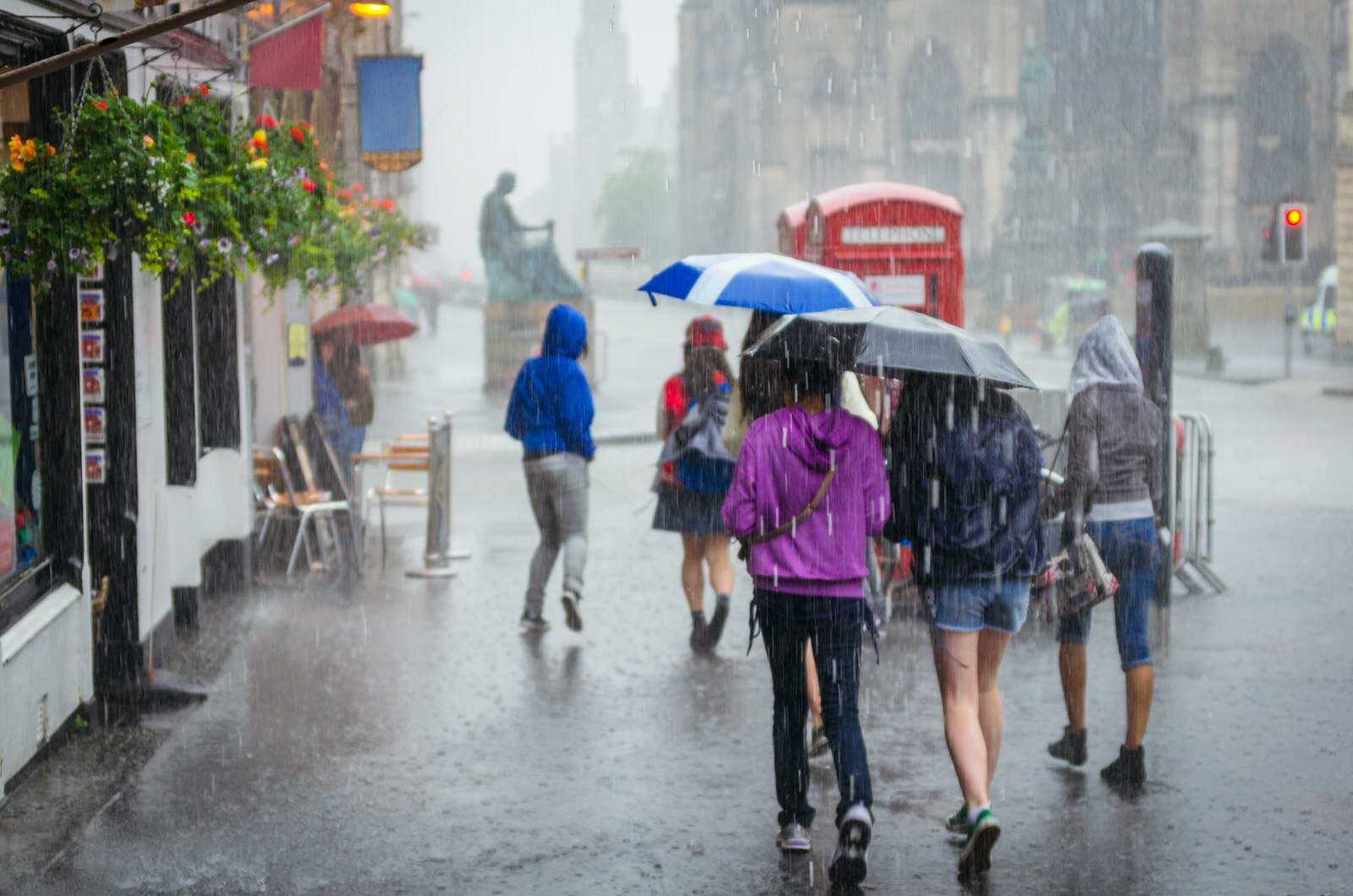} & 
    \tikzmarkin{a}(0.1,-0.1)(-0.05,1.45)
    \includegraphics[width=2.0cm]{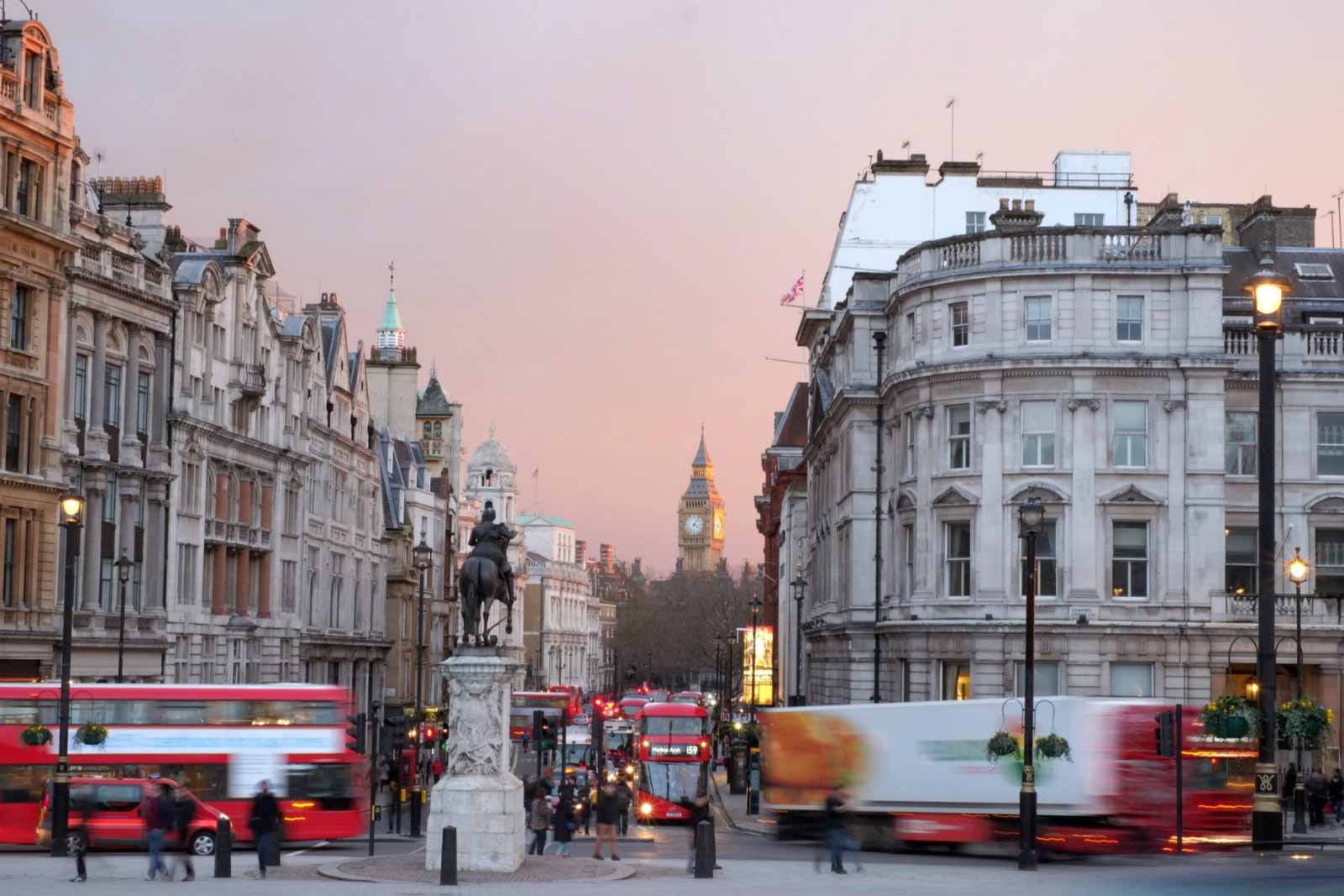}\tikzmarkend{a} &
    \includegraphics[width=2.0cm]{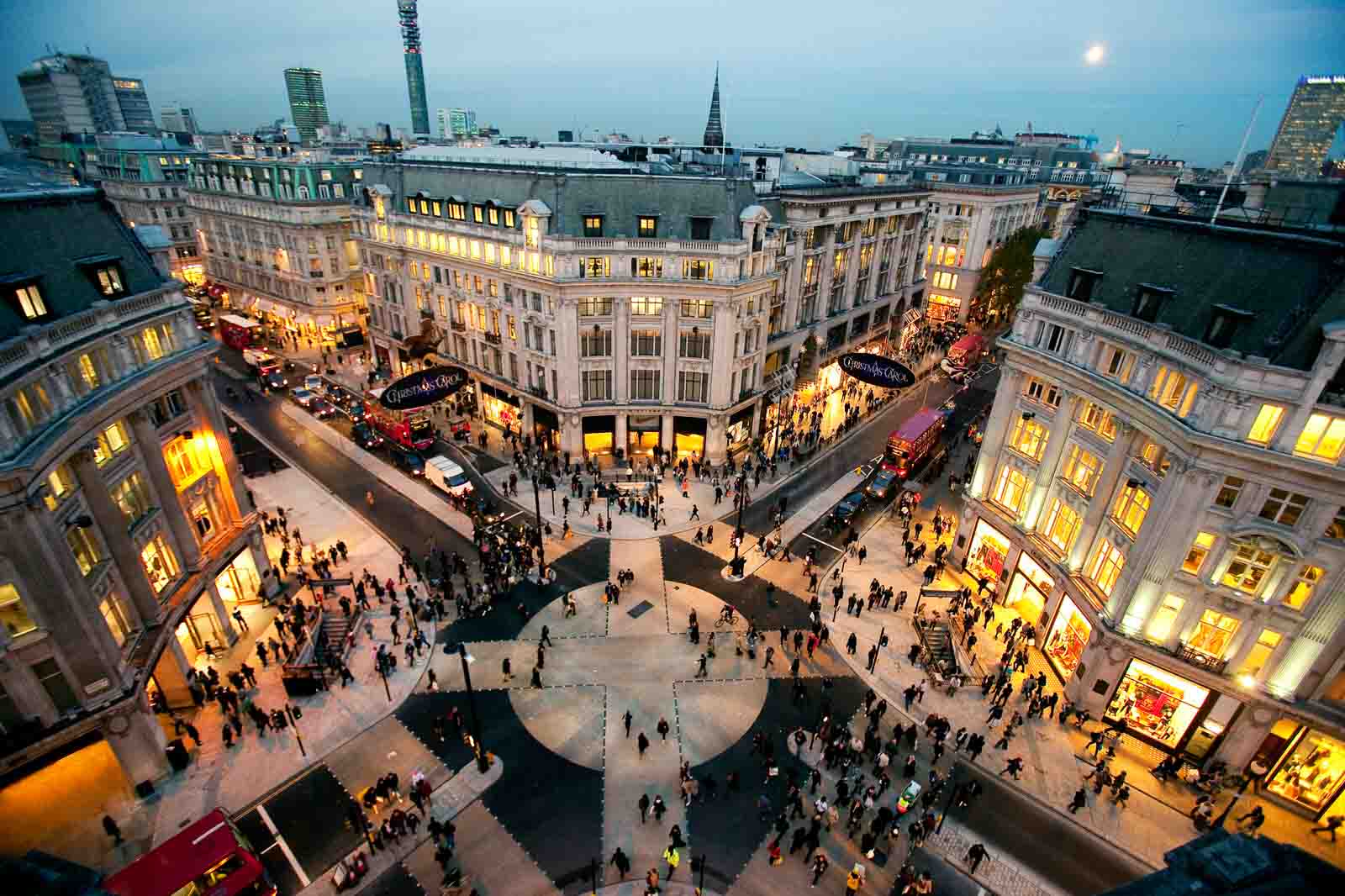} &
    \includegraphics[width=2.0cm]{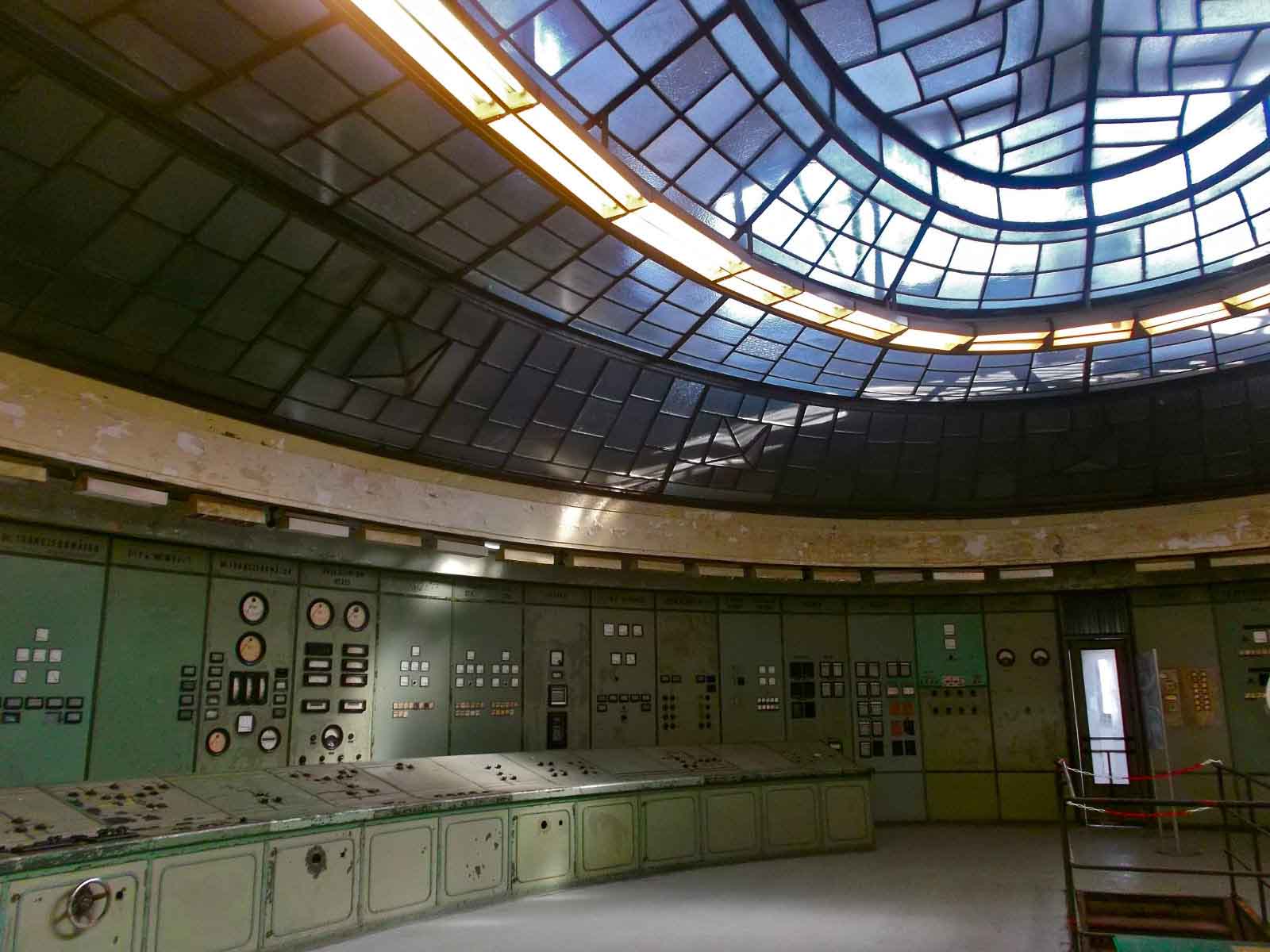} &
    \includegraphics[width=2.0cm]{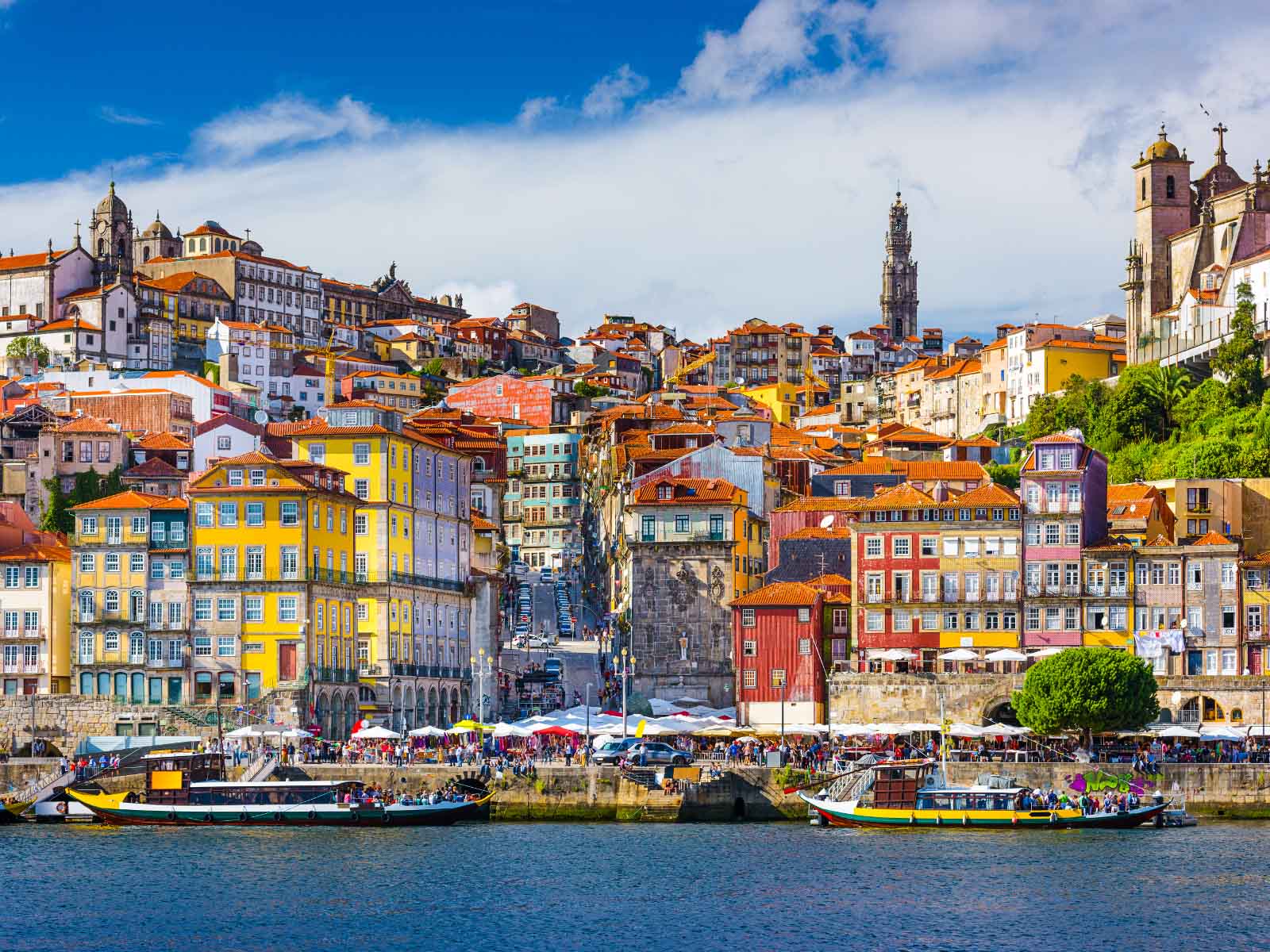} &
    \includegraphics[width=2.0cm]{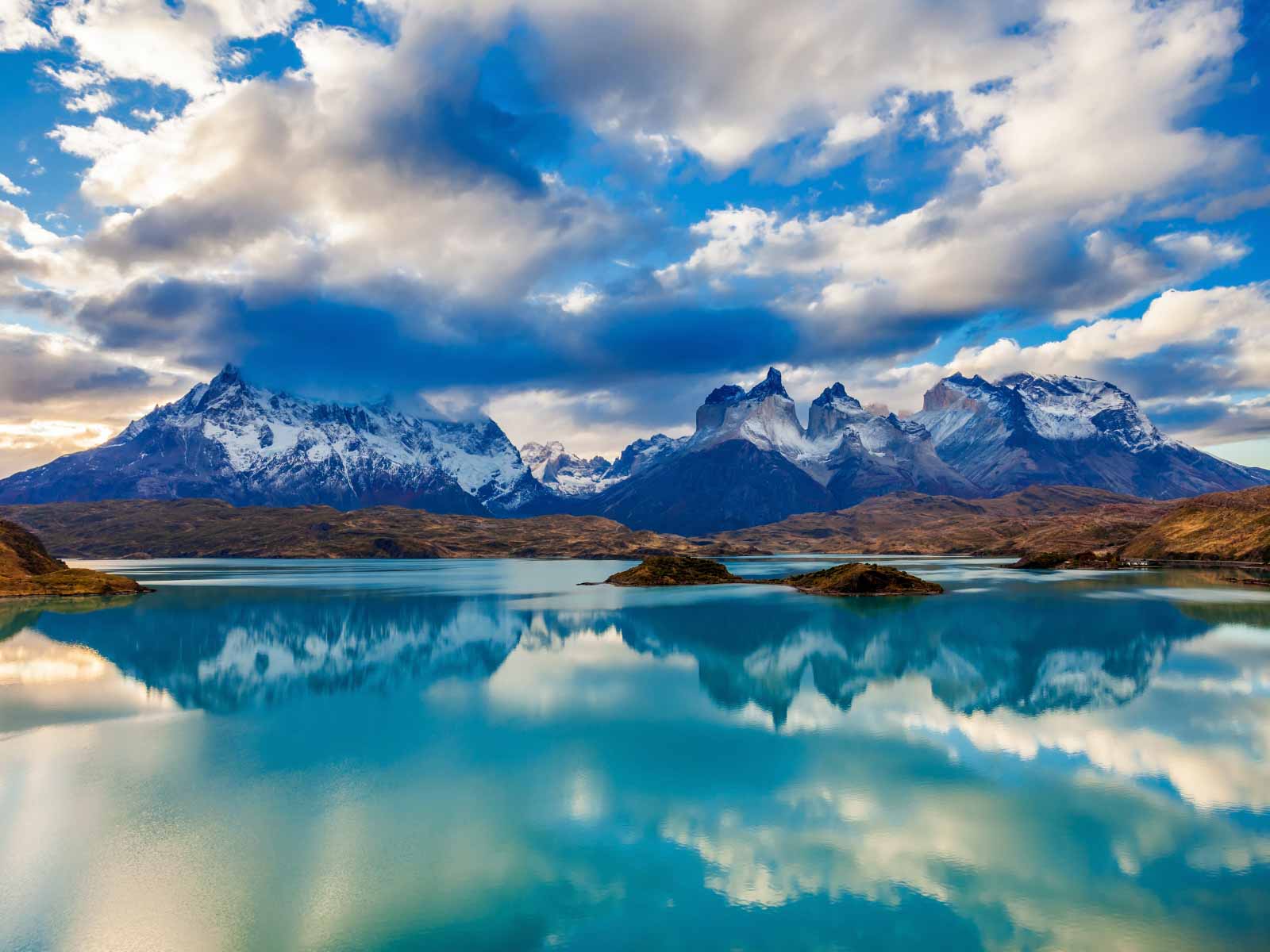} &
    \includegraphics[width=2.0cm]{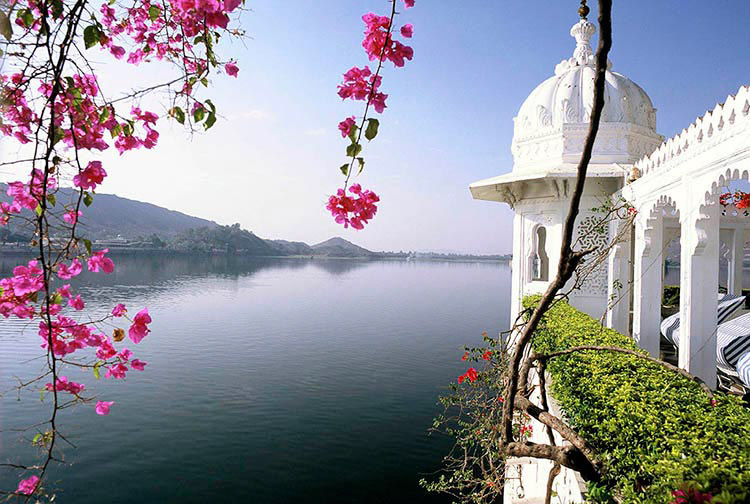} &
    \includegraphics[width=2.0cm]{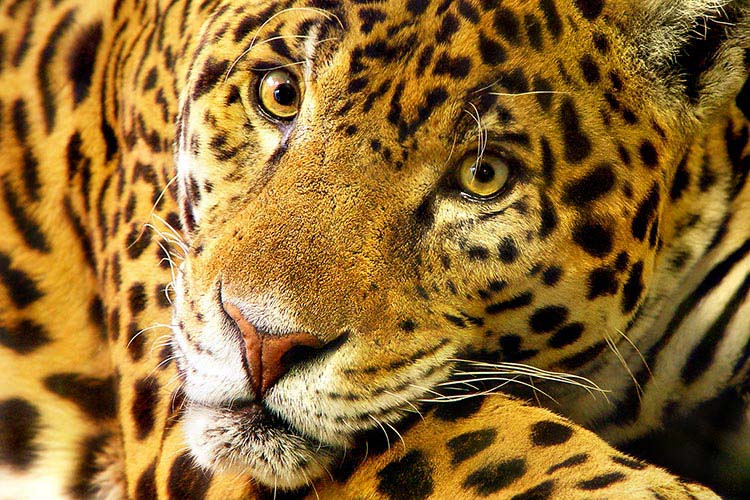} 
    \vspace{0.06cm}\\
    \rotatebox{90}{VSE++} & 
    \includegraphics[width=2.0cm]{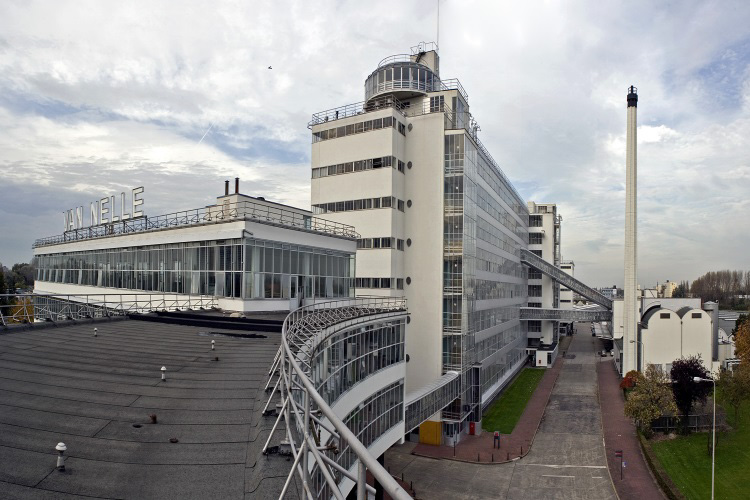} & 
    \includegraphics[width=2.0cm]{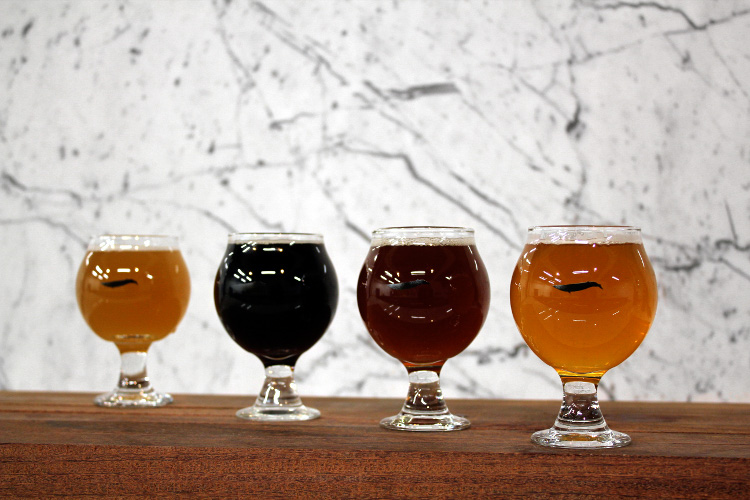} &
    \includegraphics[width=2.0cm]{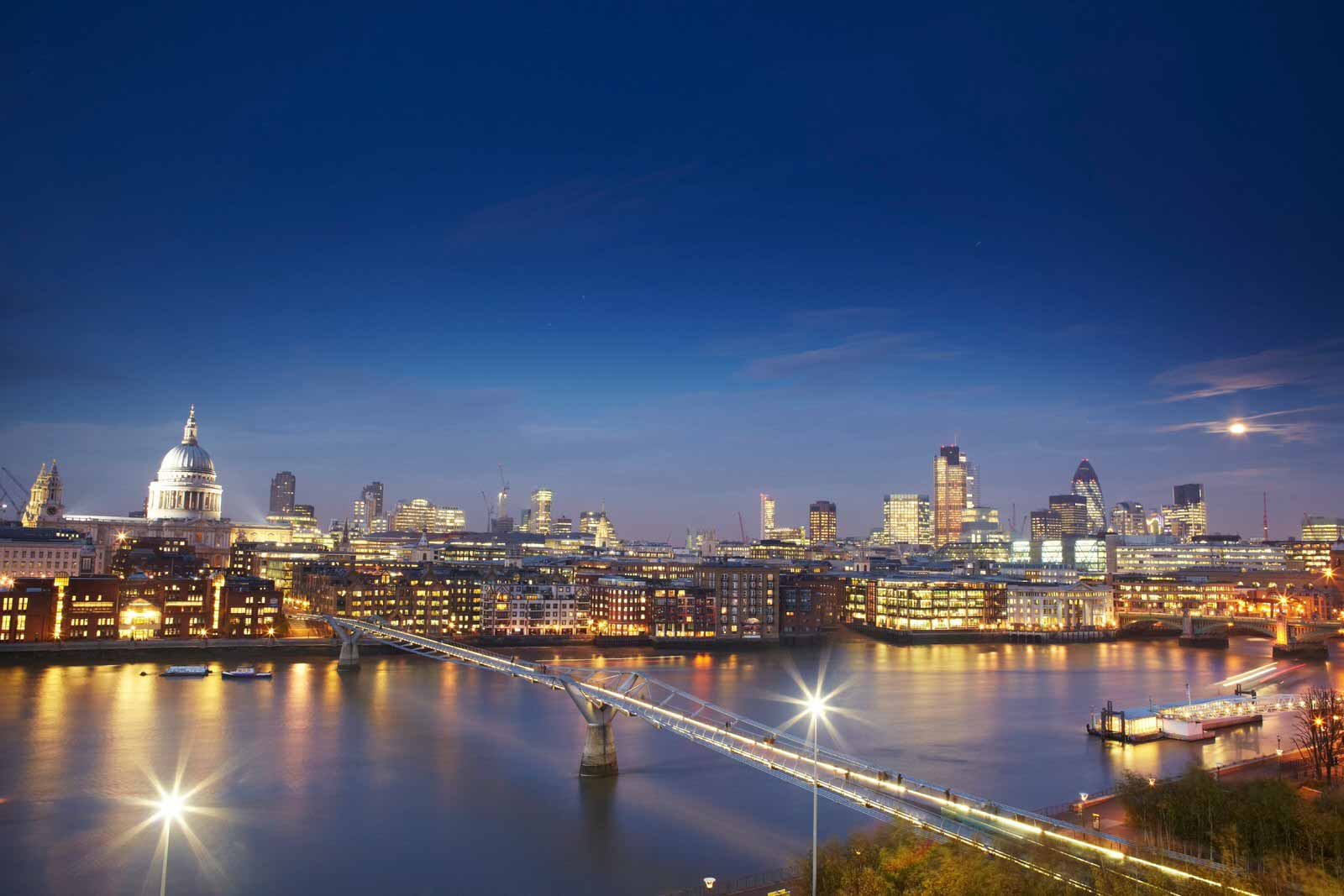} &
    \includegraphics[width=2.0cm]{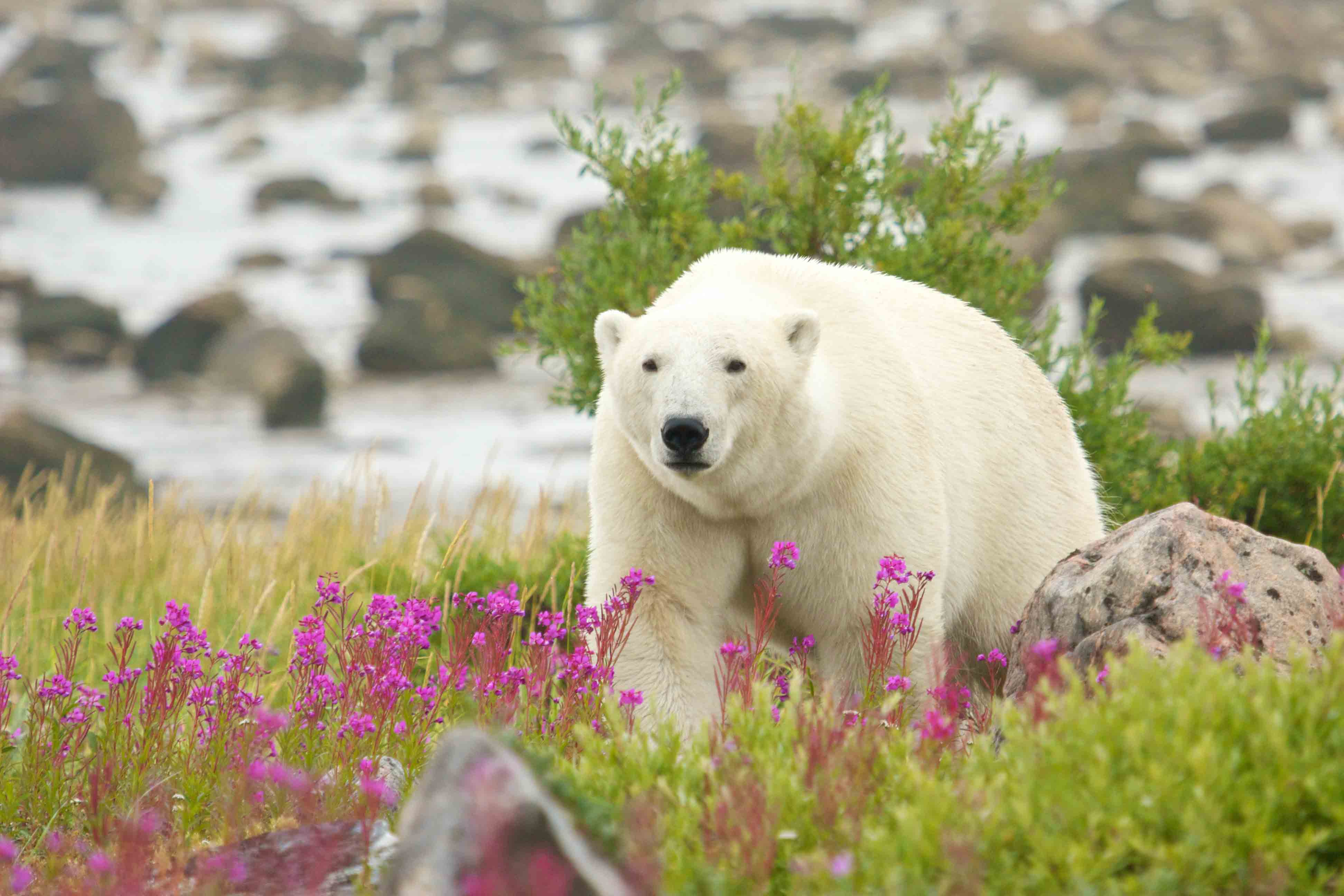} &
    \includegraphics[width=2.0cm]{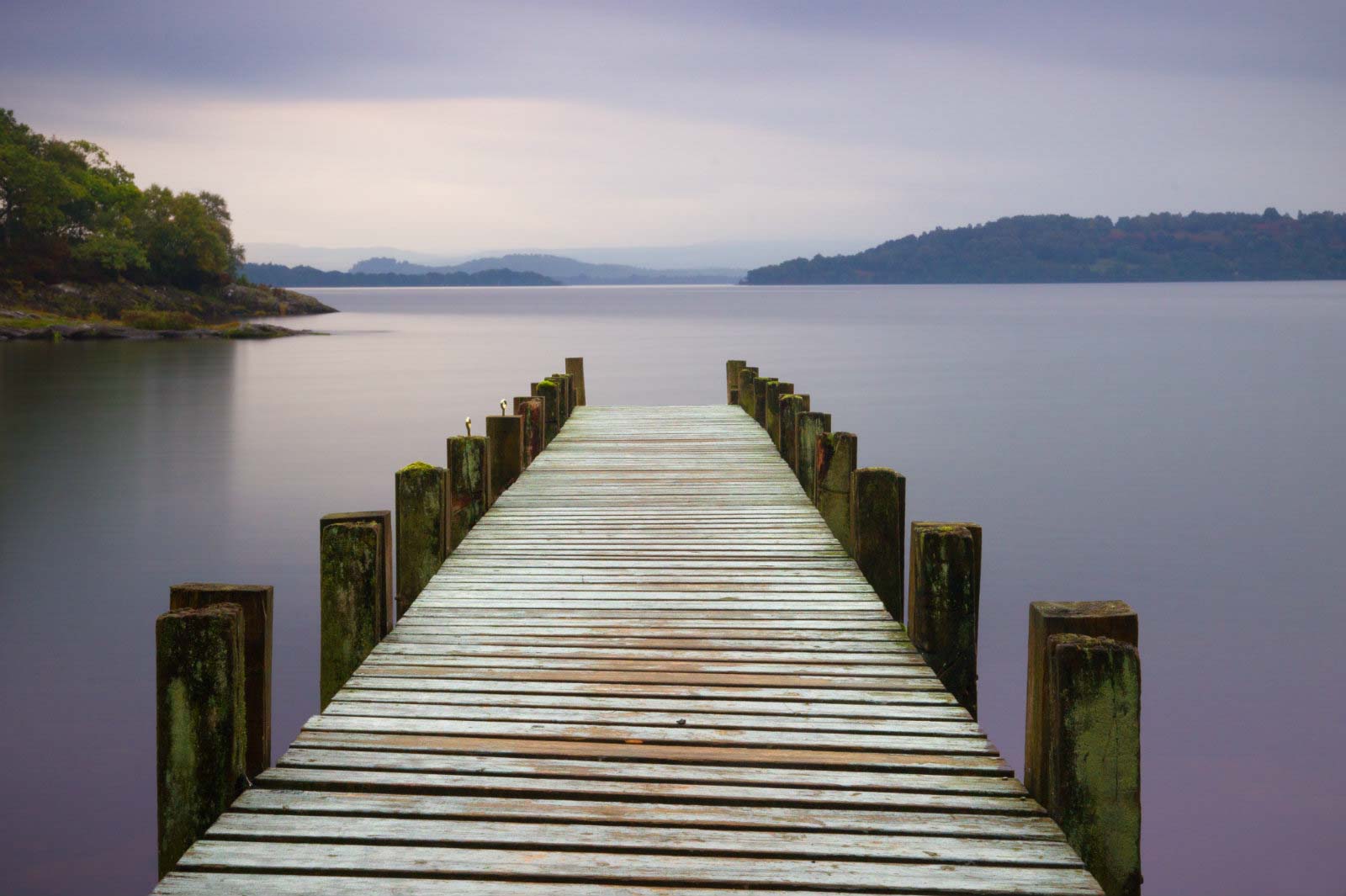} &
    \includegraphics[width=2.0cm]{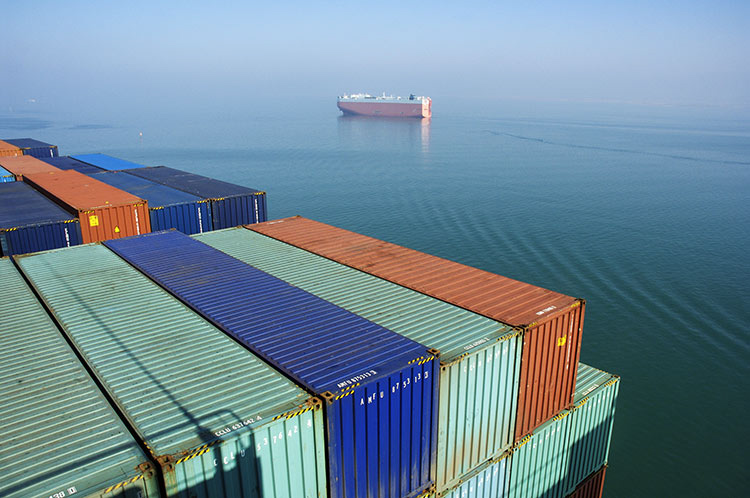} &
     \tikzmarkin{b}(0.1,-0.1)(-0.05,1.45)
     \includegraphics[width=2.0cm]{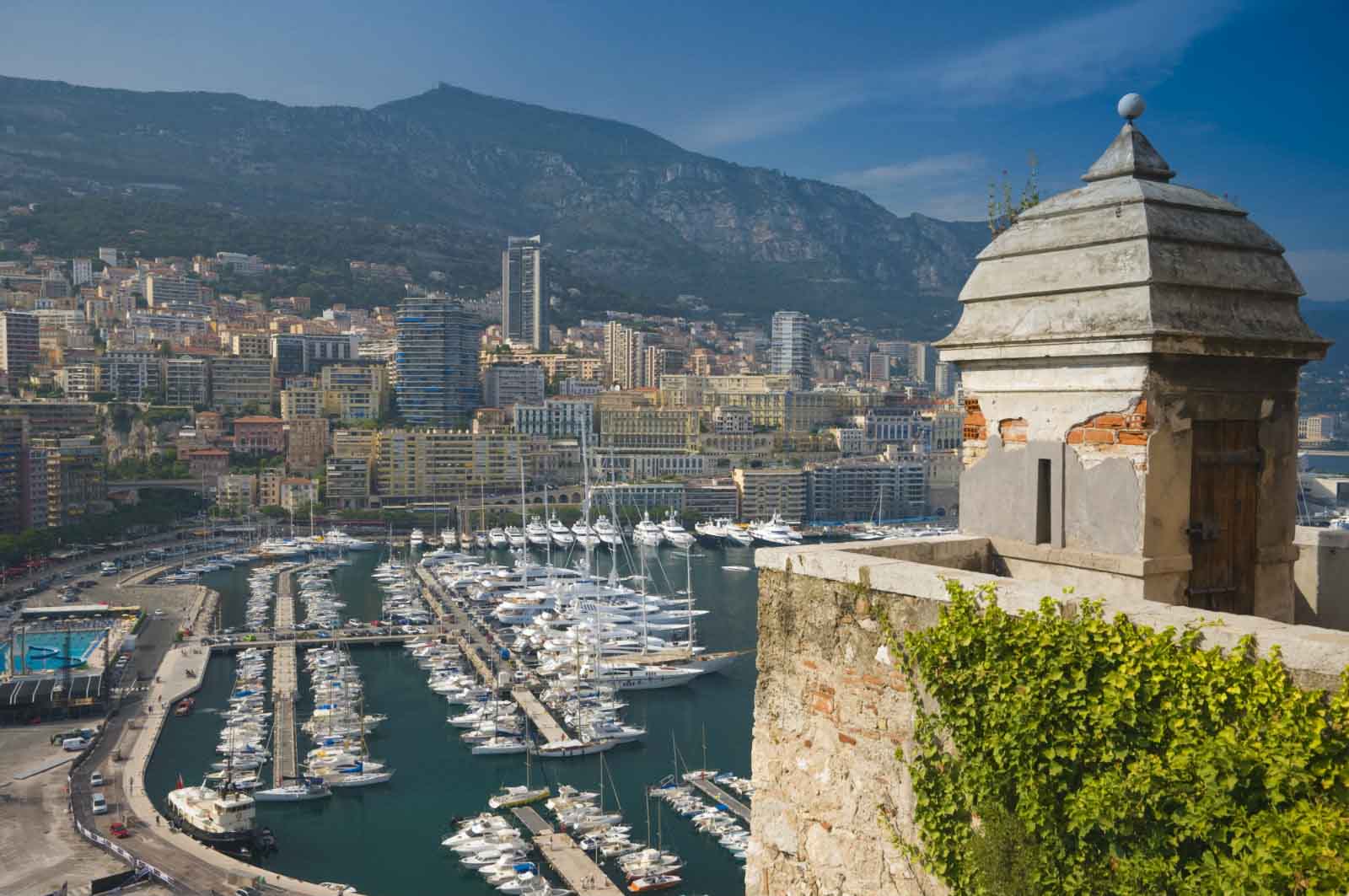}\tikzmarkend{b} &
    \includegraphics[width=2.0cm]{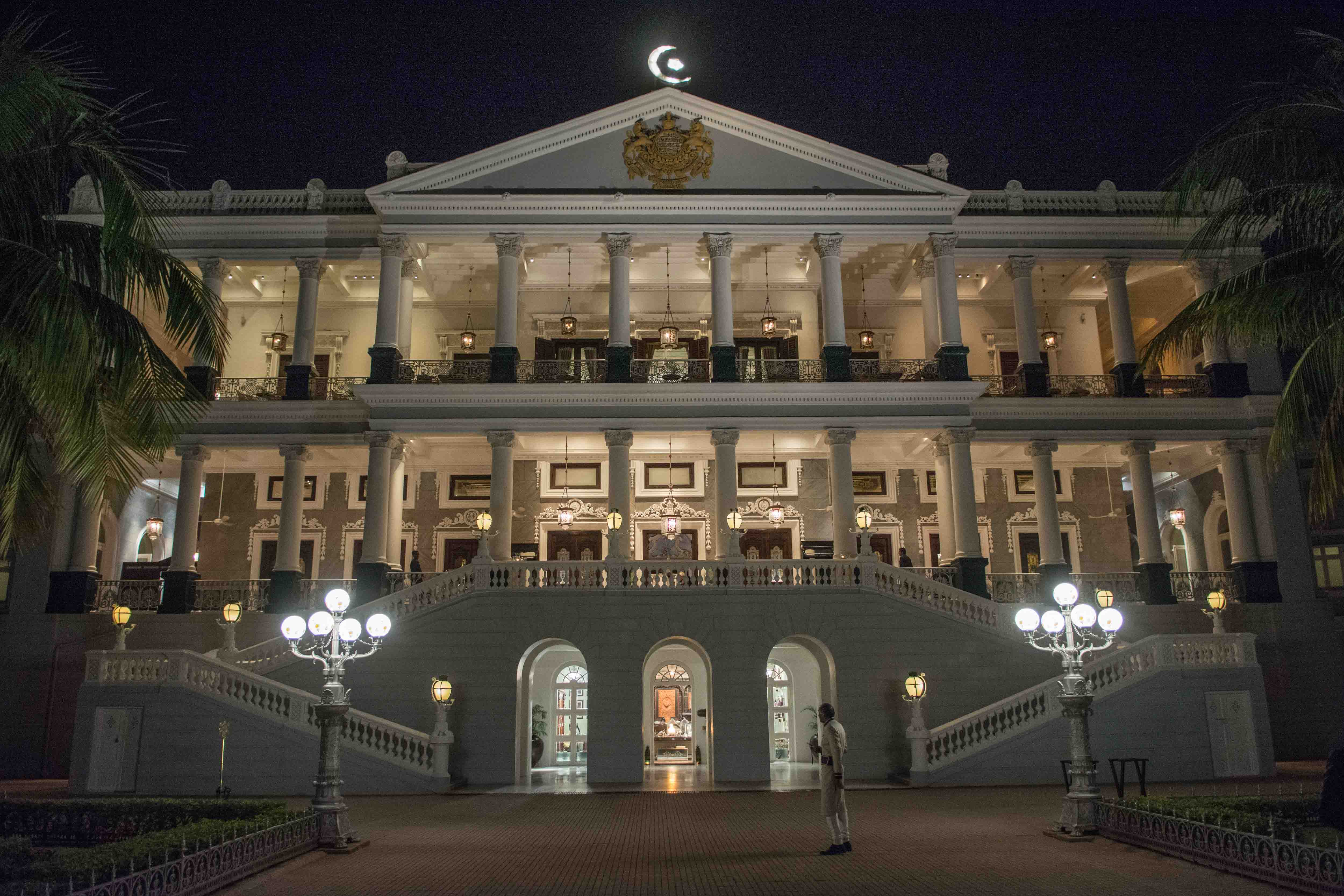} 
    \vspace{0.02cm}\\
    \rotatebox{90}{SANDI} & 
    \tikzmarkin{c}(0.1,-0.1)(-0.05,1.45)
    \includegraphics[width=2.0cm]{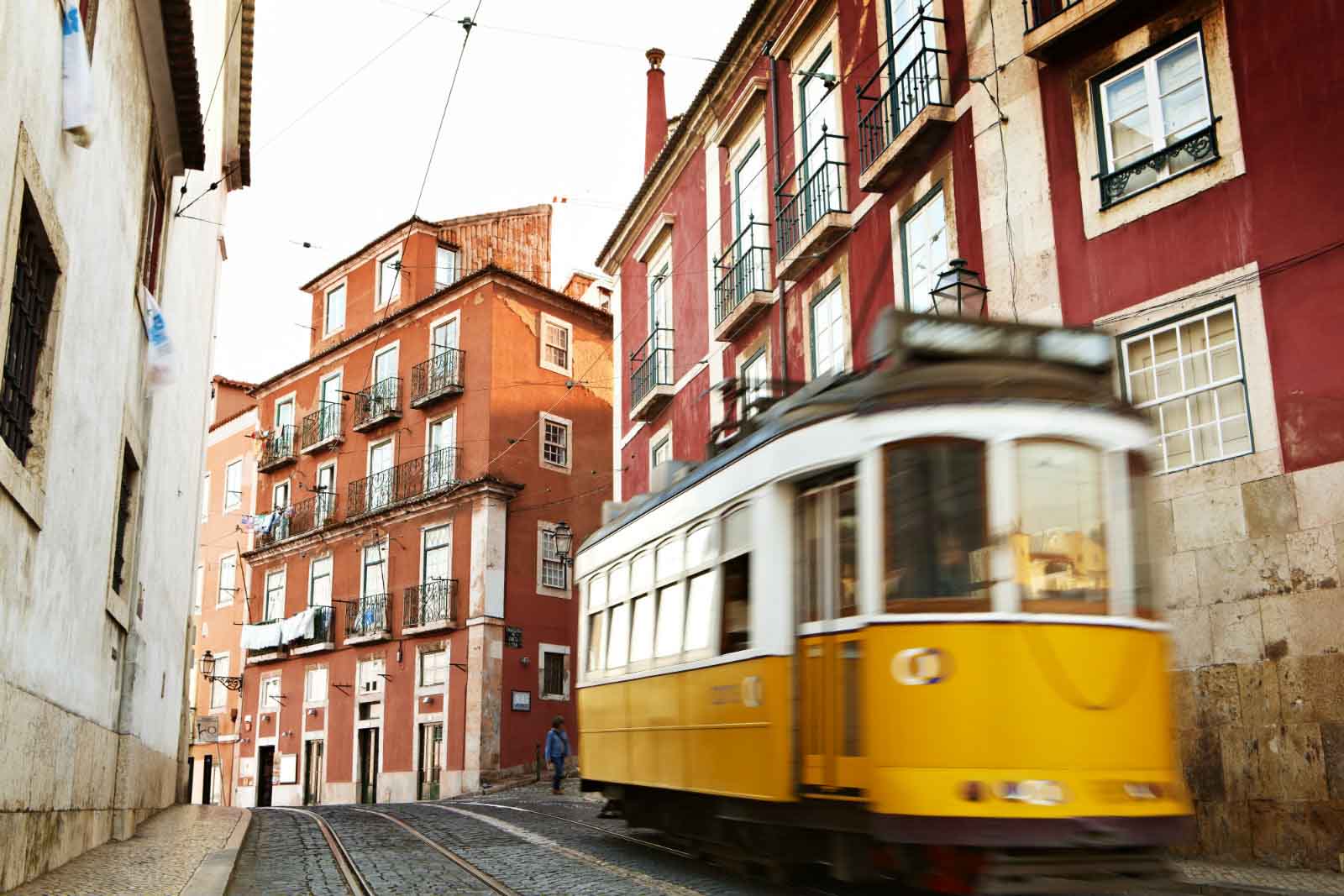} & 
    \includegraphics[width=2.0cm]{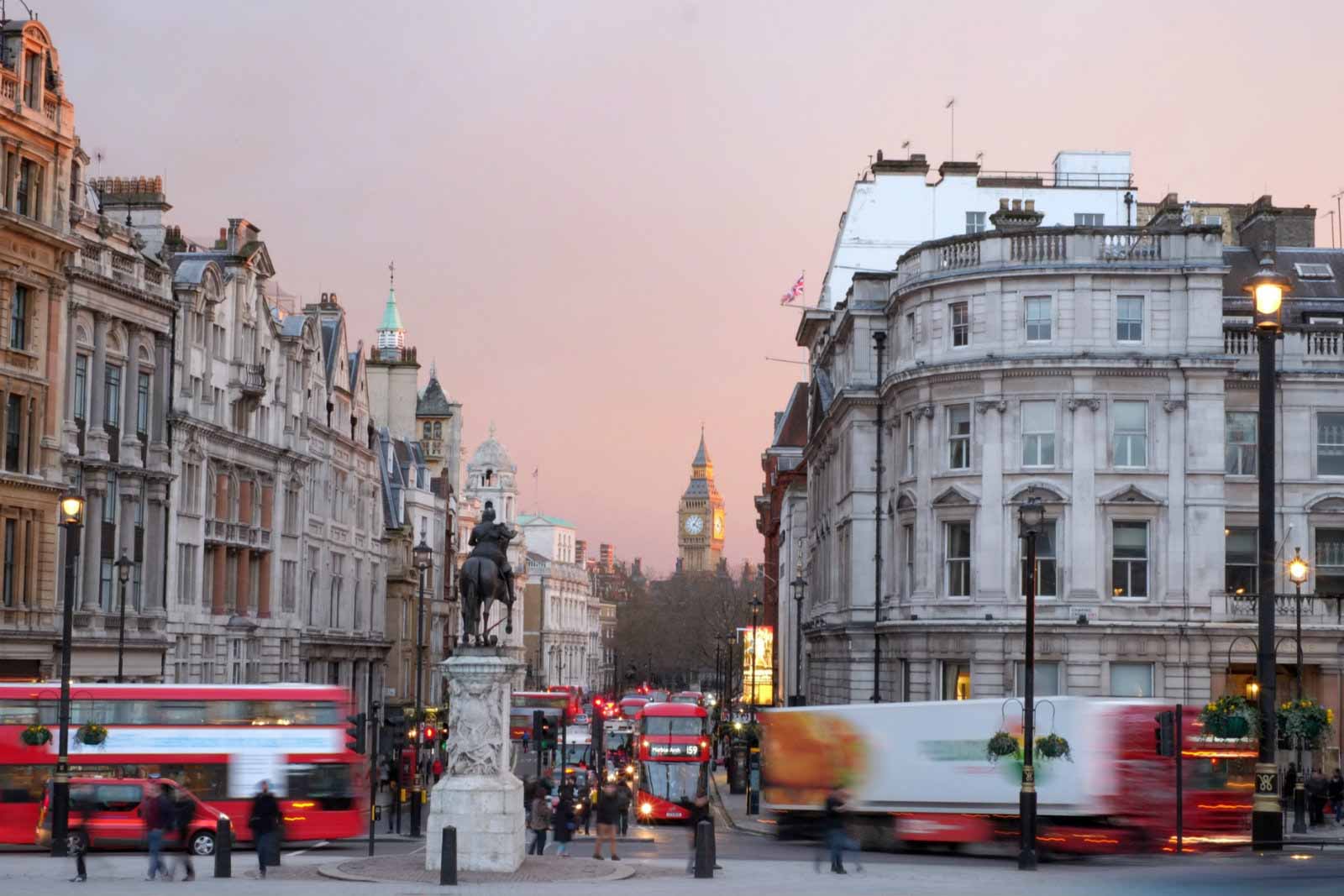}\tikzmarkend{c}& 
    \includegraphics[width=2.0cm]{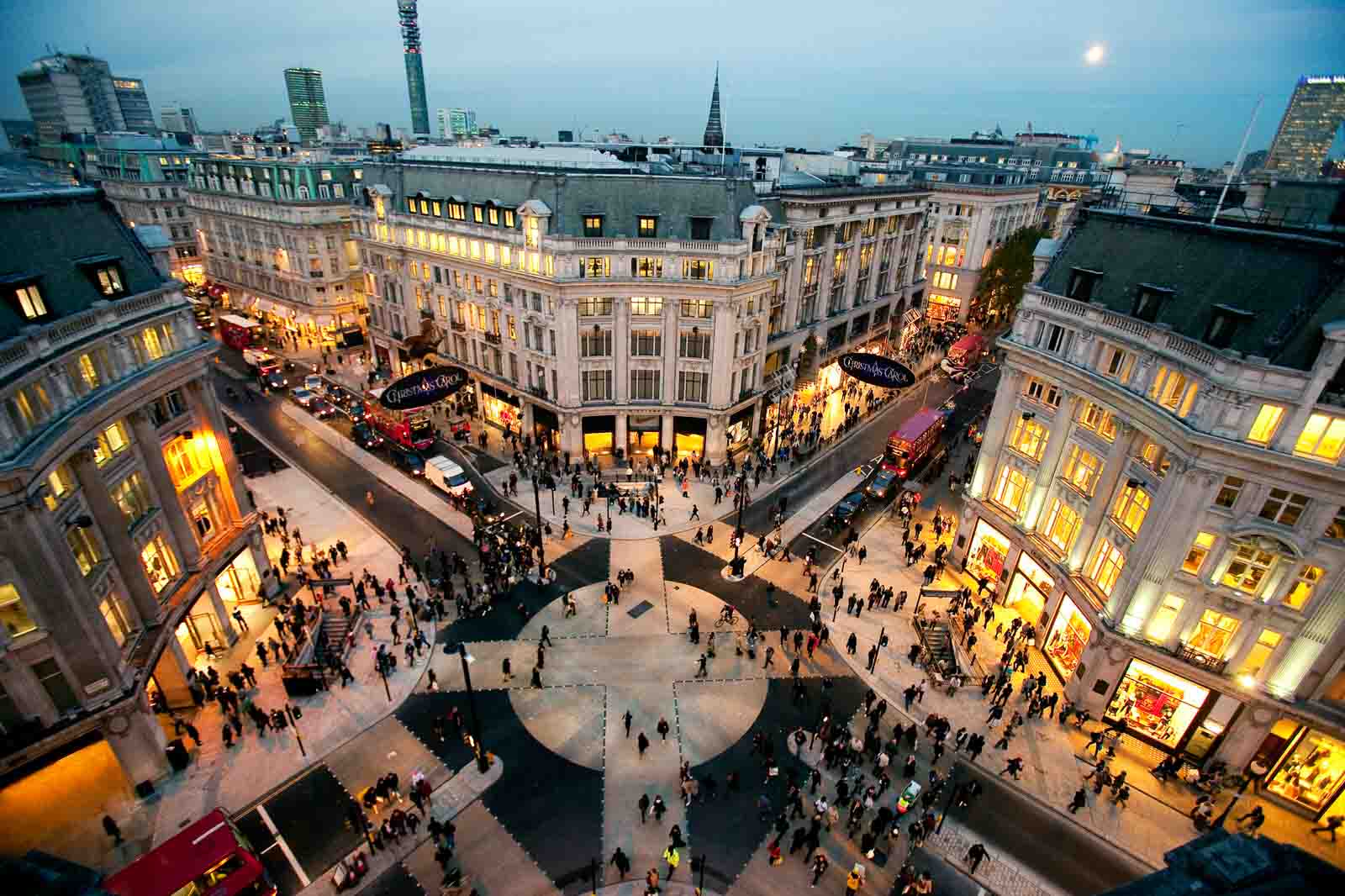} & 
    \includegraphics[width=2.0cm]{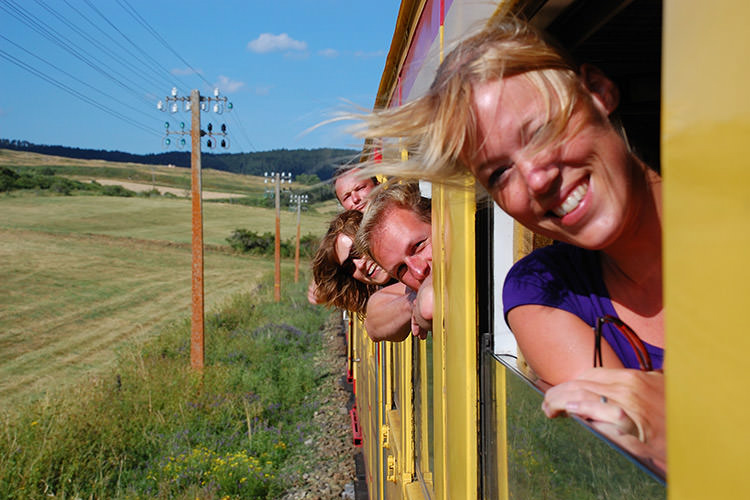} & 
    \includegraphics[width=2.0cm]{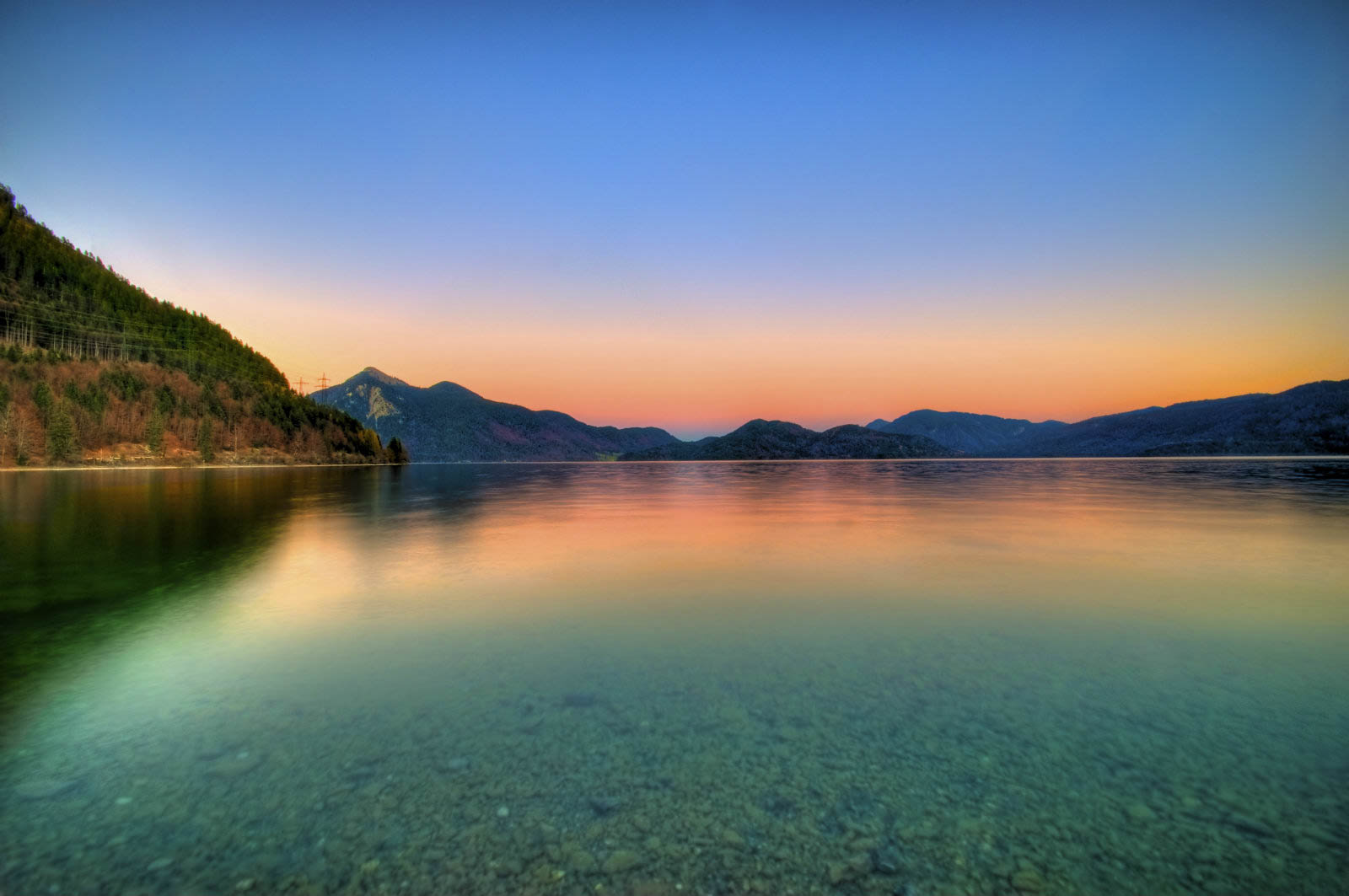} & 
    \includegraphics[width=2.0cm]{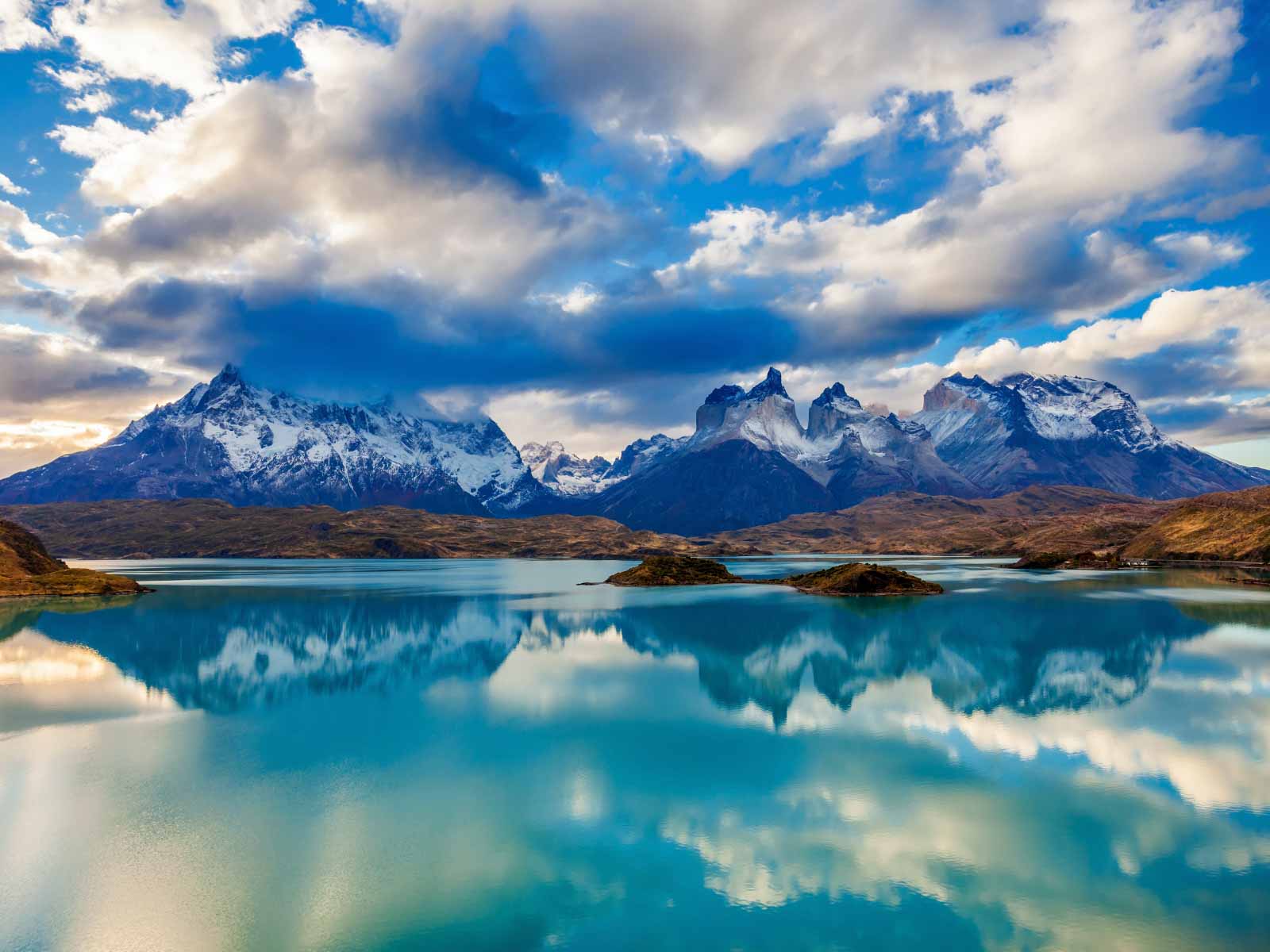} & 
    \includegraphics[width=2.0cm]{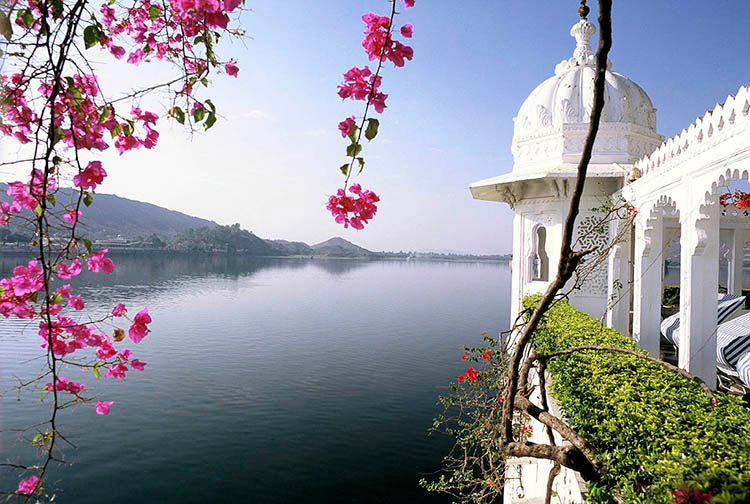} & 
    \includegraphics[width=2.0cm]{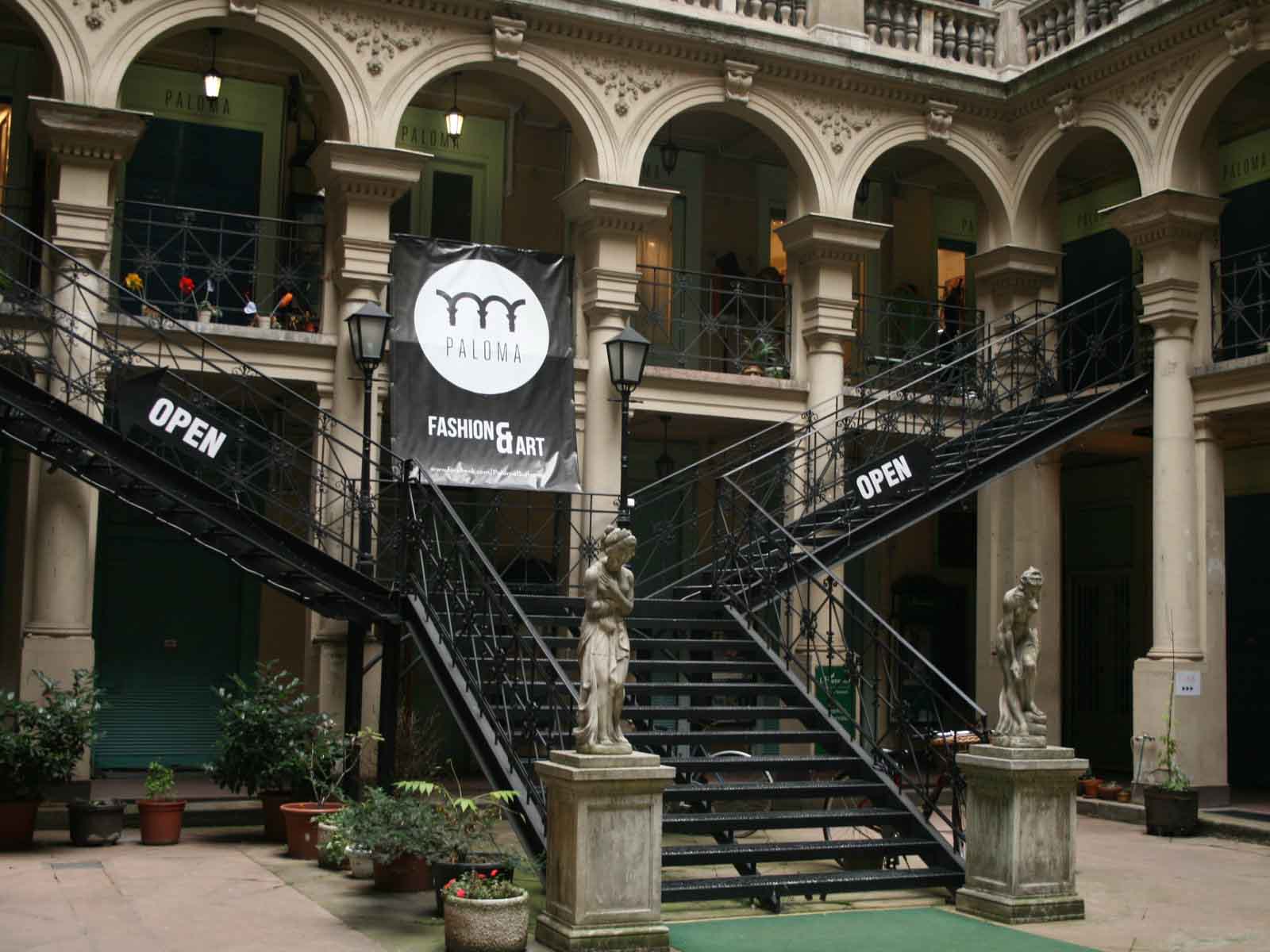}
    \vspace{0.02cm}
    \\
\hline
\vspace{-1.7em}
\end{tabular}
\caption{Image Selection. Images within green boxes are exact matches with ground truth (GT). SANDI retrieves more exact matches than the baselines (NN, VSE++). SANDI's non-exact matches are also much more thematically similar.}
\label{selectiveAlignmentAnecdote}
\end{figure*}


Figure~\ref{selectiveAlignmentAnecdote} shows anecdotal image selection results for one story. The original story contains 17 paragraphs; only the main concepts from the story have been retained in the figure for readability. SANDI is able to retrieve 2 
ground-truth images out of 8, while the baselines retrieve 1 each. Note that the remaining non-exact matches retrieved by SANDI are also thematically similar. This can be attributed to the wider space of concepts that SANDI explores through the different types of image descriptors described in Section~\ref{features}. 

\vspace{0.5cm}
\noindent
\textbf{\large6.3.2 \hspace{0.6em} Image Placement}

\vspace{0.5em}
\noindent
Having selected thematically related images from a big image pool, SANDI places them within contextual paragraphs of the story. 
Note that SANDI actually integrates the Image Selection and Image Placement
stages into joint inference on selective alignment seamlessly,
whereas the baselines operate in two sequential steps.

We evaluate the alignments by the measures from Section~\ref{image_placement_metrics}.
Note that the measure \textit{OrderPreserve} does not apply to Selective Alignment since the images are selected from a pool of mixed images which cannot be ordered.
Tables~\ref{evaluation-selectiveAlignment-lonelyPlanet} and ~\ref{evaluation-selectiveAlignment-asiaExchange} show results for the Lonely Planet and Asia Exchange datasets respectively.
We observe that SANDI outperforms the baselines by a clear
margin, harnessing its more expressive pool of tags.
This holds for all the different metrics (to various degrees).
We show anecdotal evidence of the diversity of our image descriptors in Figure~\ref{tagSources} and Table~\ref{example-studyAbroad}.%


\begin{table}[b]
\centering
\caption{Selective Alignment on the Asia Exchange dataset.}
\begin{tabular}{l|lllll}
\hline
          & \rotatebox{60}{BLEU} & \rotatebox{60}{ROUGE} & \rotatebox{60}{SemSim} & \rotatebox{60}{ParaRank} \\
          \hline
Random & 2.06 & 1.37 & 53.14 & 58.28  \\
VSE++ \cite{DBLP:conf/bmvc/FaghriFKF18}      & 2.66 & 1.39   & 58.00   &   64.34  \\
VSE++ ILP     & 2.78 & 1.47 & 57.65 & 64.29 \\[0.2em]
\hdashline
SANDI-CV  & 1.04 & 1.51  &  60.28  &  75.42   \\
SANDI-MAN & \textbf{3.49} & \textbf{2.98}  &  61.11  &  \textbf{82.00}   \\
SANDI-BD  & 1.68 & 1.52  &  \textbf{76.86}  &  70.41    \\
SANDI{\LARGE $\ast$}    & 1.53 & 1.84  &  64.76  &  80.57    \\ 
\hline
\end{tabular}
\label{evaluation-selectiveAlignment-asiaExchange}
\vspace{-0.1cm}
\end{table}

\subsection{Role of Model Components}

\ourheading{Image Descriptors}
Table~\ref{example-studyAbroad} shows alignments for a section of a single story from three SANDI variants. Each of the variants capture special characteristics of the images, hence aligning to different paragraphs. The paragraphs across variants are quite semantically similar. The highlighted key concepts bring out the similarities and justification of alignment. The wide variety of image descriptors that SANDI leverages (CV, BD, MAN, CSK) is unavailable to VSE++, attributing to the latter's poor performance.

\ourheading{Embeddings} 
The nature of embeddings is decisive towards alignment quality. Joint visual-semantic-embeddings trained on MSCOCO (used by VSE++) fall short in capturing high-level semantics between images and story. Word2Vec embeddings trained on a much larger and domain-independent Google News corpus better represents high-level image-story interpretations.

\ourheading{ILP} 
Combinatorial optimization (Integer Linear Programming) wins in performance over greedy optimization approaches. In 
Tables \ref{storyIllustration-lonelyPlanet} and \ref{storyIllustration-asiaExchange} 
this phenomenon can be observed between NN (greedy) and SANDI (ILP). This pair of approaches make use of the same embedding space, with SANDI outperforming NN. 

\begin{table*}[t]
\small
\centering
\caption{Example alignments. Highlighted texts show similar concepts between image and aligned paragraphs.}%
\begin{tabular}{p{4.1cm}|p{4.1cm}|p{4.1cm}|p{4.1cm}}
Image and detected concepts				& SANDI-CV & SANDI-MAN & SANDI-BD \\[0.2em]
\hline
\makecell[l]{\raisebox{-\totalheight}{\includegraphics[width=4cm]{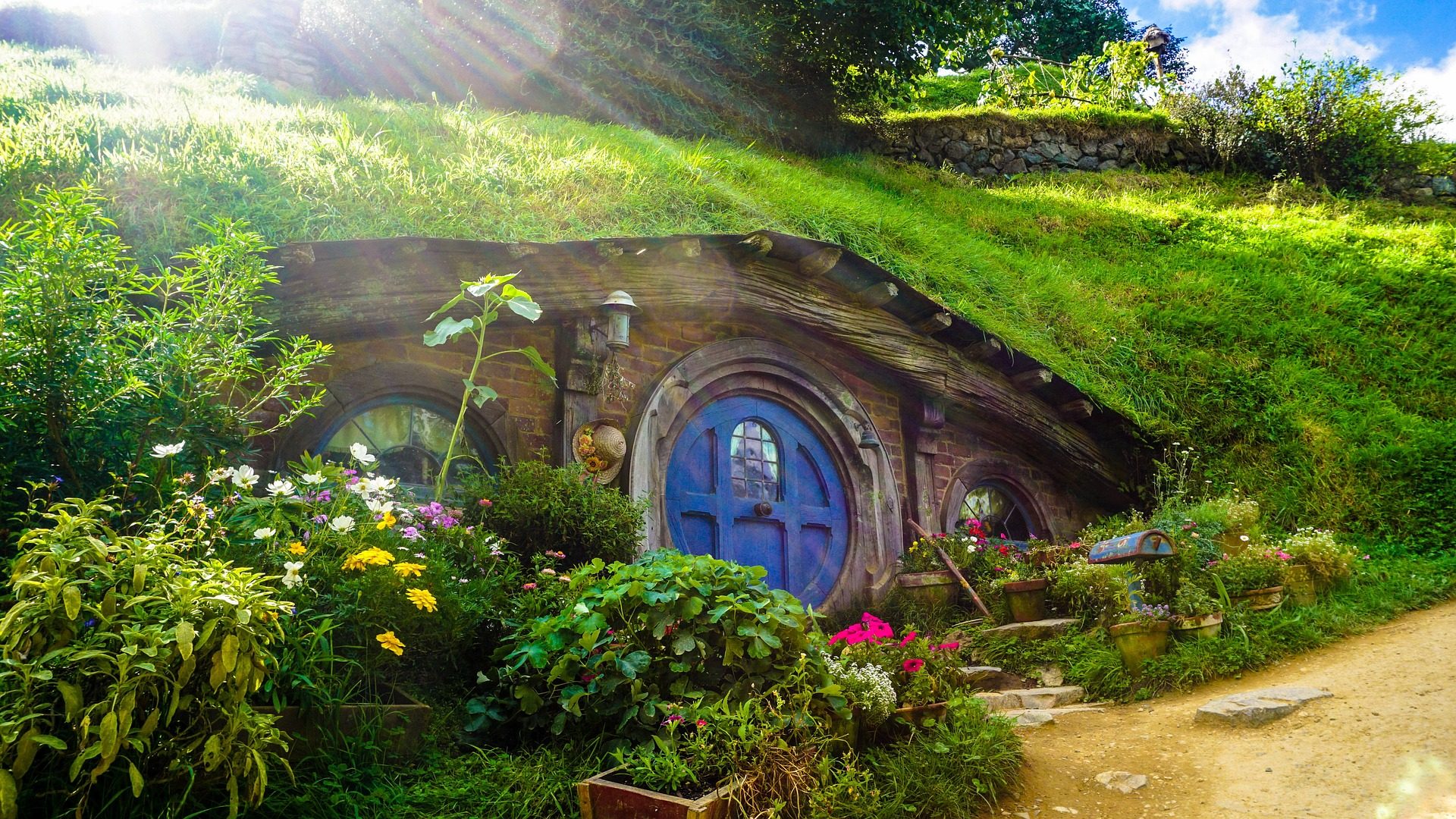}}\\[-1em]
\\
CV: \hl{cottage}, flower-pot, carrier\\%
MAN: Shire, \hl{Bilbo Baggins}\\
BD: \hl{New Zealand}, \hl{hobbit house}\\}\vspace{0.2em}
										 & \vspace{-1.7cm}Take advantage of your stay here and visit the memorable scenes shown in the movies. Visit The Shire and experience firsthand the 44 \hl{Hobbit Holes} from which Bilbo Baggins emerged to commence his grand adventure. Tongariro National Park is home to the feared Mount Doom in The Lord of the Rings. Other famous locations that you can visit are \hl{Christchurch}, Nelson and Cromwell.
                                        &\vspace{-1.7cm} Home to \hl{hobbits}, warriors, orcs and dragons. If you're a fan of the famous trilogies, \hl{Lord of the Rings} and The Hobbit, then choosing New Zealand should be a no-brainer.
                                        &\vspace{-1.7cm}Take advantage of your stay here and visit the memorable scenes shown in the movies. Visit The Shire and experience firsthand the 44 \hl{Hobbit Holes} from which Bilbo Baggins emerged to commence his grand adventure. Tongariro National Park is home to the feared Mount Doom in \hl{The Lord of the Rings}. Other famous locations that you can visit are Christchurch, Nelson and Cromwell.\\[0.4em]
\hline
\makecell[l]{\raisebox{-\totalheight}{\includegraphics[width=4cm]{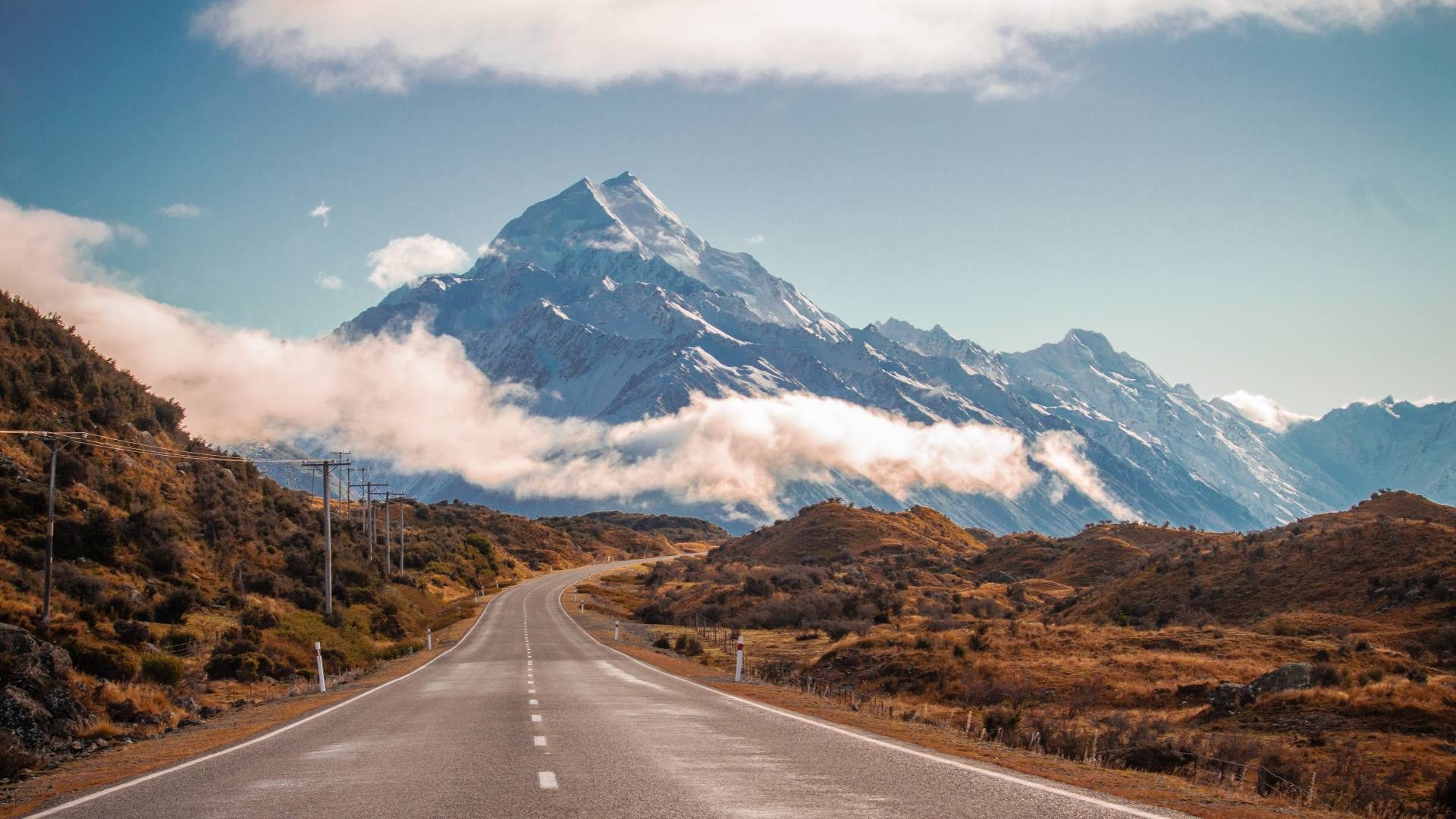}}\\[-1em]
\\
CV: \hl{snowy mountains}, \hl{massif}, alpine \\glacier, mountain range\\
MAN: \hl{outdoor lover}, \hl{New Zealand}, \\study destination, 
BD: \hl{New Zealand}}
										 & \vspace{-1.9cm}New Zealand produced the first man to ever climb \hl{Mount Everest} and also the creator of the \hl{bungee-jump}. Thus, it comes as no surprise that this country is filled with adventures and adrenaline junkies.
                                        &\vspace{-1.9cm}Moreover, the \hl{wildlife} in \hl{New Zealand} is something to behold. Try and find a Kiwi! (the bird, not the fruit). They are nocturnal creatures so it is quite a challenge. New Zealand is also home to the smallest dolphin species, Hector's Dolphin. Lastly, take the opportunity to search for the beautiful yellow-eyed penguin.
                                        &\vspace{-1.9cm}Home to hobbits, warriors, orcs and dragons. If you're a fan of the famous trilogies, Lord of the Rings and The Hobbit, then choosing \hl{New Zealand} should be a no-brainer.\\
\hline
\makecell[l]{\raisebox{-\totalheight}{\includegraphics[width=4cm]{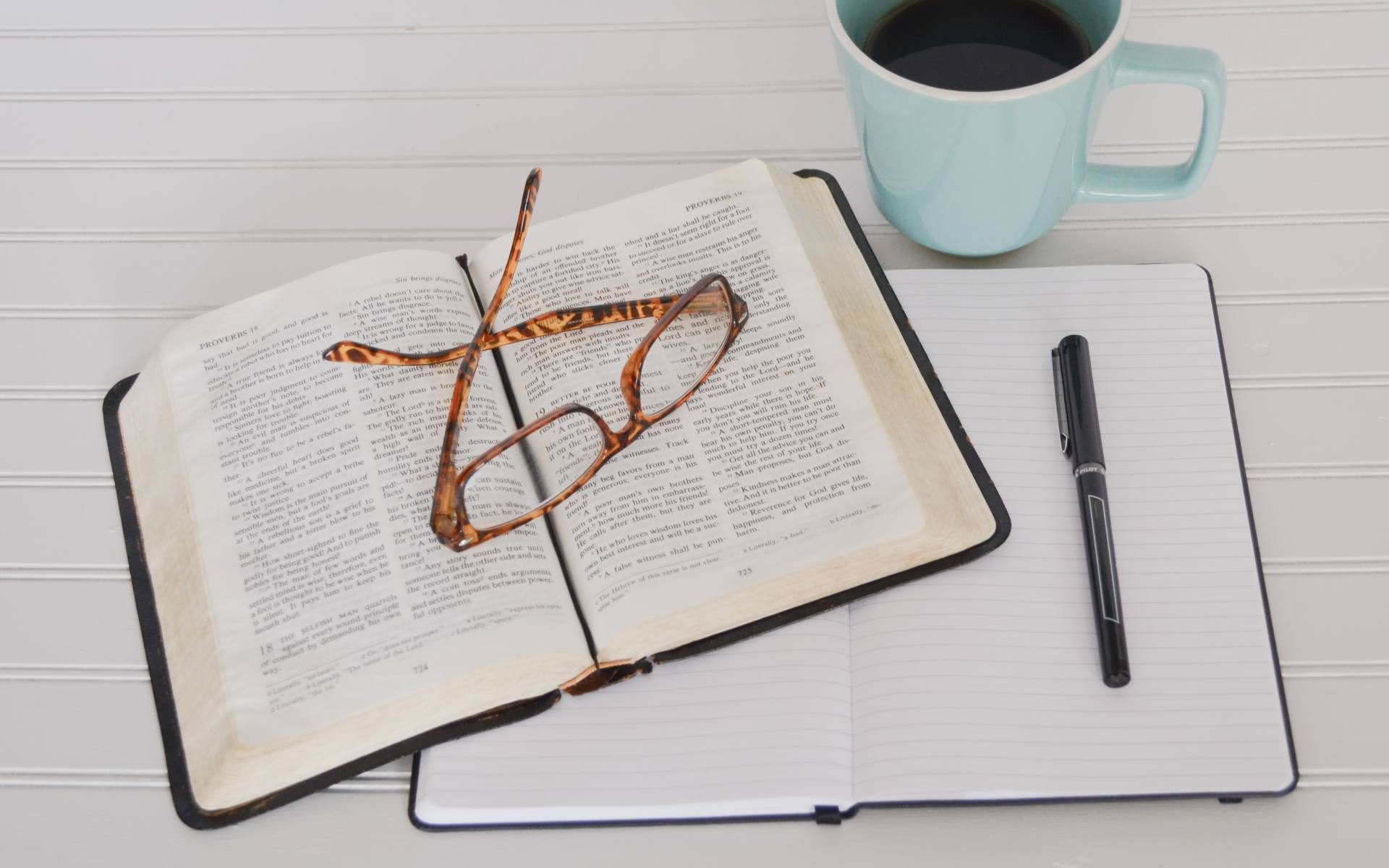}}\\[-1em]
\\
CV: cup, \hl{book}, knife, art gallery,\\cigarette holder,
MAN: \hl{New Zealand},\\
\hl{educational rankings},
BD: \hl{books}}
										 & \vspace{-1.8cm}\hl{Study} in New Zealand and your \hl{CV} will gain an instant boost when it is time to go \hl{job} hunting.
                                        &\vspace{-1.8cm} The land of the \hl{Kiwis} consistently tops \hl{educational rankings} and has many top ranking universities, with 5 of them in the top 300 in the world. Furthermore, teachers are highly educated and more often than not researchers themselves. Active participation, creativity and understanding of different perspectives are some of the many qualities you can pick up by studying in \hl{New Zealand}.
                                        &\vspace{-1.8cm}New Zealand has countless \hl{student}-friendly cities. Therefore, it should hardly come as a surprise that New Zealand is constantly ranked as a top \hl{study} abroad destination. To name but a few cities: Auckland, North Palmerston, Wellington and Christchurch all have fantastic services and great \hl{universities} where you can study and live to your heart's content.\\[0.5em]
\hline
\end{tabular}%
\label{example-studyAbroad}%
\end{table*}%


\section{Conclusion}

In this paper we have introduced the problem of story-images alignment -- selecting and placing a set of representative images for a story within contextual paragraphs. We analyzed features towards meaningful alignments from two real-world multimodal datasets -- Lonely Planet and Asia Exchange blogs -- and defined measures to evaluate text-image alignments. We presented SANDI, a methodology for automating such alignments by a constrained optimization problem maximizing semantic coherence between text-image pairs jointly for the entire story. Evaluations show that SANDI produces alignments which are semantically meaningful. Nevertheless, some follow-up questions arise.

\ourheading{Additional Features}
While our feature space covers most natural aspects, in downstream applications additional image metadata such as GPS location or timestamps may be available. GPS location may provide cues for geographic named entities, lifting the reliance on user-provided tags. Timestamps might prove to be useful for temporal aspects of a storyline.

\ourheading{Abstract and Metaphoric Relations} Our alignments were focused largely on visual and contextual features. We do not address stylistic elements like metaphors and sarcasm in text, which would entail more challenging alignments. For example, the text ``the news was a dagger to his heart'' should not be paired with a picture of a dagger. Although user knowledge may provide some cues towards such abstract relationships, 
a deeper understanding of semantic coherence is desired.

\vspace{0.3em}
The datasets used in this paper, along with the image descriptors, will be made available. 


\bibliographystyle{ACM-Reference-Format}

\begin{thebibliography}{40}


\ifx \showCODEN    \undefined \def \showCODEN     #1{\unskip}     \fi
\ifx \showDOI      \undefined \def \showDOI       #1{#1}\fi
\ifx \showISBNx    \undefined \def \showISBNx     #1{\unskip}     \fi
\ifx \showISBNxiii \undefined \def \showISBNxiii  #1{\unskip}     \fi
\ifx \showISSN     \undefined \def \showISSN      #1{\unskip}     \fi
\ifx \showLCCN     \undefined \def \showLCCN      #1{\unskip}     \fi
\ifx \shownote     \undefined \def \shownote      #1{#1}          \fi
\ifx \showarticletitle \undefined \def \showarticletitle #1{#1}   \fi
\ifx \showURL      \undefined \def \showURL       {\relax}        \fi
\providecommand\bibfield[2]{#2}
\providecommand\bibinfo[2]{#2}
\providecommand\natexlab[1]{#1}
\providecommand\showeprint[2][]{arXiv:#2}

\bibitem[\protect\citeauthoryear{Bernardi, {\c{C}}akici, Elliott, Erdem, Erdem,
  Ikizler{-}Cinbis, Keller, Muscat, and Plank}{Bernardi et~al\mbox{.}}{2016}]%
        {DBLP:journals/jair/BernardiCEEEIKM16}
\bibfield{author}{\bibinfo{person}{Raffaella Bernardi}, \bibinfo{person}{Ruket
  {\c{C}}akici}, \bibinfo{person}{Desmond Elliott}, \bibinfo{person}{Aykut
  Erdem}, \bibinfo{person}{Erkut Erdem}, \bibinfo{person}{Nazli
  Ikizler{-}Cinbis}, \bibinfo{person}{Frank Keller}, \bibinfo{person}{Adrian
  Muscat}, {and} \bibinfo{person}{Barbara Plank}.}
  \bibinfo{year}{2016}\natexlab{}.
\newblock \showarticletitle{Automatic Description Generation from Images: {A}
  Survey of Models, Datasets, and Evaluation Measures}.
\newblock \bibinfo{journal}{\emph{J. Artif. Intell. Res.}}
  (\bibinfo{year}{2016}).
\newblock


\bibitem[\protect\citeauthoryear{Chowdhury, Tandon, Ferhatosmanoglu, and
  Weikum}{Chowdhury et~al\mbox{.}}{2018}]%
        {DBLP:conf/wsdm/ChowdhuryTFW18}
\bibfield{author}{\bibinfo{person}{Sreyasi~Nag Chowdhury},
  \bibinfo{person}{Niket Tandon}, \bibinfo{person}{Hakan Ferhatosmanoglu},
  {and} \bibinfo{person}{Gerhard Weikum}.} \bibinfo{year}{2018}\natexlab{}.
\newblock \showarticletitle{{VISIR:} Visual and Semantic Image Label
  Refinement}. \bibinfo{booktitle}{\emph{{WSDM}}}.
\newblock


\bibitem[\protect\citeauthoryear{Chowdhury, Tandon, and Weikum}{Chowdhury
  et~al\mbox{.}}{2016}]%
        {DBLP:conf/akbc/ChowdhuryTW16}
\bibfield{author}{\bibinfo{person}{Sreyasi~Nag Chowdhury},
  \bibinfo{person}{Niket Tandon}, {and} \bibinfo{person}{Gerhard Weikum}.}
  \bibinfo{year}{2016}\natexlab{}.
\newblock \showarticletitle{Know2Look: Commonsense Knowledge for Visual
  Search}. \bibinfo{booktitle}{\emph{AKBC}}.
\newblock


\bibitem[\protect\citeauthoryear{Chu and Kao}{Chu and Kao}{2017}]%
        {DBLP:conf/ism/ChuK17}
\bibfield{author}{\bibinfo{person}{Wei{-}Ta Chu} {and}
  \bibinfo{person}{Ming{-}Chih Kao}.} \bibinfo{year}{2017}\natexlab{}.
\newblock \showarticletitle{Blog Article Summarization with Image-Text
  Alignment Techniques}. \bibinfo{booktitle}{\emph{ISM}}.
\newblock


\bibitem[\protect\citeauthoryear{Delgado, Magalh{\~{a}}es, and Correia}{Delgado
  et~al\mbox{.}}{2010}]%
        {DBLP:conf/semco/DelgadoMC10}
\bibfield{author}{\bibinfo{person}{Diogo Delgado}, \bibinfo{person}{Jo{\~{a}}o
  Magalh{\~{a}}es}, {and} \bibinfo{person}{Nuno Correia}.}
  \bibinfo{year}{2010}\natexlab{}.
\newblock \showarticletitle{Automated Illustration of News Stories}.
  \bibinfo{booktitle}{\emph{ICSC}}.
\newblock


\bibitem[\protect\citeauthoryear{Faghri, Fleet, Kiros, and Fidler}{Faghri
  et~al\mbox{.}}{2018}]%
        {DBLP:conf/bmvc/FaghriFKF18}
\bibfield{author}{\bibinfo{person}{Fartash Faghri}, \bibinfo{person}{David~J.
  Fleet}, \bibinfo{person}{Jamie Kiros}, {and} \bibinfo{person}{Sanja Fidler}.}
  \bibinfo{year}{2018}\natexlab{}.
\newblock \showarticletitle{{VSE++:} Improving Visual-Semantic Embeddings with
  Hard Negatives}. \bibinfo{booktitle}{\emph{{BMVC}}}.
\newblock


\bibitem[\protect\citeauthoryear{Frome, Corrado, Shlens, Bengio, Dean, Ranzato,
  and Mikolov}{Frome et~al\mbox{.}}{2013}]%
        {DBLP:conf/nips/FromeCSBDRM13}
\bibfield{author}{\bibinfo{person}{Andrea Frome}, \bibinfo{person}{Gregory~S.
  Corrado}, \bibinfo{person}{Jonathon Shlens}, \bibinfo{person}{Samy Bengio},
  \bibinfo{person}{Jeffrey Dean}, \bibinfo{person}{Marc'Aurelio Ranzato}, {and}
  \bibinfo{person}{Tomas Mikolov}.} \bibinfo{year}{2013}\natexlab{}.
\newblock \showarticletitle{DeViSE: {A} Deep Visual-Semantic Embedding Model}.
  \bibinfo{booktitle}{\emph{NIPS}}.
\newblock


\bibitem[\protect\citeauthoryear{Gan, Gan, He, Gao, and Deng}{Gan
  et~al\mbox{.}}{2017}]%
        {DBLP:conf/cvpr/GanGHGD17}
\bibfield{author}{\bibinfo{person}{Chuang Gan}, \bibinfo{person}{Zhe Gan},
  \bibinfo{person}{Xiaodong He}, \bibinfo{person}{Jianfeng Gao}, {and}
  \bibinfo{person}{Li Deng}.} \bibinfo{year}{2017}\natexlab{}.
\newblock \showarticletitle{StyleNet: Generating Attractive Visual Captions
  with Styles}. \bibinfo{booktitle}{\emph{CVPR}}.
\newblock


\bibitem[\protect\citeauthoryear{Gkioxari, Girshick, and Malik}{Gkioxari
  et~al\mbox{.}}{2015}]%
        {DBLP:conf/iccv/GkioxariGM15}
\bibfield{author}{\bibinfo{person}{Georgia Gkioxari}, \bibinfo{person}{Ross~B.
  Girshick}, {and} \bibinfo{person}{Jitendra Malik}.}
  \bibinfo{year}{2015}\natexlab{}.
\newblock \showarticletitle{Contextual Action Recognition with R*CNN}.
  \bibinfo{booktitle}{\emph{ICCV}}.
\newblock


\bibitem[\protect\citeauthoryear{Gupta, Li, Yin, and Han}{Gupta
  et~al\mbox{.}}{2010}]%
        {DBLP:journals/sigkdd/GuptaLYH10}
\bibfield{author}{\bibinfo{person}{Manish Gupta}, \bibinfo{person}{Rui Li},
  \bibinfo{person}{Zhijun Yin}, {and} \bibinfo{person}{Jiawei Han}.}
  \bibinfo{year}{2010}\natexlab{}.
\newblock \showarticletitle{Survey on social tagging techniques}.
\newblock \bibinfo{journal}{\emph{{SIGKDD} Explorations}}
  (\bibinfo{year}{2010}).
\newblock


\bibitem[\protect\citeauthoryear{Hoffman, Guadarrama, Tzeng, Hu, Donahue,
  Girshick, Darrell, and Saenko}{Hoffman et~al\mbox{.}}{2014}]%
        {DBLP:conf/nips/HoffmanGTHDGDS14}
\bibfield{author}{\bibinfo{person}{Judy Hoffman}, \bibinfo{person}{Sergio
  Guadarrama}, \bibinfo{person}{Eric Tzeng}, \bibinfo{person}{Ronghang Hu},
  \bibinfo{person}{Jeff Donahue}, \bibinfo{person}{Ross~B. Girshick},
  \bibinfo{person}{Trevor Darrell}, {and} \bibinfo{person}{Kate Saenko}.}
  \bibinfo{year}{2014}\natexlab{}.
\newblock \showarticletitle{{LSDA:} Large Scale Detection through Adaptation}.
  \bibinfo{booktitle}{\emph{NIPS}}.
\newblock


\bibitem[\protect\citeauthoryear{Joshi, Wang, and Li}{Joshi
  et~al\mbox{.}}{2006}]%
        {DBLP:journals/tomccap/JoshiWL06}
\bibfield{author}{\bibinfo{person}{Dhiraj Joshi}, \bibinfo{person}{James~Ze
  Wang}, {and} \bibinfo{person}{Jia Li}.} \bibinfo{year}{2006}\natexlab{}.
\newblock \showarticletitle{The Story Picturing Engine - a system for automatic
  text illustration}.
\newblock \bibinfo{journal}{\emph{{TOMCCAP}}} \bibinfo{volume}{2},
  \bibinfo{number}{1} (\bibinfo{year}{2006}), \bibinfo{pages}{68--89}.
\newblock


\bibitem[\protect\citeauthoryear{Karpathy and Li}{Karpathy and Li}{2015}]%
        {DBLP:conf/cvpr/KarpathyL15}
\bibfield{author}{\bibinfo{person}{Andrej Karpathy} {and}
  \bibinfo{person}{Fei{-}Fei Li}.} \bibinfo{year}{2015}\natexlab{}.
\newblock \showarticletitle{Deep visual-semantic alignments for generating
  image descriptions}. \bibinfo{booktitle}{\emph{CVPR}}.
\newblock


\bibitem[\protect\citeauthoryear{Kiros, Salakhutdinov, and Zemel}{Kiros
  et~al\mbox{.}}{2014}]%
        {DBLP:journals/corr/KirosSZ14}
\bibfield{author}{\bibinfo{person}{Ryan Kiros}, \bibinfo{person}{Ruslan
  Salakhutdinov}, {and} \bibinfo{person}{Richard~S. Zemel}.}
  \bibinfo{year}{2014}\natexlab{}.
\newblock \showarticletitle{Unifying Visual-Semantic Embeddings with Multimodal
  Neural Language Models}.
\newblock \bibinfo{journal}{\emph{CoRR}} (\bibinfo{year}{2014}).
\newblock


\bibitem[\protect\citeauthoryear{Kong, Lin, Bansal, Urtasun, and Fidler}{Kong
  et~al\mbox{.}}{2014}]%
        {DBLP:conf/cvpr/KongLBUF14}
\bibfield{author}{\bibinfo{person}{Chen Kong}, \bibinfo{person}{Dahua Lin},
  \bibinfo{person}{Mohit Bansal}, \bibinfo{person}{Raquel Urtasun}, {and}
  \bibinfo{person}{Sanja Fidler}.} \bibinfo{year}{2014}\natexlab{}.
\newblock \showarticletitle{What Are You Talking About? Text-to-Image
  Coreference}. \bibinfo{booktitle}{\emph{CVPR}}.
\newblock


\bibitem[\protect\citeauthoryear{Krause, Johnson, Krishna, and Fei}{Krause
  et~al\mbox{.}}{2017}]%
        {DBLP:conf/cvpr/KrauseJKF17}
\bibfield{author}{\bibinfo{person}{Jonathan Krause}, \bibinfo{person}{Justin
  Johnson}, \bibinfo{person}{Ranjay Krishna}, {and} \bibinfo{person}{Li~Fei
  Fei}.} \bibinfo{year}{2017}\natexlab{}.
\newblock \showarticletitle{A Hierarchical Approach for Generating Descriptive
  Image Paragraphs}. \bibinfo{booktitle}{\emph{CVPR}}.
\newblock


\bibitem[\protect\citeauthoryear{Lester}{Lester}{2013}]%
        {lester2013visual}
\bibfield{author}{\bibinfo{person}{Paul~Martin Lester}.}
  \bibinfo{year}{2013}\natexlab{}.
\newblock \bibinfo{booktitle}{\emph{Visual communication: Images with
  messages}}.
\newblock \bibinfo{publisher}{Cengage Learning}.
\newblock


\bibitem[\protect\citeauthoryear{Lieberman and Liu}{Lieberman and Liu}{2002}]%
        {DBLP:conf/ah/LiebermanL02}
\bibfield{author}{\bibinfo{person}{Henry Lieberman} {and} \bibinfo{person}{Hugo
  Liu}.} \bibinfo{year}{2002}\natexlab{}.
\newblock \showarticletitle{Adaptive Linking between Text and Photos Using
  Common Sense Reasoning}. \bibinfo{booktitle}{\emph{AH}}.
\newblock


\bibitem[\protect\citeauthoryear{Lin, Maire, Belongie, Hays, Perona, Ramanan,
  Doll{\'{a}}r, and Zitnick}{Lin et~al\mbox{.}}{2014}]%
        {DBLP:conf/eccv/LinMBHPRDZ14}
\bibfield{author}{\bibinfo{person}{Tsung{-}Yi Lin}, \bibinfo{person}{Michael
  Maire}, \bibinfo{person}{Serge~J. Belongie}, \bibinfo{person}{James Hays},
  \bibinfo{person}{Pietro Perona}, \bibinfo{person}{Deva Ramanan},
  \bibinfo{person}{Piotr Doll{\'{a}}r}, {and} \bibinfo{person}{C.~Lawrence
  Zitnick}.} \bibinfo{year}{2014}\natexlab{}.
\newblock \showarticletitle{Microsoft {COCO:} Common Objects in Context}.
  \bibinfo{booktitle}{\emph{ECCV}}.
\newblock


\bibitem[\protect\citeauthoryear{Lu, Krishna, Bernstein, and Li}{Lu
  et~al\mbox{.}}{2016}]%
        {DBLP:conf/eccv/LuKBL16}
\bibfield{author}{\bibinfo{person}{Cewu Lu}, \bibinfo{person}{Ranjay Krishna},
  \bibinfo{person}{Michael~S. Bernstein}, {and} \bibinfo{person}{Fei{-}Fei
  Li}.} \bibinfo{year}{2016}\natexlab{}.
\newblock \showarticletitle{Visual Relationship Detection with Language
  Priors}. \bibinfo{booktitle}{\emph{ECCV}}.
\newblock


\bibitem[\protect\citeauthoryear{Lu, Xiong, Parikh, and Socher}{Lu
  et~al\mbox{.}}{2017}]%
        {DBLP:conf/cvpr/LuXPS17}
\bibfield{author}{\bibinfo{person}{Jiasen Lu}, \bibinfo{person}{Caiming Xiong},
  \bibinfo{person}{Devi Parikh}, {and} \bibinfo{person}{Richard Socher}.}
  \bibinfo{year}{2017}\natexlab{}.
\newblock \showarticletitle{Knowing When to Look: Adaptive Attention via a
  Visual Sentinel for Image Captioning}. \bibinfo{booktitle}{\emph{CVPR}}.
\newblock


\bibitem[\protect\citeauthoryear{Messaris and Abraham}{Messaris and
  Abraham}{2001}]%
        {messaris2001role}
\bibfield{author}{\bibinfo{person}{Paul Messaris} {and} \bibinfo{person}{Linus
  Abraham}.} \bibinfo{year}{2001}\natexlab{}.
\newblock \showarticletitle{The role of images in framing news stories}.
\newblock \bibinfo{booktitle}{\emph{Framing public life}}.
  \bibinfo{publisher}{Routledge}, \bibinfo{pages}{231--242}.
\newblock


\bibitem[\protect\citeauthoryear{Mikolov, Sutskever, Chen, Corrado, and
  Dean}{Mikolov et~al\mbox{.}}{2013}]%
        {DBLP:conf/nips/MikolovSCCD13}
\bibfield{author}{\bibinfo{person}{Tomas Mikolov}, \bibinfo{person}{Ilya
  Sutskever}, \bibinfo{person}{Kai Chen}, \bibinfo{person}{Gregory~S. Corrado},
  {and} \bibinfo{person}{Jeffrey Dean}.} \bibinfo{year}{2013}\natexlab{}.
\newblock \showarticletitle{Distributed Representations of Words and Phrases
  and their Compositionality}. \bibinfo{booktitle}{\emph{NIPS}}.
\newblock


\bibitem[\protect\citeauthoryear{Ravi, Wang, Mu{\~{n}}iz, Sigal, Metaxas, and
  Kapadia}{Ravi et~al\mbox{.}}{2018}]%
        {DBLP:conf/cvpr/RaviWMSMK18}
\bibfield{author}{\bibinfo{person}{Hareesh Ravi}, \bibinfo{person}{Lezi Wang},
  \bibinfo{person}{Carlos Mu{\~{n}}iz}, \bibinfo{person}{Leonid Sigal},
  \bibinfo{person}{Dimitris~N. Metaxas}, {and} \bibinfo{person}{Mubbasir
  Kapadia}.} \bibinfo{year}{2018}\natexlab{}.
\newblock \showarticletitle{Show Me a Story: Towards Coherent Neural Story
  Illustration}. \bibinfo{booktitle}{\emph{CVPR}}.
\newblock


\bibitem[\protect\citeauthoryear{Redmon and Farhadi}{Redmon and
  Farhadi}{2017}]%
        {DBLP:conf/cvpr/RedmonF17}
\bibfield{author}{\bibinfo{person}{Joseph Redmon} {and} \bibinfo{person}{Ali
  Farhadi}.} \bibinfo{year}{2017}\natexlab{}.
\newblock \showarticletitle{{YOLO9000:} Better, Faster, Stronger}.
  \bibinfo{booktitle}{\emph{CVPR}}.
\newblock


\bibitem[\protect\citeauthoryear{Ren, He, Girshick, and Sun}{Ren
  et~al\mbox{.}}{2015}]%
        {DBLP:conf/nips/RenHGS15}
\bibfield{author}{\bibinfo{person}{Shaoqing Ren}, \bibinfo{person}{Kaiming He},
  \bibinfo{person}{Ross~B. Girshick}, {and} \bibinfo{person}{Jian Sun}.}
  \bibinfo{year}{2015}\natexlab{}.
\newblock \showarticletitle{Faster {R-CNN:} Towards Real-Time Object Detection
  with Region Proposal Networks}. \bibinfo{booktitle}{\emph{NIPS}}.
\newblock


\bibitem[\protect\citeauthoryear{Schwarz, Rojtberg, Caspar, Gurevych, Goesele,
  and Lensch}{Schwarz et~al\mbox{.}}{2010}]%
        {DBLP:conf/kes/SchwarzRCGGL10}
\bibfield{author}{\bibinfo{person}{Katharina Schwarz}, \bibinfo{person}{Pavel
  Rojtberg}, \bibinfo{person}{Joachim Caspar}, \bibinfo{person}{Iryna
  Gurevych}, \bibinfo{person}{Michael Goesele}, {and} \bibinfo{person}{Hendrik
  P.~A. Lensch}.} \bibinfo{year}{2010}\natexlab{}.
\newblock \showarticletitle{Text-to-Video: Story Illustration from Online Photo
  Collections}. \bibinfo{booktitle}{\emph{KES}}.
\newblock


\bibitem[\protect\citeauthoryear{Speer, Chin, and Havasi}{Speer
  et~al\mbox{.}}{2017}]%
        {DBLP:conf/aaai/SpeerCH17}
\bibfield{author}{\bibinfo{person}{Robert Speer}, \bibinfo{person}{Joshua
  Chin}, {and} \bibinfo{person}{Catherine Havasi}.}
  \bibinfo{year}{2017}\natexlab{}.
\newblock \showarticletitle{ConceptNet 5.5: An Open Multilingual Graph of
  General Knowledge}. \bibinfo{booktitle}{\emph{AAAI}}.
\newblock


\bibitem[\protect\citeauthoryear{Tan and Chan}{Tan and Chan}{2016}]%
        {DBLP:conf/accv/TanC16}
\bibfield{author}{\bibinfo{person}{Ying~Hua Tan} {and}
  \bibinfo{person}{Chee~Seng Chan}.} \bibinfo{year}{2016}\natexlab{}.
\newblock \showarticletitle{phi-LSTM: {A} Phrase-Based Hierarchical {LSTM}
  Model for Image Captioning}. \bibinfo{booktitle}{\emph{ACCV}}.
\newblock


\bibitem[\protect\citeauthoryear{Vendrov, Kiros, Fidler, and Urtasun}{Vendrov
  et~al\mbox{.}}{2016}]%
        {DBLP:journals/corr/VendrovKFU15}
\bibfield{author}{\bibinfo{person}{Ivan Vendrov}, \bibinfo{person}{Ryan Kiros},
  \bibinfo{person}{Sanja Fidler}, {and} \bibinfo{person}{Raquel Urtasun}.}
  \bibinfo{year}{2016}\natexlab{}.
\newblock \showarticletitle{Order-Embeddings of Images and Language}.
  \bibinfo{booktitle}{\emph{ICLR}}.
\newblock


\bibitem[\protect\citeauthoryear{Williams, Lieberman, and Winston}{Williams
  et~al\mbox{.}}{2017}]%
        {DBLP:conf/commonsense/WilliamsLW17}
\bibfield{author}{\bibinfo{person}{Bryan Williams}, \bibinfo{person}{Henry
  Lieberman}, {and} \bibinfo{person}{Patrick~H. Winston}.}
  \bibinfo{year}{2017}\natexlab{}.
\newblock \showarticletitle{Understanding Stories with Large-Scale Common
  Sense}. \bibinfo{booktitle}{\emph{COMMONSENSE}}.
\newblock


\bibitem[\protect\citeauthoryear{Wu, Shen, Liu, Dick, and van~den Hengel}{Wu
  et~al\mbox{.}}{2016}]%
        {DBLP:conf/cvpr/WuSLDH16}
\bibfield{author}{\bibinfo{person}{Qi Wu}, \bibinfo{person}{Chunhua Shen},
  \bibinfo{person}{Lingqiao Liu}, \bibinfo{person}{Anthony~R. Dick}, {and}
  \bibinfo{person}{Anton van~den Hengel}.} \bibinfo{year}{2016}\natexlab{}.
\newblock \showarticletitle{What Value Do Explicit High Level Concepts Have in
  Vision to Language Problems?}. \bibinfo{booktitle}{\emph{CVPR}}.
\newblock


\bibitem[\protect\citeauthoryear{Wu and Giles}{Wu and Giles}{2013}]%
        {DBLP:conf/naacl/WuG13}
\bibfield{author}{\bibinfo{person}{Zhaohui Wu} {and} \bibinfo{person}{C.~Lee
  Giles}.} \bibinfo{year}{2013}\natexlab{}.
\newblock \showarticletitle{Measuring Term Informativeness in Context}.
  \bibinfo{booktitle}{\emph{NAACL HLT}}.
\newblock


\bibitem[\protect\citeauthoryear{Xu, Ba, Kiros, Cho, Courville, Salakhutdinov,
  Zemel, and Bengio}{Xu et~al\mbox{.}}{2015}]%
        {DBLP:conf/icml/XuBKCCSZB15}
\bibfield{author}{\bibinfo{person}{Kelvin Xu}, \bibinfo{person}{Jimmy Ba},
  \bibinfo{person}{Ryan Kiros}, \bibinfo{person}{Kyunghyun Cho},
  \bibinfo{person}{Aaron~C. Courville}, \bibinfo{person}{Ruslan Salakhutdinov},
  \bibinfo{person}{Richard~S. Zemel}, {and} \bibinfo{person}{Yoshua Bengio}.}
  \bibinfo{year}{2015}\natexlab{}.
\newblock \showarticletitle{Show, Attend and Tell: Neural Image Caption
  Generation with Visual Attention}. \bibinfo{booktitle}{\emph{ICML}}.
\newblock


\bibitem[\protect\citeauthoryear{Yao, Jiang, Khosla, Lin, Guibas, and Li}{Yao
  et~al\mbox{.}}{2011}]%
        {DBLP:conf/iccv/YaoJKLGF11}
\bibfield{author}{\bibinfo{person}{Bangpeng Yao}, \bibinfo{person}{Xiaoye
  Jiang}, \bibinfo{person}{Aditya Khosla}, \bibinfo{person}{Andy~Lai Lin},
  \bibinfo{person}{Leonidas~J. Guibas}, {and} \bibinfo{person}{Fei-Fei Li}.}
  \bibinfo{year}{2011}\natexlab{}.
\newblock \showarticletitle{Human action recognition by learning bases of
  action attributes and parts}. \bibinfo{booktitle}{\emph{ICCV}}.
\newblock


\bibitem[\protect\citeauthoryear{Zhang, Qu, Peng, and Fan}{Zhang
  et~al\mbox{.}}{2017}]%
        {DBLP:journals/mta/ZhangQPF17}
\bibfield{author}{\bibinfo{person}{Baopeng Zhang}, \bibinfo{person}{Yanyun Qu},
  \bibinfo{person}{Jinye Peng}, {and} \bibinfo{person}{Jianping Fan}.}
  \bibinfo{year}{2017}\natexlab{}.
\newblock \showarticletitle{An automatic image-text alignment method for
  large-scale web image retrieval}.
\newblock \bibinfo{journal}{\emph{Multimedia Tools Appl.}}
  \bibinfo{volume}{76}, \bibinfo{number}{20} (\bibinfo{year}{2017}),
  \bibinfo{pages}{21401--21421}.
\newblock


\bibitem[\protect\citeauthoryear{Zhao, Ma, and You}{Zhao et~al\mbox{.}}{2017}]%
        {DBLP:conf/iccv/ZhaoMY17}
\bibfield{author}{\bibinfo{person}{Zhichen Zhao}, \bibinfo{person}{Huimin Ma},
  {and} \bibinfo{person}{Shaodi You}.} \bibinfo{year}{2017}\natexlab{}.
\newblock \showarticletitle{Single Image Action Recognition Using Semantic Body
  Part Actions}. \bibinfo{booktitle}{\emph{ICCV}}.
\newblock


\bibitem[\protect\citeauthoryear{Zhou, Lapedriza, Xiao, Torralba, and
  Oliva}{Zhou et~al\mbox{.}}{2014}]%
        {DBLP:conf/nips/ZhouLXTO14}
\bibfield{author}{\bibinfo{person}{Bolei Zhou}, \bibinfo{person}{{\`{A}}gata
  Lapedriza}, \bibinfo{person}{Jianxiong Xiao}, \bibinfo{person}{Antonio
  Torralba}, {and} \bibinfo{person}{Aude Oliva}.}
  \bibinfo{year}{2014}\natexlab{}.
\newblock \showarticletitle{Learning Deep Features for Scene Recognition using
  Places Database}. \bibinfo{booktitle}{\emph{NIPS}}.
\newblock


\bibitem[\protect\citeauthoryear{Zhou and Fan}{Zhou and Fan}{2015}]%
        {DBLP:journals/pr/ZhouF15}
\bibfield{author}{\bibinfo{person}{Ning Zhou} {and} \bibinfo{person}{Jianping
  Fan}.} \bibinfo{year}{2015}\natexlab{}.
\newblock \showarticletitle{Automatic image-text alignment for large-scale web
  image indexing and retrieval}.
\newblock \bibinfo{journal}{\emph{Pattern Recognition}} \bibinfo{volume}{48},
  \bibinfo{number}{1} (\bibinfo{year}{2015}), \bibinfo{pages}{205--219}.
\newblock


\bibitem[\protect\citeauthoryear{Zhu, Kiros, Zemel, Salakhutdinov, Urtasun,
  Torralba, and Fidler}{Zhu et~al\mbox{.}}{2015}]%
        {DBLP:conf/iccv/ZhuKZSUTF15}
\bibfield{author}{\bibinfo{person}{Yukun Zhu}, \bibinfo{person}{Ryan Kiros},
  \bibinfo{person}{Richard~S. Zemel}, \bibinfo{person}{Ruslan Salakhutdinov},
  \bibinfo{person}{Raquel Urtasun}, \bibinfo{person}{Antonio Torralba}, {and}
  \bibinfo{person}{Sanja Fidler}.} \bibinfo{year}{2015}\natexlab{}.
\newblock \showarticletitle{Aligning Books and Movies: Towards Story Like
  Visual Explanations by Watching Movies and Reading Books}.
  \bibinfo{booktitle}{\emph{ICCV}}.
\newblock


\end{thebibliography}


\end{document}